\documentclass[10pt,twocolumn,letterpaper]{article}
\usepackage[pagebackref=true,breaklinks=true,letterpaper=true,colorlinks,bookmarks=false]{hyperref}

\usepackage{cvpr}
\usepackage{times}
\usepackage{epsfig}
\usepackage{graphicx}
\usepackage{amsmath}
\usepackage{amssymb}

\usepackage{booktabs}
\usepackage{multirow}
\usepackage{algorithm}
\usepackage{algpseudocode}

\DeclareMathOperator*{\argmin}{arg\,min}

\newcommand{\ve}[1]{\mathbf{#1}}
\newcommand{\figref}[1]{Fig.~\ref{#1}}
\newcommand{\tabref}[1]{Tab.~\ref{#1}}
\newcommand{\equref}[1]{Equation~(\ref{#1})}

\makeatletter
\DeclareRobustCommand\onedot{\futurelet\@let@token\@onedot}
\def\@onedot{\ifx\@let@token.\else.\null\fi\xspace}
\def\eg{\emph{e.g}\onedot} 
\def\ie{\emph{i.e}\onedot}

\def\etal{\emph{et al}\onedot}
\makeatother

\cvprfinalcopy 

\ifcvprfinal\fi
\begin{document}

\title{DIST: Rendering Deep Implicit Signed Distance Function \\ with Differentiable Sphere Tracing}

\author{
Shaohui Liu$^{1,3}$
\thanks{Work done while Shaohui Liu was an academic guest at ETH Zurich.}
\quad Yinda Zhang$^{2}$ \quad Songyou Peng$^{1,6}$ \quad Boxin Shi$^{4,7}$ \\ Marc Pollefeys$^{1,5,6}$ \quad Zhaopeng Cui$^{1}\thanks{Corresponding author.}$
\\
$^1$ETH Zurich \quad $^2$Google \quad$^3$Tsinghua University \quad $^4$Peking University \quad $^5$Microsoft
\\
$^6$Max Planck ETH Center for Learing Systems \quad $^7$Peng Cheng Laboratory
}

\maketitle
\thispagestyle{empty}

\begin{abstract}

We propose a differentiable sphere tracing algorithm to bridge the gap between inverse graphics methods and the recently proposed deep learning based implicit signed distance function.
Due to the nature of the implicit function, the rendering process requires tremendous function queries, which is particularly problematic when the function is represented as a neural network.
We optimize both the forward and backward passes of our rendering layer to make it run efficiently with affordable memory consumption on a commodity graphics card.
Our rendering method is fully differentiable such that losses can be directly computed on the rendered 2D observations, and the gradients can be propagated backwards to optimize the 3D geometry.
We show that our rendering method can effectively reconstruct accurate 3D shapes from various inputs, such as sparse depth and multi-view images, through inverse optimization.
With the geometry based reasoning, 
our 3D shape prediction methods show excellent generalization capability and robustness against various noises.

\end{abstract}


\section{Introduction}
Solving vision problem as an inverse graphics process is one of the most fundamental approaches, where the solution is the visual structure that best explains the given observations. In the realm of 3D geometry understanding, this approach has been used since the very early age~\cite{baumgart1974geometric, patow2003survey, yu1999inverse}.
As a critical component to the inverse graphics based 3D geometric reasoning process, an efficient renderer is required to accurately simulate the observations, \eg, depth maps, from an optimizable 3D structure, and also be differentiable to back-propagate the error from the partial observation.

As a natural fit to the deep learning framework, differentiable rendering techniques have drawn great interests recently.
Various solutions for different 3D representations, \eg, voxels, point clouds, meshes, have been proposed.
However, these 3D representations are all discretized up to a certain resolution, leading to the loss of geometric details and breaking the differentiable properties \cite{li2018differentiable}.
Recently, the continuous implicit function has been used to represent the signed distance field \cite{park2019deepsdf}, which has premium capacity to encode accurate geometry when combined with the deep learning techniques.
Given a latent code as the shape representation, the function can produce a signed distance value for any arbitrary point, and thus enable unlimited resolution and better preserved geometric details for rendering purpose.
However, a differentiable rendering solution for learning-based continuous signed distance function does not exist yet.

\begin{figure}[tb]
{
\includegraphics[width=0.96\linewidth]{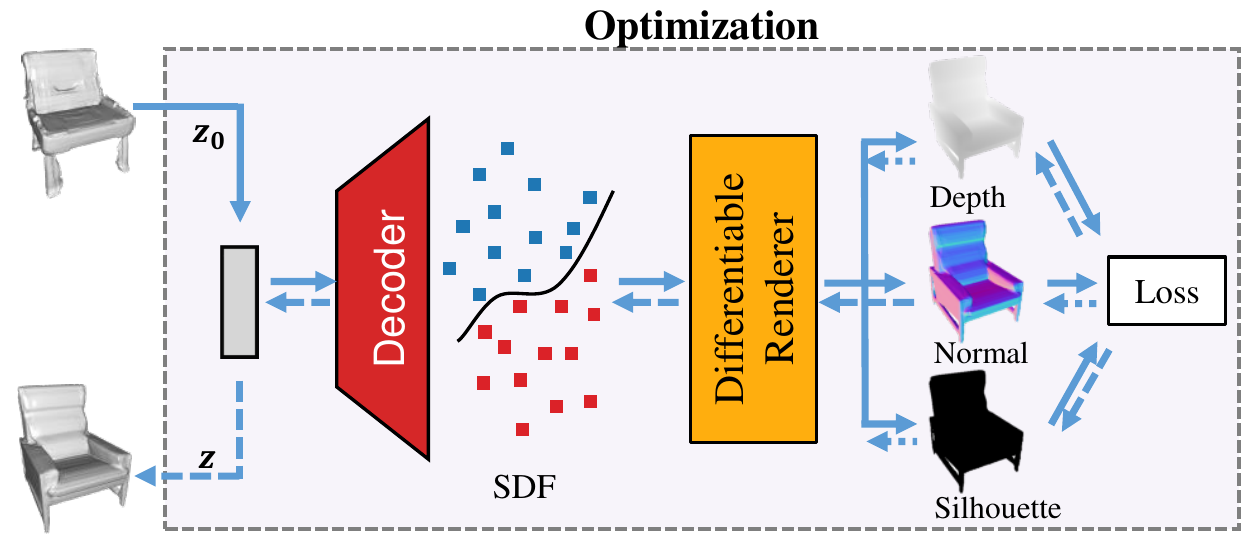}}
\centering
\caption{Illustration of our proposed differentiable renderer for continuous signed distance function. Our method enables geometric reasoning with strong generalization capability. With a random shape code $z_0$ initialized in the learned shape space, we can acquire high-quality 3D shape prediction by performing iterative optimization with various 2D supervisions.}
\vspace{-5pt}
\label{fig::teaser}
\end{figure}

In this paper, we propose a differentiable renderer for continuous implicit signed distance functions (SDF) to facilitate the 3D shape understanding via geometric reasoning in a deep learning framework (\figref{fig::teaser}). 
Our method can render an implicit SDF represented by a neural network from a latent code into various 2D observations, 
\eg, depth images, surface normals, silhouettes, and other properties encoded, from arbitrary camera viewpoints.
The rendering process is fully differentiable, such that loss functions can be conveniently defined on the rendered images and the observations, and the gradients can be propagated back to the shape generator. 
As major applications, our differentiable renderer can be applied to infer the 3D shape from various inputs, \eg, multi-view images and single depth image, through an inverse graphics process.
Specifically, given a pre-trained generative model, \eg, DeepSDF~\cite{park2019deepsdf}, we search within the latent code space for the 3D shape that produces the rendered images mostly consistent with the observation.
Extensive experiments show that our geometric reasoning based approach exhibits significantly better generalization capability than previous purely learning based approaches, and consistently produce accurate 3D shapes across datasets without finetuning.

Nevertheless, it is challenging to make differentiable rendering work on a learning-based implicit SDF with computationally affordable resources.
The main obstacle is that an implicit function provides neither the exact location nor any bound of the surface geometry as in other representations like meshes, voxels, and point clouds. 

\begin{figure}[tb]
{\includegraphics[width=0.95\linewidth,height=100pt]{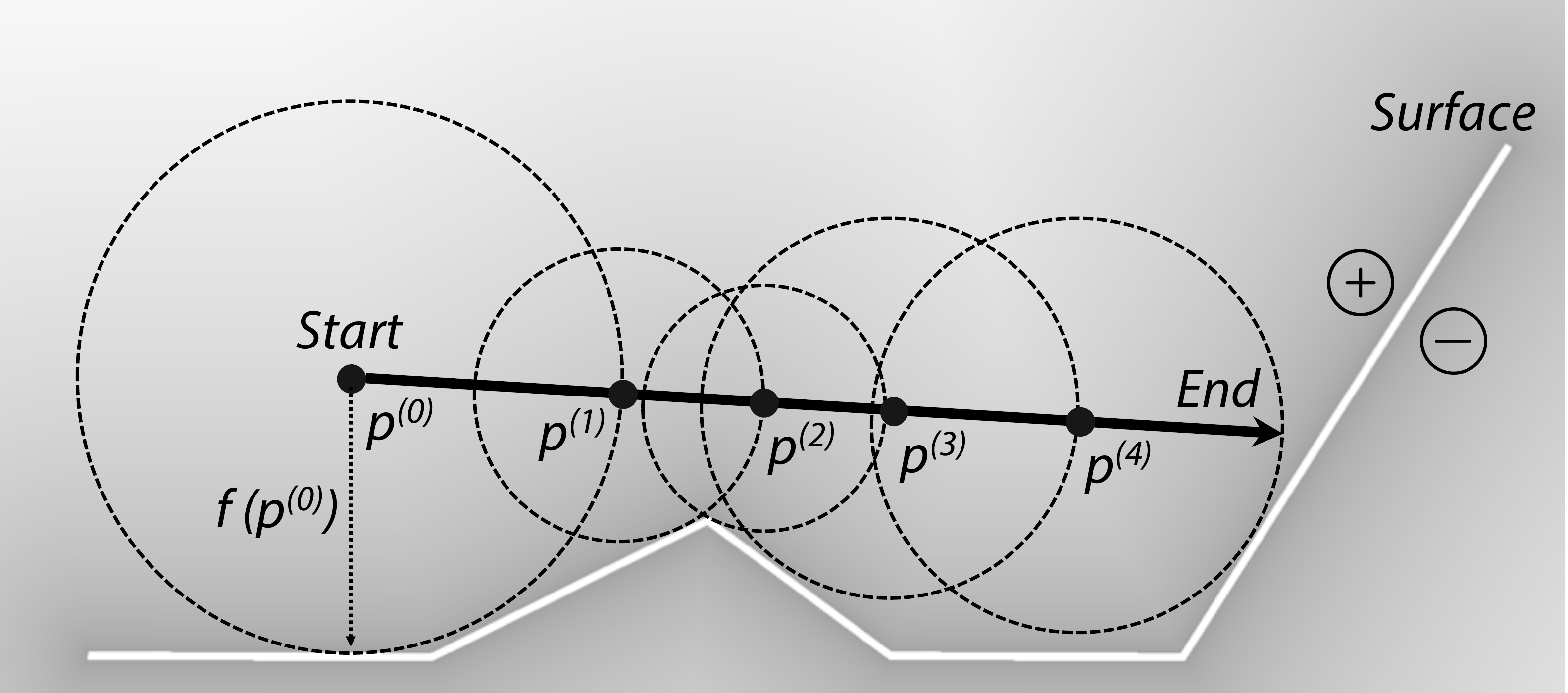}}
\centering
\caption{Illustration on the sphere tracing algorithm~\cite{hart1996sphere}. A ray is initiated at each pixel and marching along the viewing direction. The front end moves with a step size equals to the signed distance value of the current location. The algorithm converges when the current absolute SDF is smaller than a threshold, which indicates that the surface has been found.}
\vspace{-5pt}
\label{fig::sphere-tracing}
\end{figure}

Inspired by traditional ray-tracing based approaches, we adopt the sphere tracing algorithm \cite{hart1996sphere}, which marches along each pixel's ray direction with the queried signed distance until the ray hits the surface, \ie, the signed distance equals to zero (\figref{fig::sphere-tracing}). However, this is not feasible in the neural network based scenario where each query on the ray would require a forward pass and recursive computational graph for back-propagation, which is prohibitive in terms of computation and memory.

To make it work efficiently on a commodity level GPU, we optimize the full life-time of the rendering process for both forward and backward propagations. 
In the forward pass, \ie, the rendering process, we adopt a coarse-to-fine approach to save computation at initial steps, an aggressive strategy to speed up the ray marching, and a safe convergence criteria to prevent unnecessary queries and maintain resolution.
In the backward propagation, we propose a gradient approximation which empirically has negligible impact on the training performance but dramatically reduces the computation and memory consumption.
By making the rendering tractable, we show how producing 2D observations with the sphere tracing and interacting with camera extrinsics can be done in differentiable ways.

To sum up, our major contribution is to \textbf{enable efficient differentiable rendering on the implicit signed distance function represented as a neural network}.
It enables accurate 3D shape prediction via geometric reasoning in deep learning frameworks and exhibits promising generalization capability.
The differentiable renderer could also potentially benefit various vision problems thanks to the marriage of implicit SDF and inverse graphics techniques.
The code and data are available at \href{https://github.com/B1ueber2y/DIST-Renderer}{\color{cyan}{https://github.com/B1ueber2y/DIST-Renderer}}.

\section{Related Work}

\noindent
\textbf{3D Representation for Shape Learning.}
The study of 3D representations for shape learning is one of the main focuses in 3D deep learning community.
Early work quantizes shapes into 3D voxels, where each voxel contains either a binary occupancy status (occupied / not occupied) \cite{wu20153d, choy20163d, tatarchenko2017octree, riegler2017octnet, hane2017hierarchical} or a signed distance value  \cite{zeng20173dmatch, dai2017shape, stutz2018learning}.
While voxels are the most straightforward extension from the 2D image domain into the 3D geometry domain for neural network operations,
they normally require huge memory overhead and result in relatively low resolutions.
Meshes are also proposed as a more memory efficient representation for 3D shape learning \cite{wang2018pixel2mesh, groueix2018atlasnet, kong2017using, kanazawa2018end}, but the topology of meshes is normally fixed and simple. 
Many deep learning methods also utilize point clouds as the 3D representation \cite{qi2017pointnet, qi2017pointnet++}; however, the point-based representation lacks the topology information and thus makes it non-trivial to generate 3D meshes.
Very recently, the implicit functions, \eg, continuous SDF and occupancy function, are exploited as 3D representations and show much promising performance in terms of the high-frequency detail modeling and the high resolution \cite{park2019deepsdf,mescheder2019occupancy, michalkiewicz2019deep,chen2019learningimplicit}.  
Similar idea has also been used to encode other information such as textures~\cite{oechsle2019texture,saito2019pifu} and 4D dynamics~\cite{niemeyer2019occupancy}. Our work aims to design an efficient and differentiable renderer for the implicit SDF-based representation.

\noindent
\textbf{Differentiable Rendering.}
With the success of deep learning, the differentiable rendering starts to draw more attention as it is essential for end-to-end training, and solutions have been proposed for various 3D representations.
Early works focus on 3D triangulated meshes and leverage standard rasterization \cite{loper2014opendr}. Various approaches try to solve the discontinuity issue near triangle boundaries by smoothing the loss function or approximating the gradient \cite{kato2018neural,petersen2019pix2vex,liu2019soft,chen2019learning}.
Solutions for point clouds and 3D voxels are also introduced \cite{yifan2019differentiable,insafutdinov2018unsupervised, nguyen2018rendernet} to work jointly with PointNet \cite{qi2017pointnet} and 3D convolutional architectures.
However, the differentiable rendering for the implicit continuous function representation does not exist yet.
Some ray tracing based approaches are related, while they are mostly proposed for explicit representation, such as 3D voxels \cite{lombardi2019neural,nguyen2018rendernet,sitzmann2019deepvoxels,jiang2019sdfdiff} or meshes \cite{li2018differentiable}, but not implicit functions. Liu \etal \cite{liu2019learning} firstly propose to learn from 2D observations over occupancy networks \cite{mescheder2019occupancy}. However, their methods make several approximations and do not benefit from the efficiency of rendering implicit SDF.
Most related to our work, Sitzmann \etal \cite{sitzmann2019scene} propose 
a LSTM-based renderer for an implicit scene representation to generate color images, while their model focuses on simulating the rendering process with an LSTM without clear geometric meaning. This method can only generate low-resolution images due to the expensive memory consumption. 
Alternatively, our method can directly render 3D geometry represented by an implicit SDF to produce high-resolution images. It can also be applied without training to existing deep learning models.

\noindent
\textbf{3D Shape Prediction.}
3D shape prediction from 2D observations is one of the fundamental vision problems.
Early works mainly focus on multi-view reconstruction using multi-view stereo methods \cite{seitz2006comparison, hernandez2008multiview, semerjian2014new}. These purely geometry-based methods suffer from degraded performance on texture-less regions without prior knowledge \cite{cui2017polarimetric}. 
With progress of deep learning, 3D shapes can be recovered under different settings.
The simplest setting is to recover 3D shape from a single image \cite{choy20163d, girdhar2016learning, wu2016learning, johnston2017scaling}.
These systems rely heavily on priors, and are prone to weak generalization.
Deep learning based multi-view shape prediction methods \cite{yao2018mvsnet,huang2018deepmvs, im2019dpsnet, Wei_2019_CVPR,huang2019pixel2meshpp} further involve geometric constraints across views in the deep learning framework, which shows better generalization. Another thread of works \cite{dai2017shape, dai2019scan2mesh} take a single depth image as input, and the problem is usually referred as shape completion. Given the shape prior encoded in the neural network \cite{park2019deepsdf}, our rendering method can effectively predict accurate 3D object shape from a random initial shape code with various inputs, such as depth and multi-view images, through geometric optimization.  

\section{Differentiable Sphere Tracing}

\begin{figure*}[tb]
\centering
{
\small
\begin{tabular}{ccc}
     \includegraphics[width=0.28\linewidth]{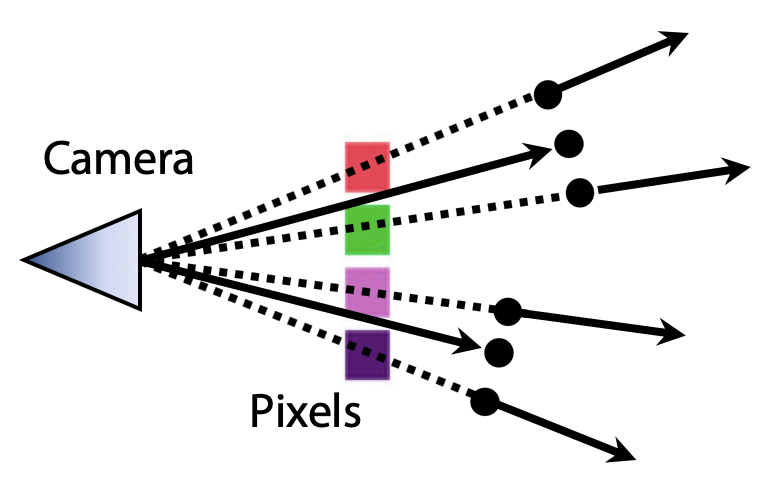}& \includegraphics[width=0.28\linewidth]{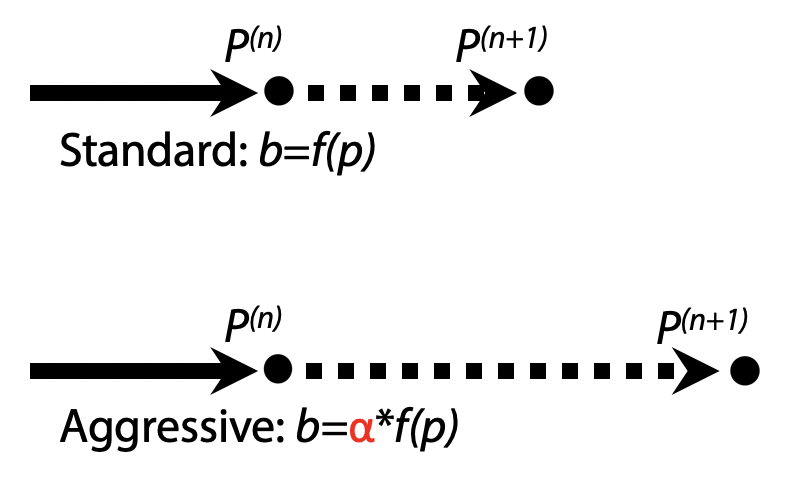} & \includegraphics[width=0.28\linewidth]{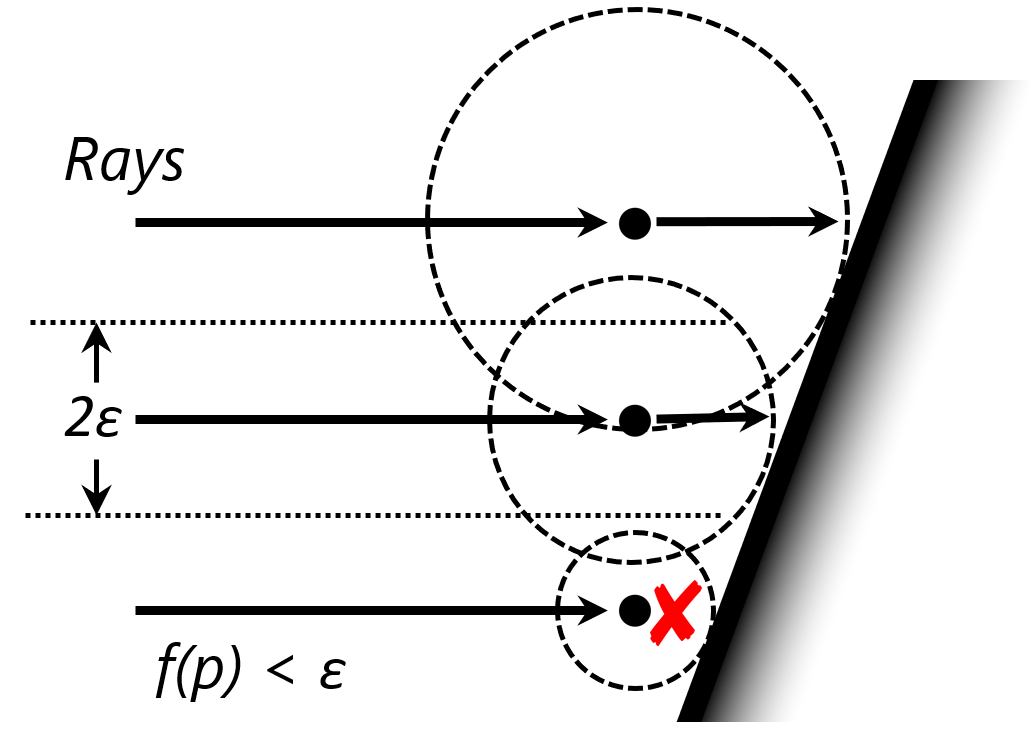} \\
     (a) Coarse-to-fine Strategy &  (b) Aggressive Marching & (c) Convergence Criteria\\
\end{tabular}
}
\caption{Strategies for our efficient forward propagation. (a) 1D illustration of our coarse-to-fine strategy, and for 2D cases, one ray will be spitted into 4 rays; (b) Comparison of standard marching and our aggressive marching; (c) We stop the marching once the SDF value is smaller than $\epsilon$, where $2\epsilon$ is the estimated minimal distance between the corresponding 3D points of two neighboring pixels.}
\vspace{-5pt}
\label{fig::lifetime}
\end{figure*}

In this section, we introduce our differentiable rendering method for the implicit signed distance function represented as a neural network, such as DeepSDF \cite{park2019deepsdf}.
In DeepSDF, a network takes a latent code and a 3D location as input, and produces the corresponding signed distance value.
Even though such a network can deliver high quality geometry, the explicit surface cannot be directly obtained and requires dense sampling in the 3D space.

Our method is inspired by Sphere Tracing \cite{hart1996sphere} designed for rendering SDF volumes, where rays are shot from the camera pinhole along the direction of each pixel to search for the surface level set according to the signed distance value.
However, it is prohibitive to apply this method directly on the implicit signed distance function represented as a neural network, since each tracing step needs a feedforward neural network and the whole algorithm requires unaffordable computational and memory resources.
To make this idea work in deep learning framework for inverse graphics, we optimize both the forward and backward propagations for efficient training and test-time optimization.
The sphere traced results, \ie, the distance along the ray, can be converted into many desired outputs, \eg, depth, surface normal, silhouette, and hence losses can be conveniently applied in an end-to-end manner.

\subsection{Preliminaries - Sphere Tracing}
To be self-contained, we first briefly introduce the traditional sphere tracing algorithm \cite{hart1996sphere}.
Sphere tracing is a conventional method specifically designed to render depth from volumetric signed distance fields.
For each pixel on the image plane, as shown in Figure \ref{fig::sphere-tracing}, a ray ($L$) is shot from the camera center ($\ve{c}$) and marches along the direction ($\Tilde{\ve{v}}$) with a step size that is equal to the queried signed distance value ($b$).
The ray marches iteratively until it hits or gets sufficiently close to the surface (\ie abs(SDF) $<$ threshold).
A more detailed algorithm can be found in Algorithm~\ref{alg::naive-sphere-tracing}.

\begin{algorithm}[tb]
    \caption{Naive sphere tracing algorithm for a camera ray $L: \ve{c} + d \Tilde{\ve{v}}$ over a signed distance fields $f: \mathbb{N}^3 \rightarrow \mathbb{R}$.}
    \label{alg::naive-sphere-tracing}
    \begin{algorithmic}[1]
    \State Initialize $n = 0$, $d^{(0)}=0$, $\ve{p}^{(0)}= \ve{c}$.
	\While {not converged}:
	\State Take the corresponding SDF value $b^{(n)}=f(\ve{p}^{(n)})$ of the location $\ve{p}^{(n)}$ and make update: $d^{(n+1)} = d^{(n)} + b^{(n)}$.
    \State $\ve{p}^{(n+1)}  = \ve{c} + d^{(n+1)} \Tilde{\ve{v}}$, $n = n + 1$.
    \State Check convergence.
    \EndWhile
    \end{algorithmic}
\end{algorithm}

\subsection{Efficient Forward Propagation}
Directly applying sphere tracing to an implicit SDF function represented by a neural network is prohibitively computational expensive, because each query of $f$ requires a forward pass of a neural network with considerable capacity.
Naive parallelization is not sufficient since essentially millions of network queries are required for a single rendering with VGA resolution ($640 \times 480$).
Therefore, we need to cut off unnecessary marching steps and safely speed up the marching process.

\noindent
\textbf{Initialization.}
Because all the 3D shapes represented by DeepSDF are bounded within the unit sphere, we initialize $\ve{p}^{(0)}$ to be the intersection between the camera ray and the unit sphere for each pixel.
Pixels with the camera rays that do not intersect with the unit sphere are set as background (\ie, infinite depth).

\noindent
\textbf{Coarse-to-fine Strategy.}
At the beginning of sphere tracing, rays for different pixels are fairly close to each other, which indicates that they will likely march in a similar way.
To leverage this nice property, we propose a coarse-to-fine sphere tracing strategy 
as shown in \figref{fig::lifetime}~(a).
We start the sphere tracing from an image with $\frac{1}{4}$ of its original resolution, and split each ray into four after every three marching steps, which is equivalent to doubling the resolution.
After six steps, each pixel in the full resolution has a corresponding ray, which keeps marching until convergence.

\noindent
\textbf{Aggressive Marching.}
After the ray marching begins, we apply an aggressive strategy (\figref{fig::lifetime}~(b)) to speed up the marching process by updating the ray with $\alpha$ times of the queried signed distance value, where $\alpha=1.5$ in our implementation. This aggressive sampling has several benefits. First, it makes the ray march faster towards the surface, especially when it is far from surface. Second, it accelerates the convergence for the ill-posed condition, where the angle between the surface normal and the ray direction is small. 
Third, the ray can pass through the surface such that space in the back (\ie, SDF $<$ 0) could be sampled. This is crucially important to apply supervision on both sides of the surface during optimization. 

\noindent
\textbf{Dynamic Synchronized Inference.}
A naive parallelization for speeding up sphere tracing is to batch rays together and synchronously update the front end positions.
However, depending on the 3D shape, some rays may converge earlier than others, thus leading to wasted computation.
We maintain a dynamic unfinished mask indicating which rays require further marching to prevent unnecessary computation.

\noindent
\textbf{Convergence Criteria.}
Even with aggressive marching, the ray movement can be extremely slow when close to the surface since $f$ is close to zero.
We define a convergence criteria to stop the marching when the accuracy is sufficiently good and the gain is marginal (\figref{fig::lifetime}(c)).
To fully maintain details supported by the 2D rendering resolution, it is sufficiently safe to stop when the sampled signed distance value does not confuse one pixel with its neighbors.
For an object with a smallest distance of 
100$mm$ captured by a camera with 60$mm$ focal length, 32$mm$ sensor width, and a resolution of $512\times 512$, the approximate minimal distance between the corresponding 3D points of two neighboring pixels is $10^{-4}m$ ($0.1mm$). 
In practice, we set the convergence threshold $\epsilon$ as $5 \times 10^{-5}$ for most of our experiments.

\subsection{Rendering 2D Observations}

After all rays converge, we can compute the distance along each ray as the following:
\vspace{-0.5em}
\begin{equation}
    d  = \alpha \sum_{n = 0}^{N-1} f(\ve{p}^{(n)}) + (1-\alpha)f(\ve{p}^{(N-1)}) = d' + e,
    \label{eq::ours-sphere-equation}
    \vspace{-0.5em}
\end{equation}
where $e=(1-\alpha)f(\ve{p}^{(N-1)})$ is the residual term on the last query. In the following part we will show how this computed ray distance is converted into 2D observations.

\noindent
\textbf{Depth and Surface Normal.} Suppose that we find the 3D surface point $\ve{p} = \ve{c} + d \Tilde{\ve{v}}$ for a pixel $(x, y)$ in the image, we can directly get the depth for each pixel as the following:
\begin{equation}
   z_c = \frac{d}{\sqrt{\Tilde{x}^2 + \Tilde{y}^2 + 1}}, 
\end{equation}
where $(\Tilde{x}, \Tilde{y}, 1)^\top = K^{-1}(x, y, 1)^\top$  is the normalized homogeneous  coordinate.

The surface normal of the point  $\ve{p}(x, y, z)$ can be computed as the normalized gradient of the function $f$.
Since $f$ is an implicit signed distance function, we take the approximation of the gradient by sampling neighboring locations:
\begin{equation}
\ve{n} = \frac{1}{2\delta} \begin{bmatrix} f(x+\delta, y, z) - f(x-\delta, y, z) \\ 
 f(x, y+\delta, z) - f(x, y-\delta, z) \\
  f(x, y, z+\delta) - f(x, y, z-\delta)
 \end{bmatrix}, \Tilde{\ve{n}} = \frac{\ve{n}}{|\ve{n}|}.
 \vspace{-0.5em}
\end{equation}

\noindent
\textbf{Silhouette.}
The silhouette is a commonly used supervision for 3D shape prediction.
To make the rendering of silhouettes differentiable,
we get the minimum absolute signed distance value for each pixel along its ray and subtract it by the convergence threshold $\epsilon$.
This produces a tight approximation of the silhouette, where pixels with positive values belong to the background, and vice versa.
Note that directly checking if ray marching stops at infinity can also generate the silhouette but it is not differentiable.

\noindent
\textbf{Color and Semantics.}
Recently, it has been shown that texture can also be represented as an implicit function parameterized with a neural network \cite{oechsle2019texture}. Not only color, other spatially varying properties, like semantics, material, etc, can all be potentially learned by implicit functions.
These information can be rendered jointly with the implicit SDF to produce corresponding 2D observations, and some examples are depicted in \figref{fig::texture_render}.

\subsection{Approximated Gradient Back-Propagation}
\label{sec::approximated_gradient}
DeepSDF \cite{park2019deepsdf} uses the conditional implicit function to represent a 3D shape as $f_{\theta}(\ve{p}, \ve{z})$, where $\theta$ is the network parameters, and $\ve{z}$ is the latent code representing a certain shape. 
As a result, each queried point $\ve{p}$ in the sphere tracing process is determined by $\theta$ and the shape code $\ve{z}$, which requires to unroll the network for multiple times and costs huge memory for back-propagation with respect to $\ve{z}$:
\vspace{-0.5em}
\begin{equation}
\small
    \begin{aligned}
    \frac{\partial d'}{\partial \ve{z}}|_{\ve{z}_0} & = \alpha \sum_{i = 0}^{N-1} \frac{\partial f_{\theta}(\ve{p}^{(i)}(\ve{z}),  \ve{z})}{\partial \ve{z}}|_{\ve{z}_0} \\
       &= \alpha \sum_{i = 0}^{N-1} (\frac{\partial f_{\theta}(\ve{p}^{(i)}(\ve{z_0}),  \ve{z})}{\partial \ve{z}} + \frac{\partial f_{\theta}(\ve{p}^{(i)}(\ve{z}),  \ve{z_0})}{\partial \ve{p}^{(i)}(\ve{z})} \frac{\partial \ve{p}^{(i)}(\ve{z_0})}{\partial \ve{z}}).
        \label{eqn::gradient}
    \end{aligned}
\end{equation}

Practically, we ignore the gradients from the residual term $e$ in \equref{eq::ours-sphere-equation}. In order to make back-propagation feasible, we define a loss for $K$ samples with the minimum absolute SDF value on the ray to encourage more signals near the surface. For each sample, we calculate the gradient with only the first term in \equref{eqn::gradient} as the high-order gradients empirically have less impact on the optimization process.
In this way, our differentiable renderer is particularly useful to bridge the gap between this strong shape prior and some partial observations.
Given a certain observation, we can search for the code that minimizes the difference between the rendering from our network and the observation.
This allows a number of applications which will be introduced in the next section.

\begin{table}[tb]
\centering
\begin{tabular}{c|c|c|c|c}
\hline
Method & size & \#step & \#query & time \\
\hline
Naive sphere tracing & $512^2$ & 50 & N/A & N/A \\
+ practical grad. & $512^2$ & 50 & 6.06M & 1.6h \\
+ parallel & $512^2$ & 50 & 6.06M & 3.39s \\
+ dynamic & $512^2$ & 50 & 1.99M & 1.23s \\
+ aggressive & $512^2$ & 50 & 1.43M & 1.08s \\
+ coarse-to-fine & $512^2$ & 50 & \textbf{887K} & \textbf{0.99s}  \\
+ coarse-to-fine & $512^2$ & 100 & \textbf{898K} & \textbf{1.24s} \\
\hline
\end{tabular}
\centering
\vspace{5pt}
\caption{Ablation studies on the cost-efficient feedforward design of our method. The average time for each optimization step was tested on a single NVIDIA GTX-1080Ti over the architecture of DeepSDF \cite{park2019deepsdf}. Note that the number of initialized rays is quadratic to the image size, and the numbers are reported for the resolution of $512 \times 512$.}
\vspace{-5pt}
\label{tab::statistics}
\end{table}

\section{Experiments and Results}
In this section, we first verify the efficacy of our differentiable sphere tracing algorithm, and then show that 3D shape understanding can be achieved through geometry based reasoning by our method.

\subsection{Rendering Efficiency and Quality}

\begin{figure}[tb]
\centering
\setlength\tabcolsep{2.3pt} 
\scriptsize
\begin{tabular}{cccc}
parallel & + dynamic & + aggressive & + coarse-to-fine

\\
{\includegraphics[width=0.21\linewidth]{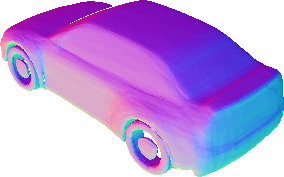}} &
{\includegraphics[width=0.21\linewidth]{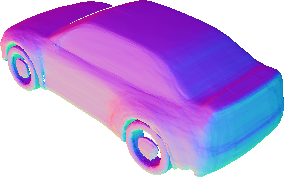}} &
{\includegraphics[width=0.21\linewidth]{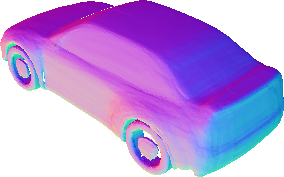}} &
{\includegraphics[width=0.21\linewidth]{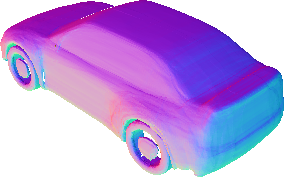}}
\\
\end{tabular}
\centering
\caption{The surface normal rendered with different speed up strategies turned on. Note that adding up these components does not deteriorate the rendering quality.}
\label{fig::validity_check}
\vspace{-5pt}
\end{figure}

\noindent
\textbf{Run-time Efficiency.}
In this section, we evaluate the run-time efficiency promoted by each design in our differentiable sphere tracing algorithm.
The number of queries and runtime for both forward and backward passes at a resolution of $512 \times 512$ on a single NVIDIA GTX-1080Ti are reported in  Tab.~\ref{tab::statistics}, and the corresponding rendered surface normal are shown in Fig.~\ref{fig::validity_check}.
We can see that the proposed back-propagation prunes the graph and reduces the memory usage significantly, making the rendering tractable with a standard graphics card. 
The dynamic synchronized inference, aggressive marching and coarse-to-fine strategy all speed up rendering. With all these designs, we can render an image with only 887K query steps within 0.99s when the maximum tracing step is set to 50. The number of query steps only increases slightly when the maximum step is set to 100, indicating that most of the pixels converge safely within 50 steps. Note that related works usually render at a much lower resolution \cite{sitzmann2019scene}.

\begin{figure}[tb]
\centering
\scriptsize
\setlength\tabcolsep{2.3pt} 
\begin{tabular}{cccccc}
\multicolumn{3}{c}{\includegraphics[width=0.49\linewidth]{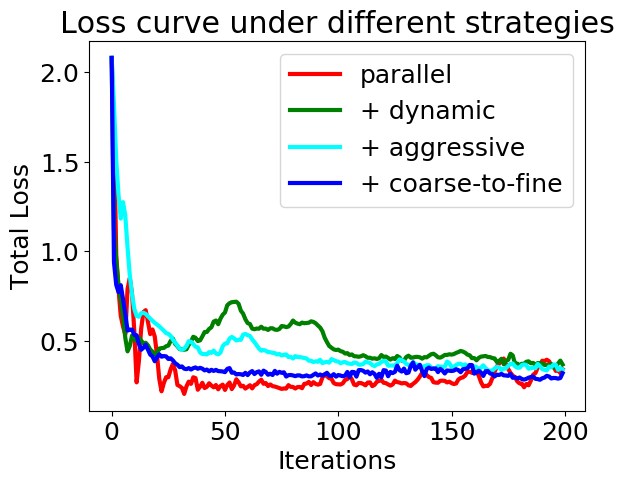}} &
\multicolumn{3}{c}{\includegraphics[width=0.49\linewidth]{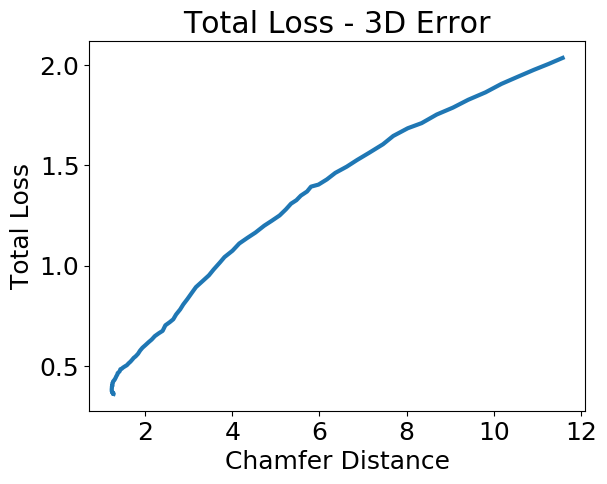}}
\end{tabular}
\centering
\caption{Loss curves for 3D prediction from partial depth. Our accelerated rendering does not impair the back-propagation. The loss on the depth image is tightly correlated with the Chamfer distance on 3D shapes, which indicates effective back-propagation.
}
\label{fig::2d_optimization}
\end{figure}

\begin{figure}[tb]
\centering
\scriptsize
\setlength\tabcolsep{2.3pt} 
\begin{tabular}{ccccc}
initial & \multicolumn{3}{c}{{\includegraphics[trim={5 30 5 10}, clip, width=0.60\linewidth]{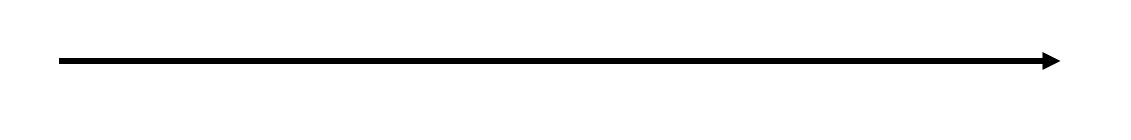}}} & optimized
\\
{\includegraphics[trim={0 0 0 0}, clip, width=0.18\linewidth]{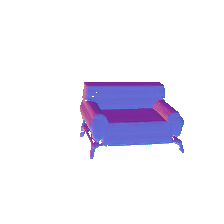}} & 
{\includegraphics[trim={0 0 0 0}, clip,width=0.18\linewidth]{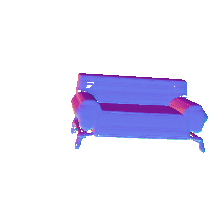}} & 
{\includegraphics[trim={0 0 0 0}, clip,width=0.18\linewidth]{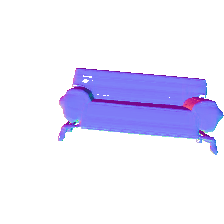}} & 
{\includegraphics[trim={0 0 0 0}, clip,width=0.18\linewidth]{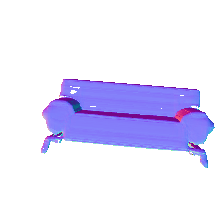}} & 
{\includegraphics[trim={0 0 0 0}, clip,width=0.18\linewidth]{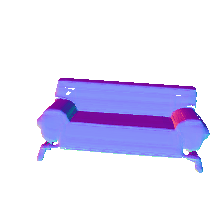}}
\vspace{-1.2em}
\\
{\includegraphics[width=0.18\linewidth]{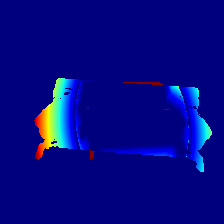}} &
{\includegraphics[width=0.18\linewidth]{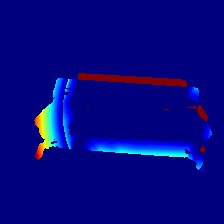}} &
{\includegraphics[width=0.18\linewidth]{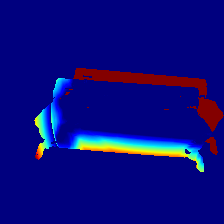}} &
{\includegraphics[width=0.18\linewidth]{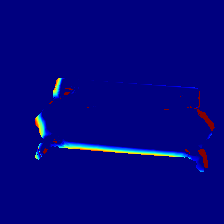}} &
{\includegraphics[width=0.18\linewidth]{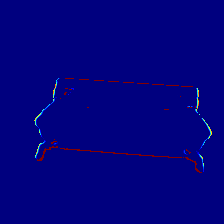}}
\end{tabular}
\centering
\caption{Illustration of the optimization process over the camera extrinsic parameters. Our differentiable renderer is able to propagate the error from the image plane to the camera. Top row: rendered surface normal. Bottom row: error map on the silhouette.}
\label{fig::demo_camera}
\vspace{-1.0em}
\end{figure}

\noindent
\textbf{Back-Propagation Effectiveness.}
We conduct sanity checks to verify the effectiveness of the back-propagation with our approximated gradient.
We take a pre-trained DeepSDF \cite{park2019deepsdf} model and run geometry based optimization to recover the 3D shape and camera extrinsics separately using our differentiable renderer.
We first assume camera pose is known and optimize the latent code for 3D shape w.r.t the given ground truth depth map, surface normal and silhouette.
As can be seen in Fig.~\ref{fig::2d_optimization} (left), the loss drops quickly, and using acceleration strategies does not hurt the optimization.
Fig.~\ref{fig::2d_optimization} (right) shows the total loss on the 2D image plane is highly correlated with the Chamfer distance on the predicted 3D shape, indicating that the gradients originated from the 2D observation are successfully back-propagated to the shape.
We then assume a known shape (fixed latent code) and optimize the camera pose using a depth image and a binary silhouette.
Fig.~\ref{fig::demo_camera} shows that a random initial camera pose can be effectively optimized toward the ground truth pose by minimizing the gradients on 2D observations. 

\begin{figure}[tb]
\centering
\small
\setlength\tabcolsep{2.3pt} 
\begin{tabular}{cccc}
$\epsilon =5\times10^{-2}$ & $\epsilon=5\times 10^{-4}$ & $\epsilon=5\times 10^{-6}$ & $\epsilon=5\times 10^{-8}$
\\
{\includegraphics[trim={0 0 100 120}, clip, width=0.23\linewidth]{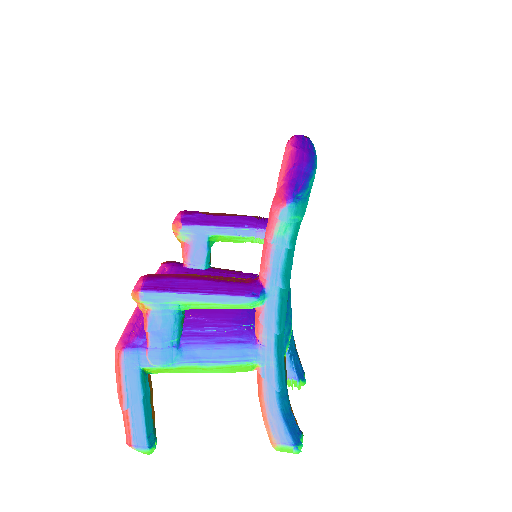}} &
{\includegraphics[trim={0 0 100 120}, clip, width=0.23\linewidth]{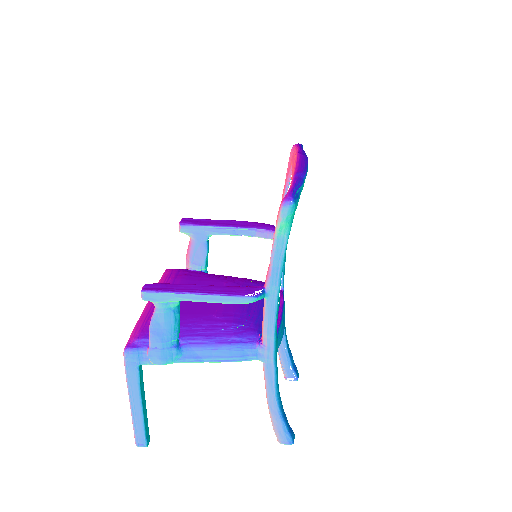}} &
{\includegraphics[trim={0 0 100 120}, clip, width=0.23\linewidth]{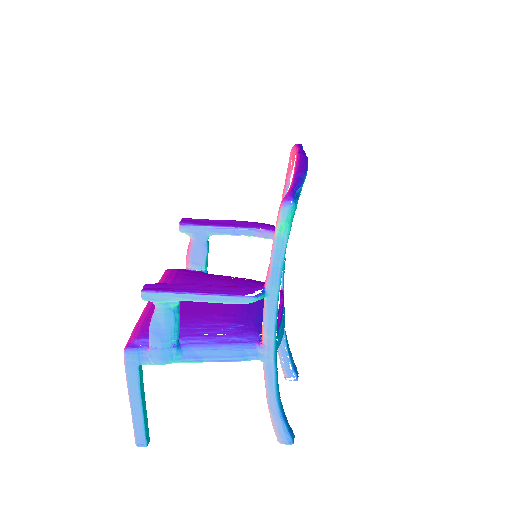}} &
{\includegraphics[trim={0 0 100 120}, clip, width=0.23\linewidth]{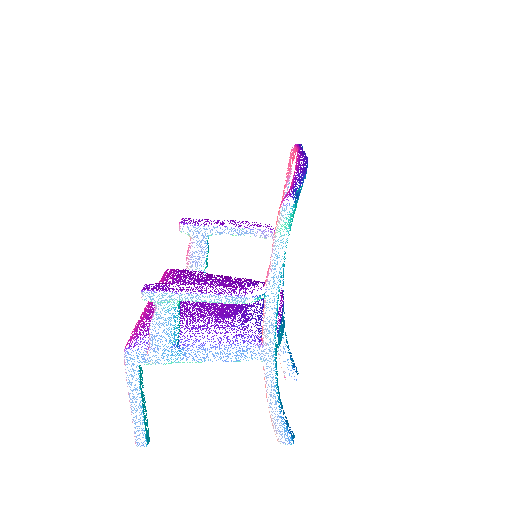}}
\\
\end{tabular}
\vspace{-10pt}
\caption{Effects on choices of different convergence thresholds. Under the same marching step, a very large threshold can incur dilation around boundaries while a small threshold may lead to erosion. We pick $5\times10^{-5}$ for all of our experiments.}
\label{fig::threshold}
\end{figure}

\noindent
\textbf{Convergence Criteria.}
The convergence criteria, \ie, the threshold on signed distance to stop the ray tracing, has a direct impact on the rendering quality.
Fig.~\ref{fig::threshold} shows the rendering result under different thresholds.
As can be seen, rendering with a large threshold will dilate the shape, which lost boundary details. Using a small threshold, on the other hand, may produces incomplete geometry. 
This parameter can be tuned according to applications, but in practice we found our threshold is effective in producing complete shape with details up to the image resolution.

\begin{figure}[tb]
\centering
\scriptsize
\setlength\tabcolsep{2.3pt} 
\begin{tabular}{cccc}
LR texture & 32x HR texture &  HR Relighting & HR 2nd View
\\
{\includegraphics[width=0.23\linewidth]{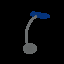}} &
{\includegraphics[width=0.23\linewidth]{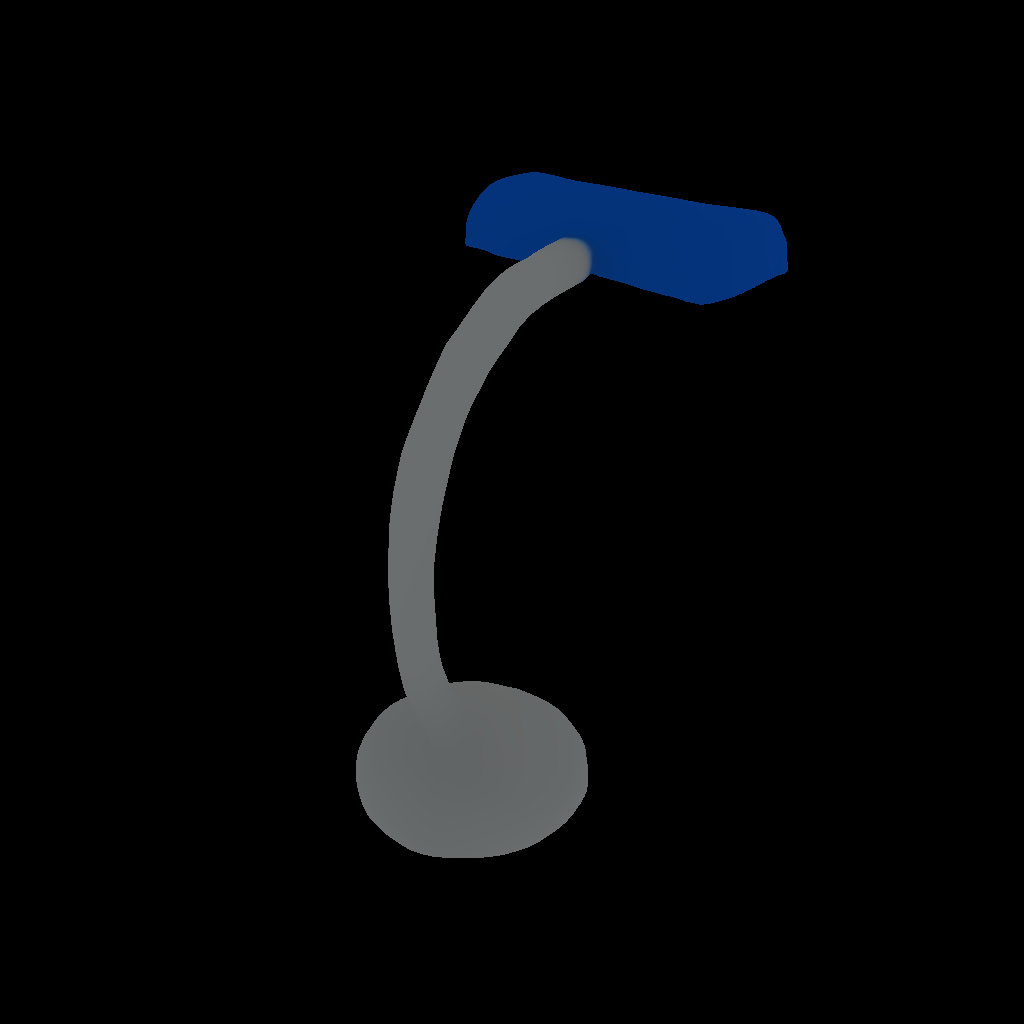}} &
{\includegraphics[width=0.23\linewidth]{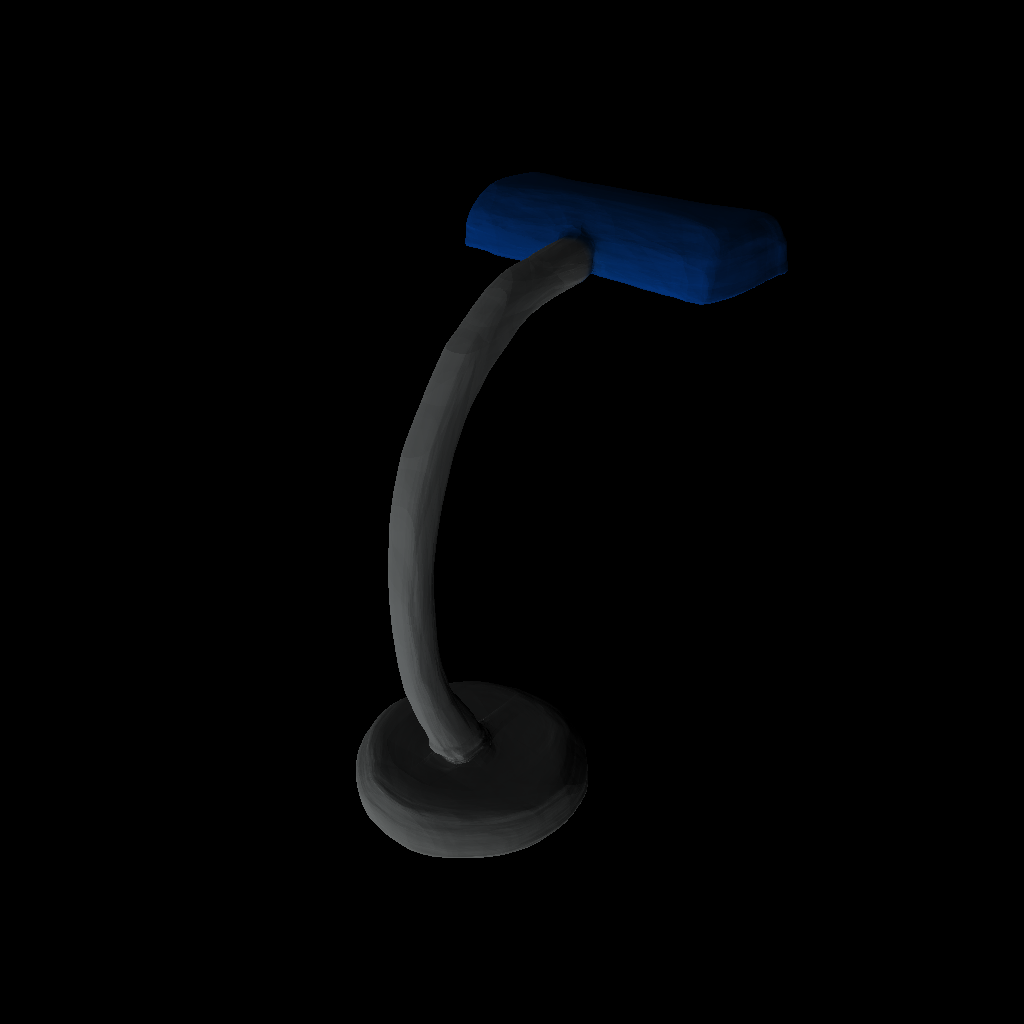}} &
{\includegraphics[width=0.23\linewidth]{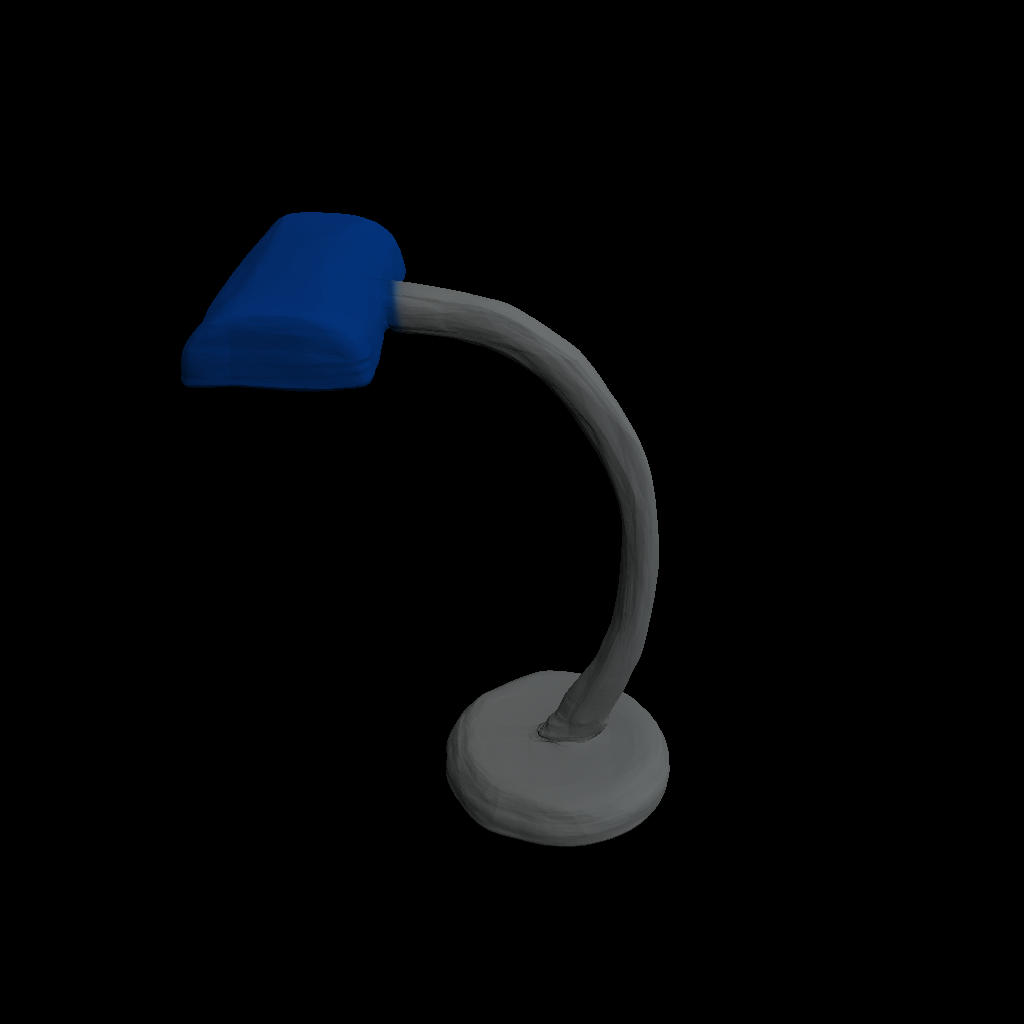}}\\
{\includegraphics[width=0.23\linewidth]{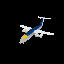}} &
{\includegraphics[width=0.23\linewidth]{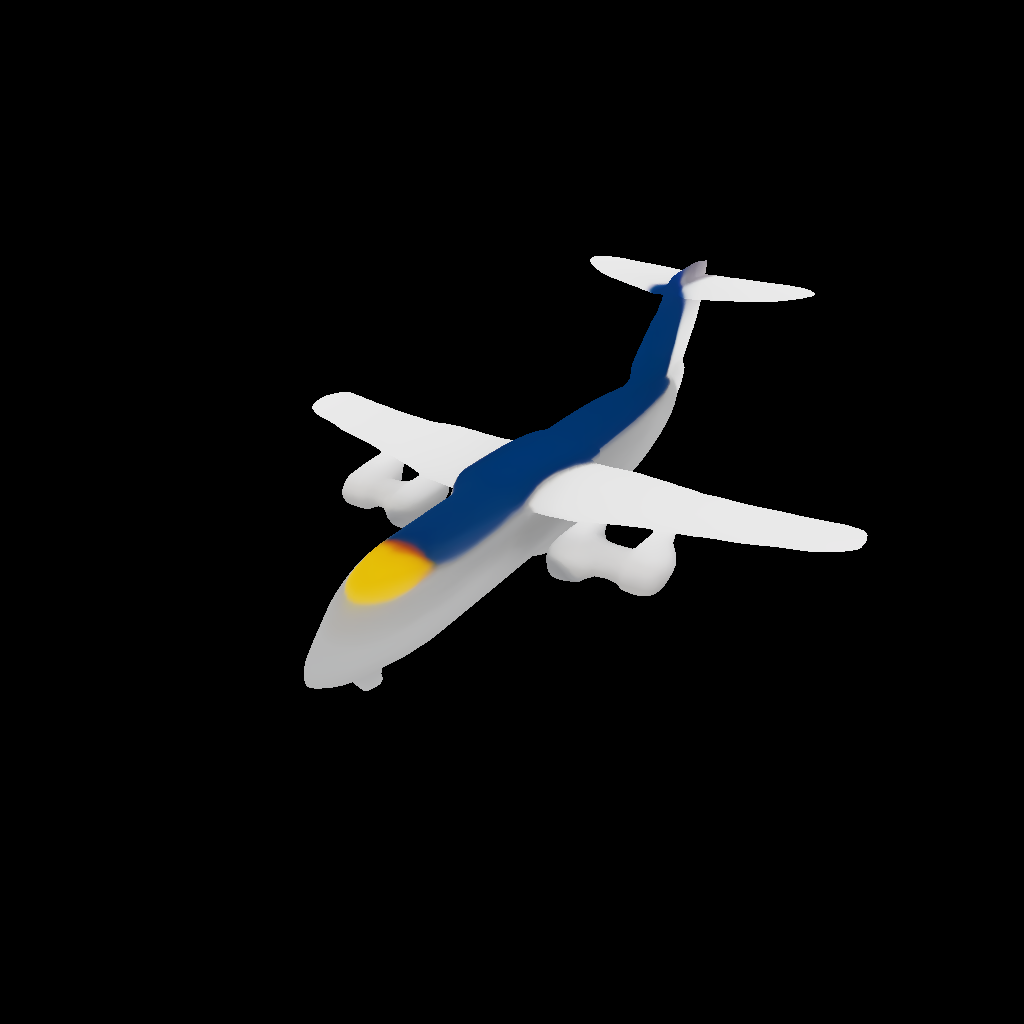}} &
{\includegraphics[width=0.23\linewidth]{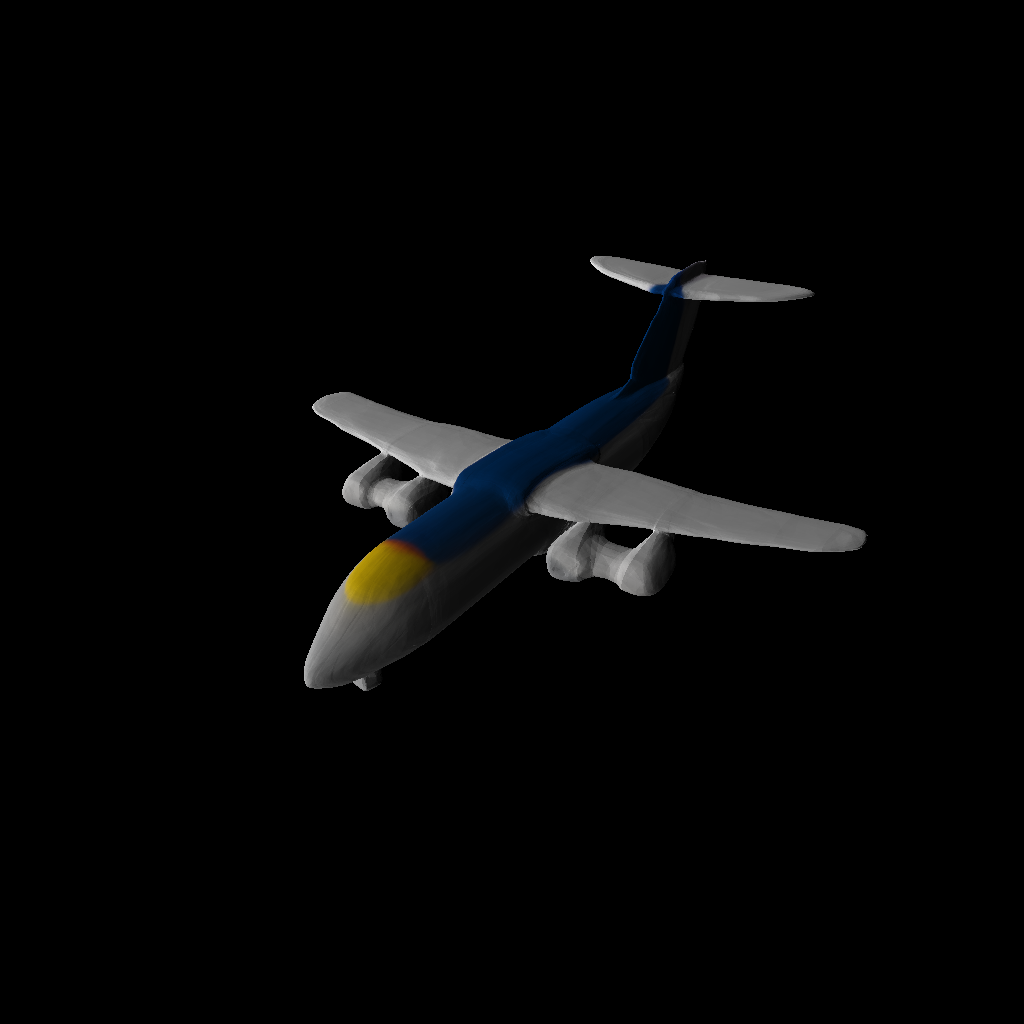}} &
{\includegraphics[width=0.23\linewidth]{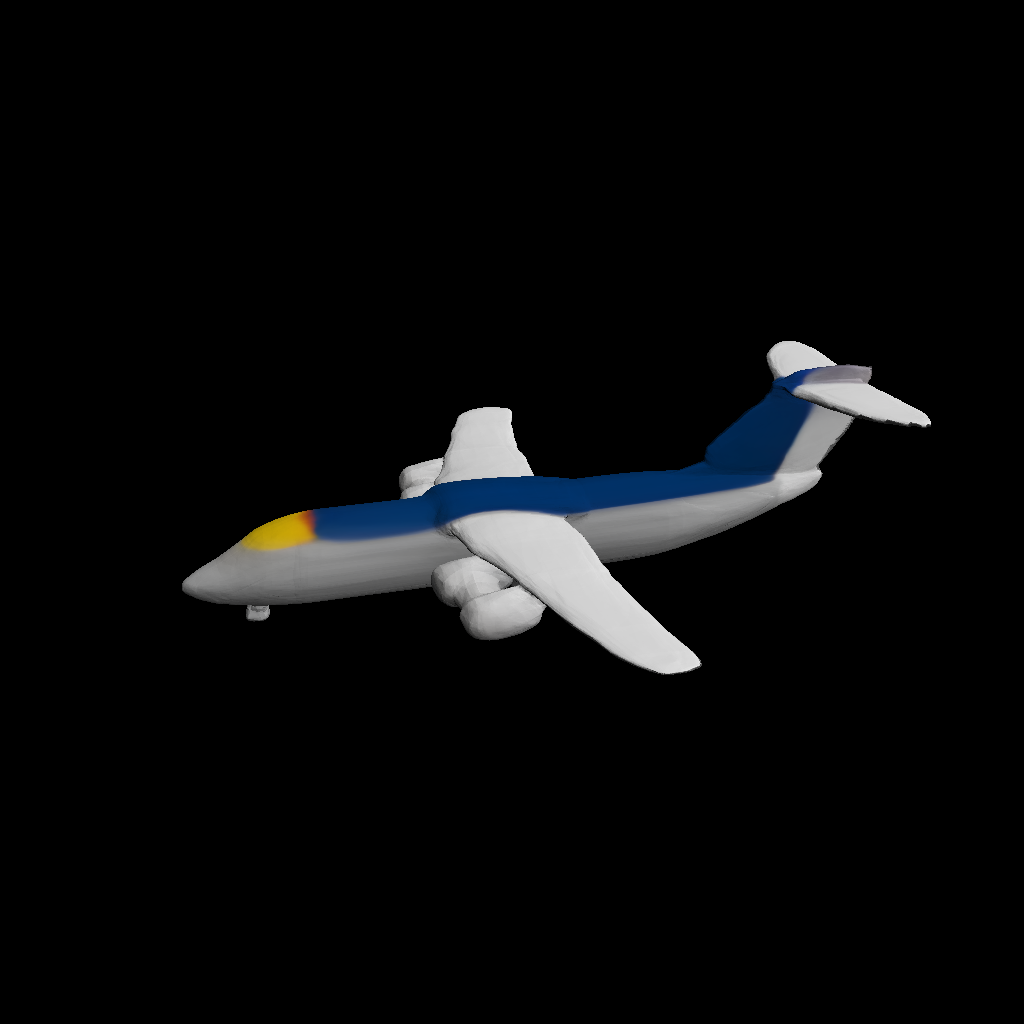}}
\end{tabular}
\centering
\caption{Our method can render information encoded in the implict function other than depth. With a pre-trained network encoding textured meshes, we can render high resolution color images under various resolution, camera viewpoints, and illumination.
}
\label{fig::texture_render}
\end{figure}

\noindent
\textbf{Rendering Other Properties.}
Not only the signed distance function for 3D shape, implicit functions can also encode other spatially variant information.
As an example, we train a network to predict both signed distance and color for each 3D location, and this grants us the capability of rendering color images.
In Fig.~\ref{fig::texture_render}, we show that with a 512-dim latent code learned from textured meshes as the ground truth, color images can be rendered in arbitrary resolution, camera viewpoints, and illumination.
Note that the latent code size is significantly smaller than the mesh (vertices+triangles+texture map), and thus can be potentially used for model compression.
Other per-vertex properties, such as semantic segmentation and material, can also be rendered in the same differentiable way.

\subsection{3D Shape Prediction}
Our differentiable implicit SDF renderer builds up the connection between 3D shape and 2D observations and enables geometry based reasoning.
In this section, we show results of 3D shape prediction from a single depth image, or multi-view color images using DeepSDF as the shape generator.
On a high-level, we take a pre-trained DeepSDF and fixed the decoder parameters.
When given 2D observations, we define proper loss functions and propagate the gradient back to the latent code, as introduced in Section~\ref{sec::approximated_gradient}, to generate 3D shape.
This method does not require any additional training and only need to run optimization at test time, which is intuitively less vulnerable to overfitting or domain gap issues in pure learning based approach.
In this section, we specifically focus on evaluating the generalization capability while maintaining high shape quality.

\subsubsection{3D Shape Prediction from Single Depth Image}

With the development of commodity range sensors, the dense or sparse depth images can be easily acquired, and several methods have been proposed to solve the problem of 3D shape prediction from a single depth image. 
DeepSDF \cite{park2019deepsdf} has shown state-of-the-art performance for this task,
however requires an offline pre-processing to lift the input 2D depth map into 3D space in order to sample the SDF values with the assistance of the surface normal.
Our differentiable render makes 3D shape prediction from a depth image more convenient by directly rendering the depth image given a latent code and comparing it with the given depth. 
Moreover, with the silhouette 
calculated from the depth map or provided from the rendering, our renderer can also leverage it as an additional supervision.
Formally, we obtain the complete 3D shape by solving the following optimization:
\vspace{-0.2em}
\begin{equation}
    \argmin_{\ve{z}} \mathcal{L}_d(\mathcal{R}_d(f(\ve{z})), I_d) + \mathcal{L}_s(\mathcal{R}_s(f(\ve{z})), I_s),
    \vspace{-0.2em}
\end{equation}
where $f(\ve{z})$ is the pre-trained neural network encoding shape priors, $\mathcal{R}_d$ and $\mathcal{R}_s$ represent the rendering function for the depth and silhouette respectively, $\mathcal{L}_d$ is the $L_1$ loss of depth observation, and $\mathcal{L}_s$ is the loss defined based on the differentiably rendered silhouette. In our experiment, the initial latent shape $\ve{z}_0$ is chosen as the mean shape.

We test our method and DeepSDF \cite{park2019deepsdf} on 200 models on plane, sofa and table category respectively from ShapeNet Core \cite{chang2015shapenet}. Specifically, for each model, we use the first camera in the dataset of Choy \etal \cite{choy20163d} to generate dense depth images for testing. 
The comparison between DeepSDF and our method is listed in \tabref{tab::depth_completion}. We can see that our method with only depth supervision performs better than DeepSDF \cite{park2019deepsdf} when dense depth image is given. 
This is probably because that DeepSDF samples the 3D space with pre-defined rule (at fixed distances along the normal direction), which may not necessarily sample correct location especially near object boundary or thin structures.
In contrast, our differentiable sphere tracing algorithm samples the space adaptively with the current estimation of shape. 

\begin{table}[tb]
\setlength\tabcolsep{4pt}
    \centering
 	\small
    \begin{tabular}{|l|c|c|c|c|c|c|}
    \hline
     & dense & 50\% & 10\% & 100pts & 50pts & 20pts \\
    \hline
    \multicolumn{7}{|l|}{sofa} \\
    \hline
    DeepSDF & 5.37 & 5.56 & 5.50 & 5.93 & 6.03 & 7.63 \\
    Ours & \textbf{4.12} & 5.75 & 5.49 & 5.72 & 5.57 & 6.95 \\
    Ours (mask) & \textbf{4.12} & \textbf{3.98} & \textbf{4.31} & \textbf{3.98} & \textbf{4.30} & \textbf{4.94} \\
    \hline
    \multicolumn{7}{|l|}{plane} \\
    \hline
    DeepSDF & 3.71 & 3.73 & 4.29 & 4.44 & 4.40 & 5.39 \\
    Ours & \textbf{2.18} & 4.08 & 4.81 & 4.44 & 4.51 & 5.30 \\
    Ours (mask) & \textbf{2.18} & \textbf{2.08} & \textbf{2.62} & \textbf{2.26} & \textbf{2.55} & \textbf{3.60} \\
    \hline
    \multicolumn{7}{|l|}{table} \\
    \hline
    DeepSDF & 12.93 & 12.78 & 11.67 & 12.87 & 13.76 & 15.77 \\
    Ours & \textbf{5.37} & 12.05 & 11.42 & 11.70 & 13.76 & 15.83 \\
    Ours (mask) & \textbf{5.37} & \textbf{5.15} & \textbf{5.16} & \textbf{5.26} & \textbf{6.33} & \textbf{7.62} \\
    \hline
    \end{tabular}
    \vspace{5pt}
    \caption{Quantitative comparison between our geometric optimization with DeepSDF \cite{park2019deepsdf} for shape completion over partial dense and sparse depth observation on ShapeNet dataset \cite{chang2015shapenet}. We report the median Chamfer Distance on the first 200 instances of the dataset of \cite{choy20163d}. We give DeepSDF \cite{park2019deepsdf} the groundtruth normal otherwise they could not be applied on the sparse depth.} 
    \label{tab::depth_completion}
    \vspace{-5pt}
\end{table}

\vspace{2pt}
\noindent
\textbf{Robustness against sparsity.} The depth from laser scanners can be very sparse, so we also study the robustness of our method and DeepSDF against sparse depth. 
The results are shown in \tabref{tab::depth_completion}. 
Specifically, we randomly sample different percentages or fixed numbers of points from the original dense depth for testing. 
To make a competitive baseline, we provide DeepSDF ground truth normals to sample SDF, since it cannot be reliably estimated from sparse depth.
From the table, we can see that even with very sparse depth observations, our method still recovers accurate shapes and gets consistently better performance than DeepSDF with additional normal information.
When the silhouette is available, our method achieves significantly better performance and robustness against the sparsity, indicating that our rendering method can back-propagate gradients effectively from the silhouette loss.

\subsubsection{3D Shape Prediction from Multiple Images}
\begin{figure}[!tb]
\centering
\scriptsize
\setlength\tabcolsep{2.3pt} 
\begin{tabular}{ccccc}
Video sequence & \multicolumn{4}{c}{Optimization process}
\\
{\includegraphics[width=0.18\linewidth]{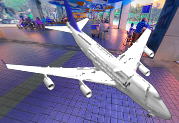}} &
{\includegraphics[width=0.18\linewidth]{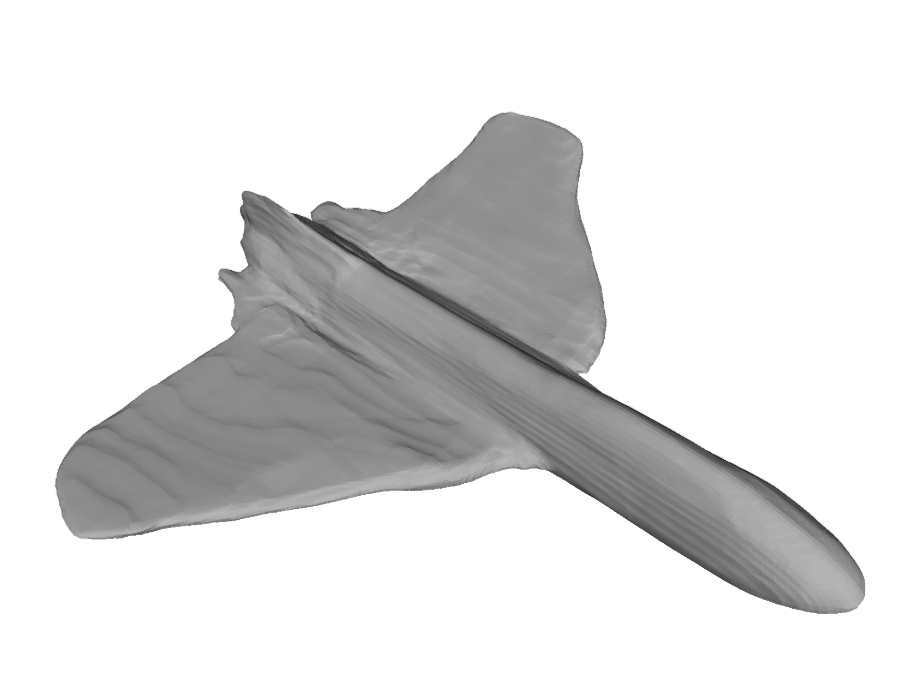}} &
{\includegraphics[width=0.18\linewidth]{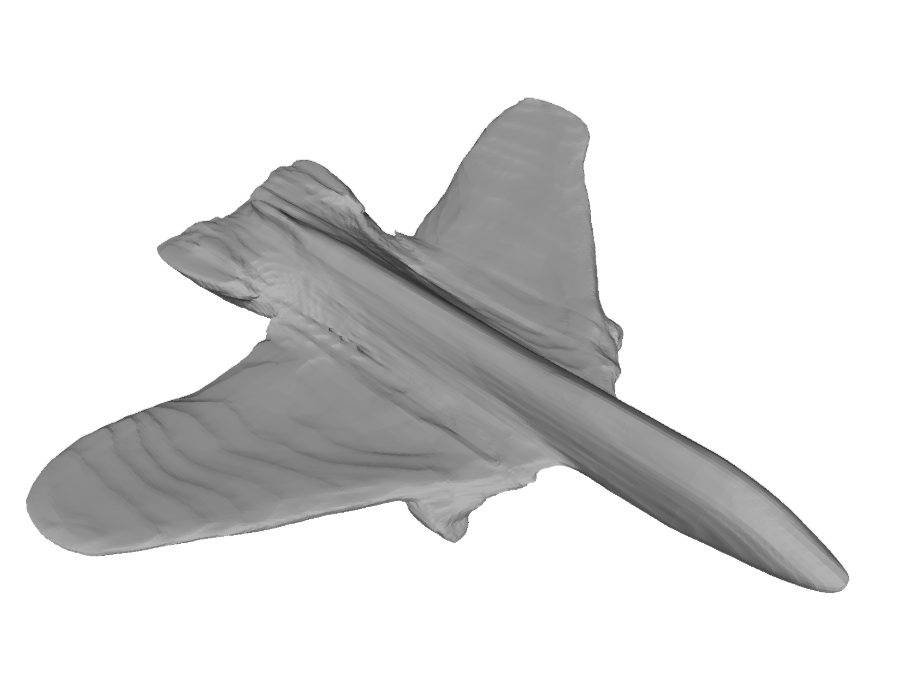}} &
{\includegraphics[width=0.18\linewidth]{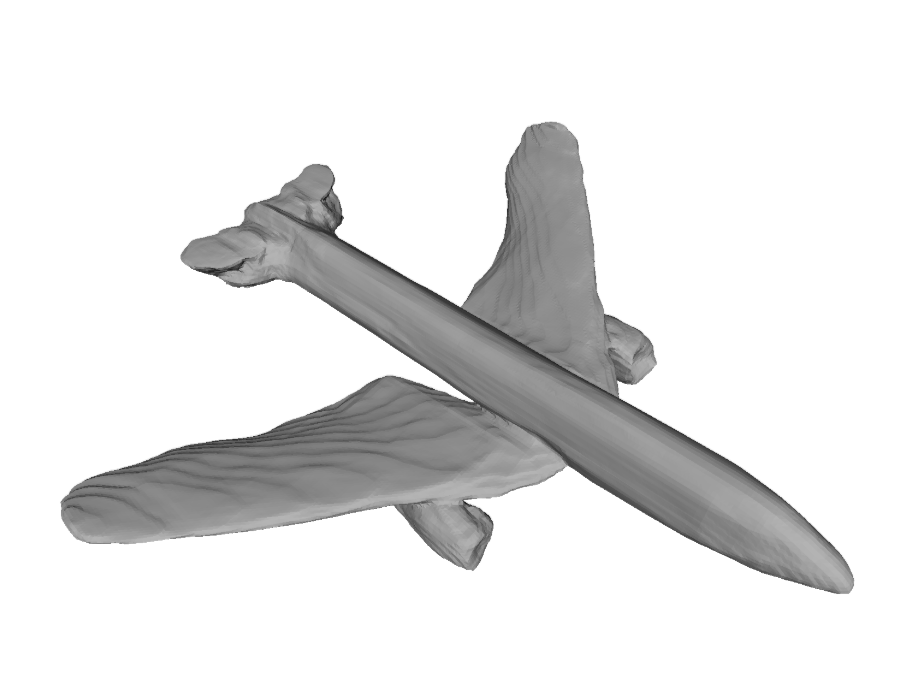}} &
{\includegraphics[width=0.18\linewidth]{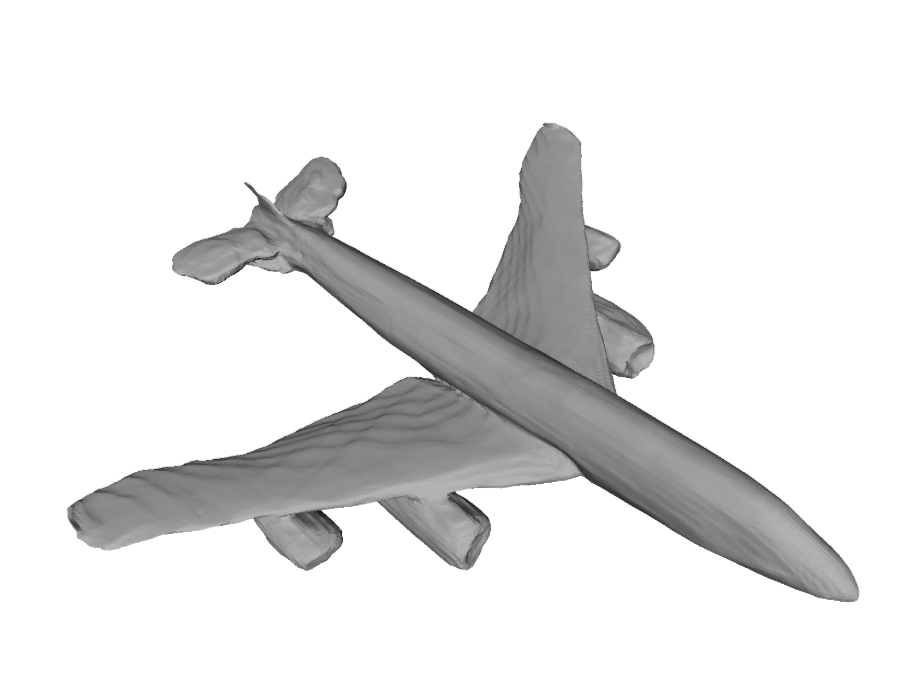}}
\end{tabular}
\centering
\caption{Illustration of the optimization process under multi-view setup. Our differentiable renderer is able to successfully recover 3D geometry from a random code with only the photometric loss.}
\label{fig::mv_dynamic}
\vspace{-3pt}
\end{figure}

Our differentiable renderer can also enable geometry based reasoning for shape prediction from multi-view color images by leveraging cross-view photometric consistency.

Specifically, we first initialize the latent code with a random vector and render a depth image for each of the input views.
We then warp each color image to other input views using the rendered depth image and the known camera pose.
The difference between the warped and input images are then defined as the photometric loss, and the shape can be predicted by minimizing this loss.
To sum up, the optimization problem is formulated as follows,

\begin{equation}
\vspace{-0.3em}
    \argmin_{\ve{z}} \sum_{i=0}^{N-1} \sum_{j \in \mathcal{N}_i} \Vert I_i - I_{j\rightarrow i}(\mathcal{R}^i_d(f(\ve{z})) \Vert,
\end{equation}
where  $\mathcal{R}^i_d$ represents the rendered depth image at view $i$, $\mathcal{N}_i$ are the neighboring images of $I_i$,  and $I_{j\rightarrow i}$ is the warped image from view $j$ to view $i$ using the rendered depth. Note that no mask is required under the multi-view setup.
\figref{fig::mv_dynamic} shows an example of the optimization process of our method. As can be seen, the shape is gradually improved while the loss is being optimized.

\begin{table}[tb]
\setlength\tabcolsep{4pt}
\begin{tabular}{l|cc}\hline
Method                    & car   & plane  \\\hline
PMO (original)            & 0.661 & 1.129  \\
PMO (rand init)           & 1.187 & 6.124  \\\hline
\textbf{Ours} (rand init) & 0.919 & 1.595  \\\hline
\end{tabular}
\centering
\vspace{5pt}
\caption{Quantitative results on 3D shape prediction from multi-view images under the metric of Chamfer Distance (only in the direction of gt$\rightarrow$pred for fair comparison).  
We randomly picked 50 instances from the PMO test set to perform the evaluation. 10000 points are sampled from meshes for evaluation.}
\label{tab::mv_comp_pmo}
\end{table}

We take PMO~\cite{lin2019photometric} as a competitive baseline, since they also perform deep learning based geometric reasoning via optimization over a pre-trained decoder, but use the triangular mesh representation. 
Their model first predicts an initial mesh from a selected input view and improve the quality via cross-view photo-consistency.
Both the synthetic and real datasets provided in~\cite{lin2019photometric} are used for evaluation.

In \tabref{tab::mv_comp_pmo}, we show quantitative comparison to PMO on their synthetic test set.
It can be seen that our method achieves comparable results with PMO~\cite{lin2019photometric} from only random initializations. 
Note that while PMO uses both the encoder and decoder trained on the PMO training set, our DeepSDF decoder was neither trained nor finetuned on it.
Besides, if the shape code for PMO, instead of being predicted from their trained image encoder, is also initialized randomly, their performance decreases dramatically, which indicates that with our rendering method, our geometric reasoning becomes more effective. Our method can be further improved with good initialization.

\begin{figure}[!tb]
\centering
\scriptsize
\setlength\tabcolsep{2.3pt} 
\begin{tabular}{cc}
{\includegraphics[width=0.43\linewidth]{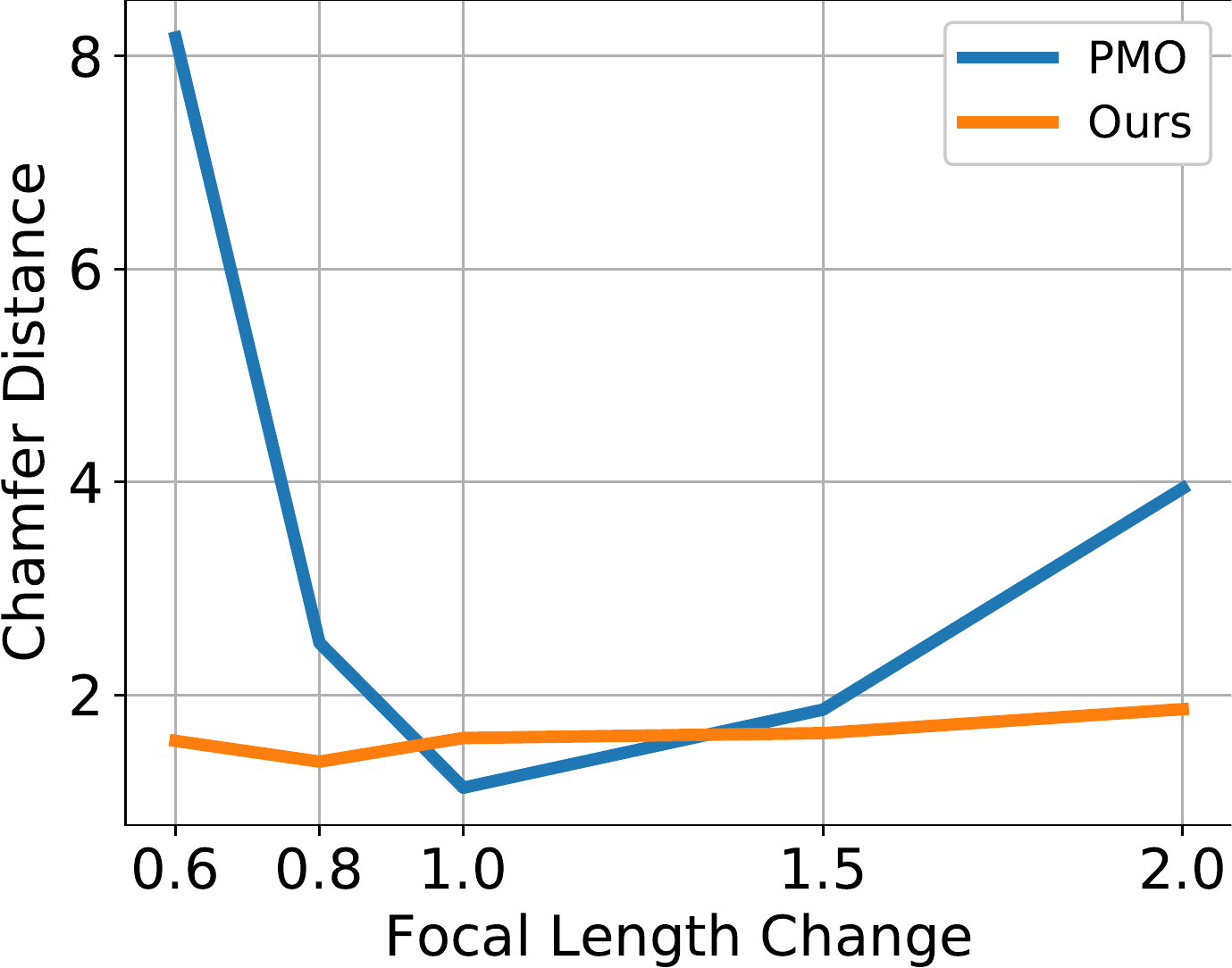}} &
{\includegraphics[width=0.43\linewidth]{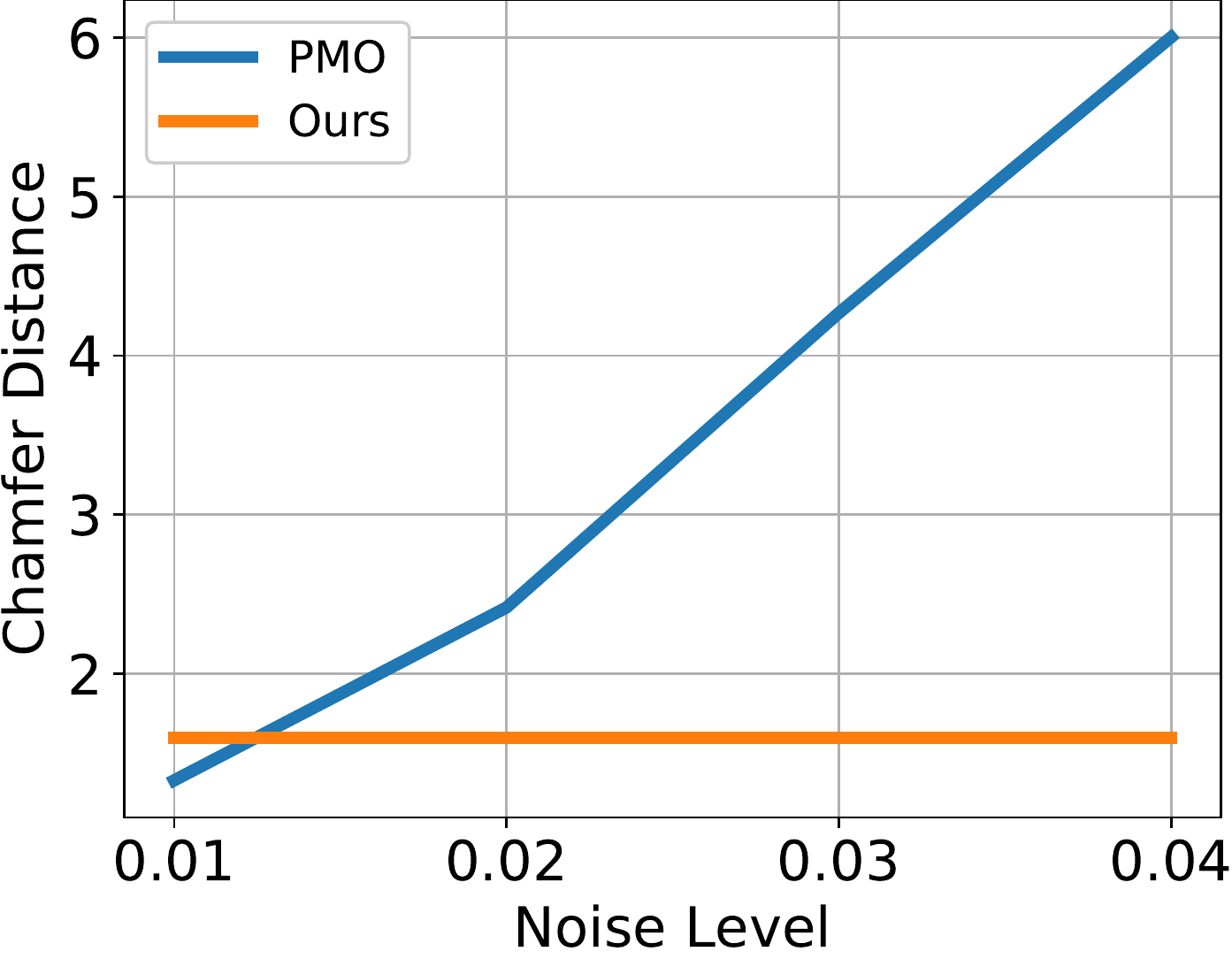}}
\\
(a) & (b)
\end{tabular}
\vspace{2pt}
\centering
\caption{Robustness of geometric reasoning via multi-view photometric optimization. (a) Performance w.r.t changes on camera focal length. (b) Performance w.r.t noise in the initialization code.
Our model is robust against focal length change and not affected by noise in the latent code since we start from random initialization.
In contrast, PMO is very sensitive to both factors, and the performance drops significantly when the testing images are different from the training set.
}
\label{fig::mv_ablation}
\vspace{-3pt}
\end{figure}

\begin{figure}[!htb]
\centering
\scriptsize
\setlength\tabcolsep{2.3pt} 
\begin{tabular}{cccc}
Video sequence & PMO (rand init)& PMO & \textbf{Ours}
\\
{\includegraphics[trim={0 0 0 10}, clip, width=0.25\linewidth]{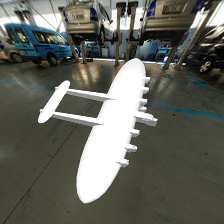}} &
{\includegraphics[width=0.22\linewidth]{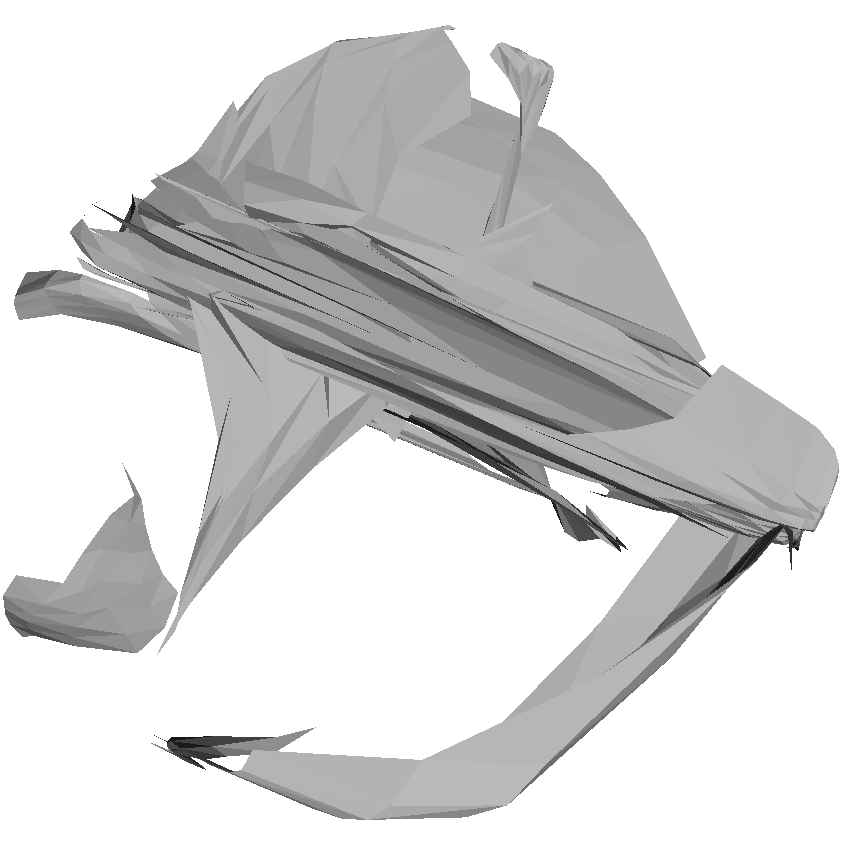}} &
{\includegraphics[width=0.22\linewidth]{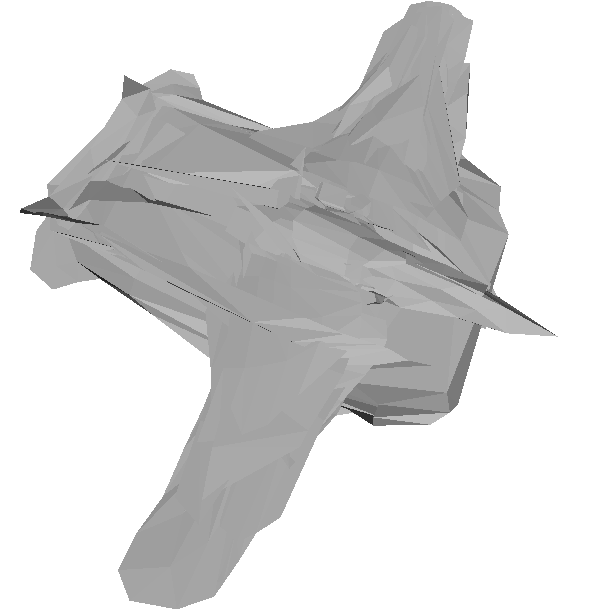}} &
{\includegraphics[width=0.22\linewidth]{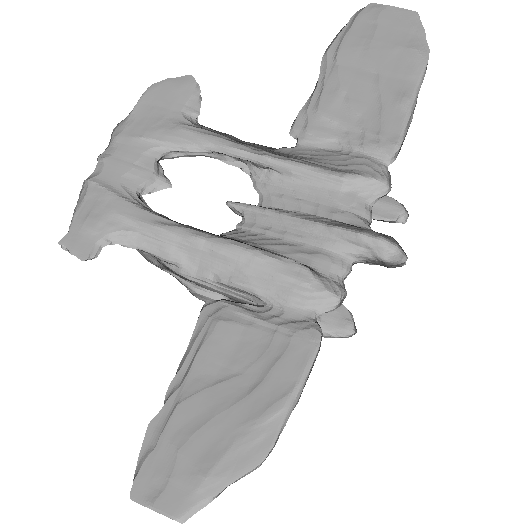}}
\\
{\includegraphics[trim={0 0 0 10}, clip, width=0.25\linewidth]{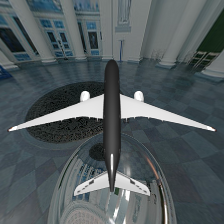}} &
{\includegraphics[width=0.22\linewidth]{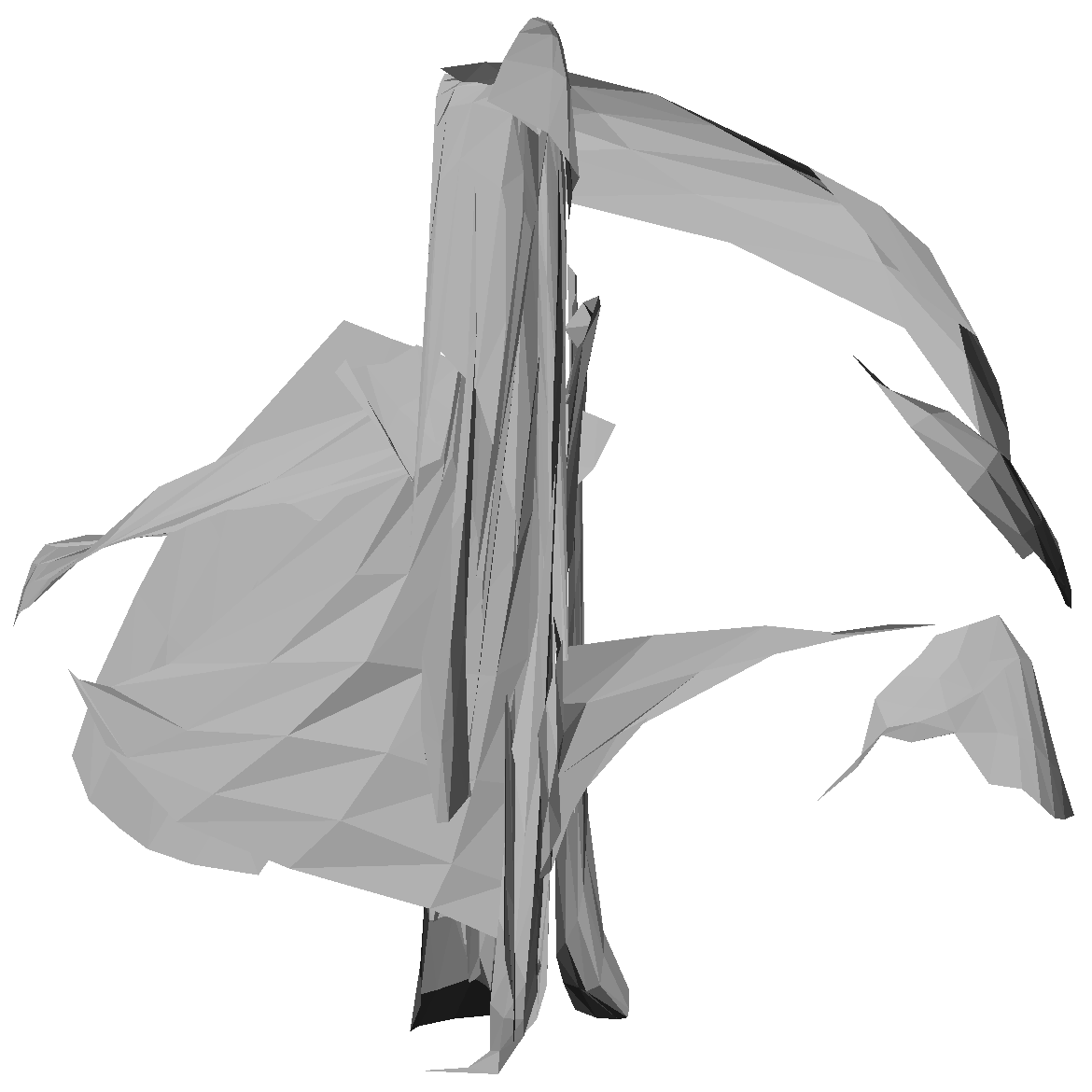}} &
{\includegraphics[width=0.22\linewidth]{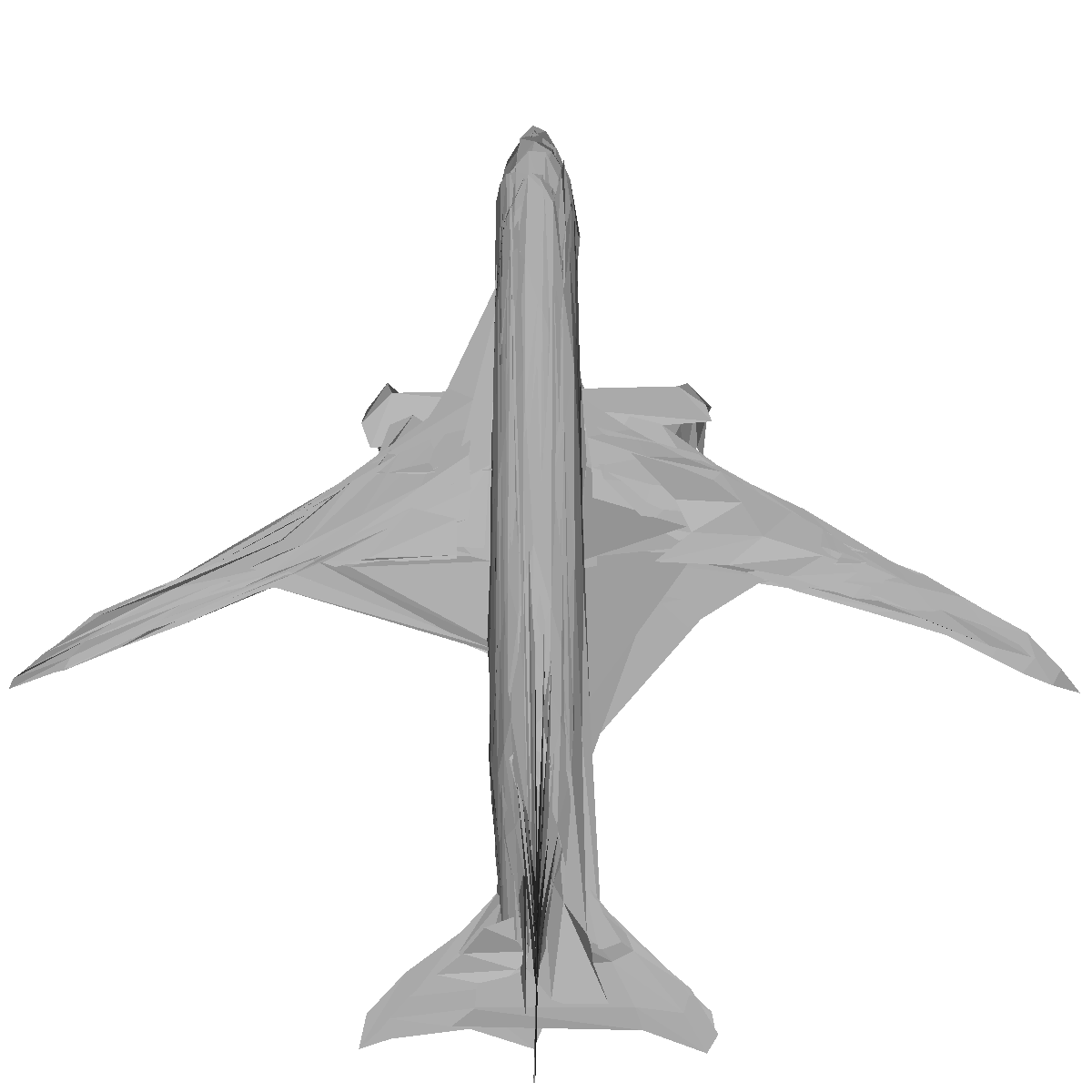}} &
{\includegraphics[width=0.2\linewidth]{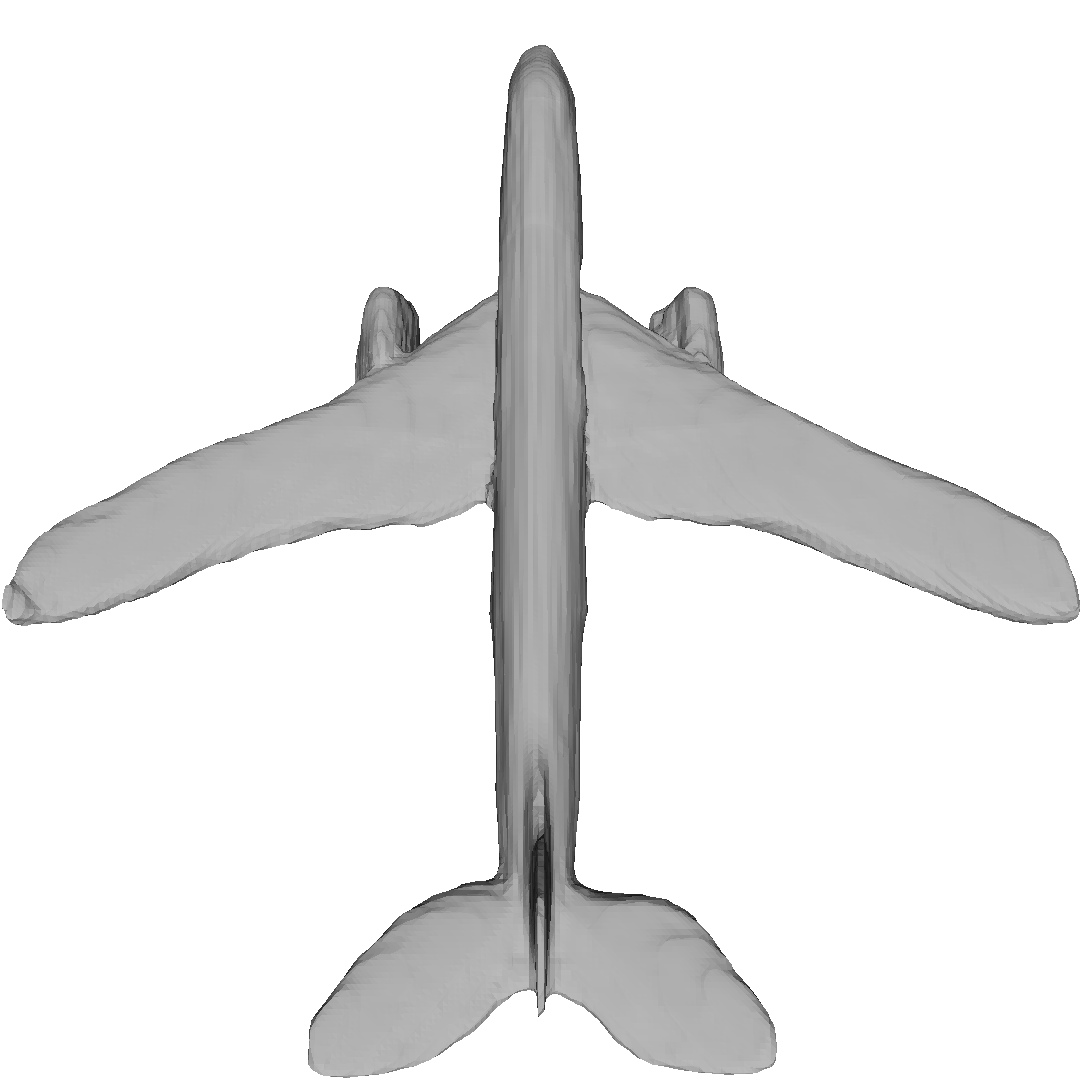}}
\\
{\includegraphics[width=0.25\linewidth]{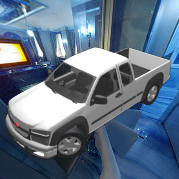}} &
{\includegraphics[width=0.22\linewidth]{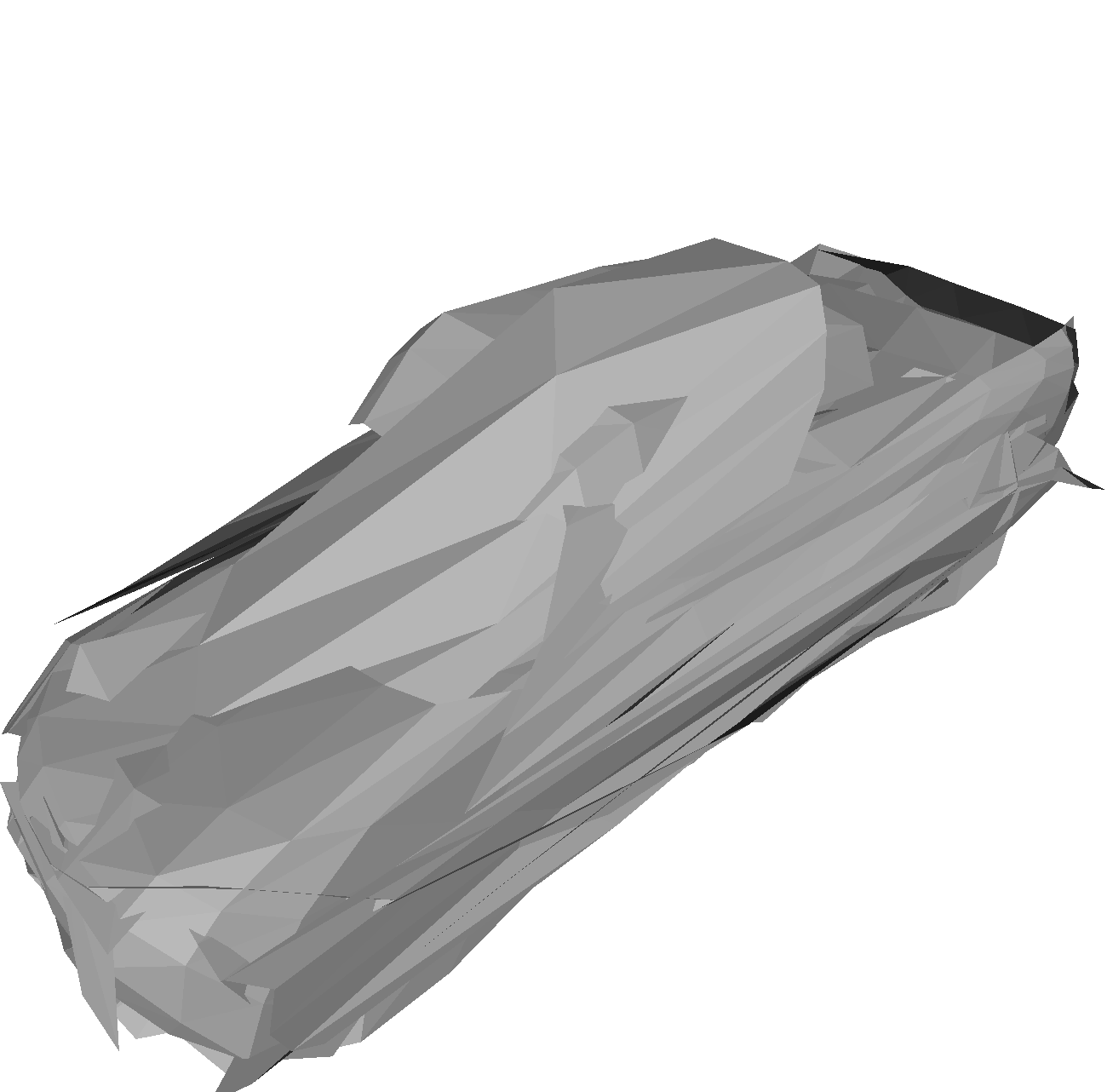}} &
{\includegraphics[width=0.22\linewidth]{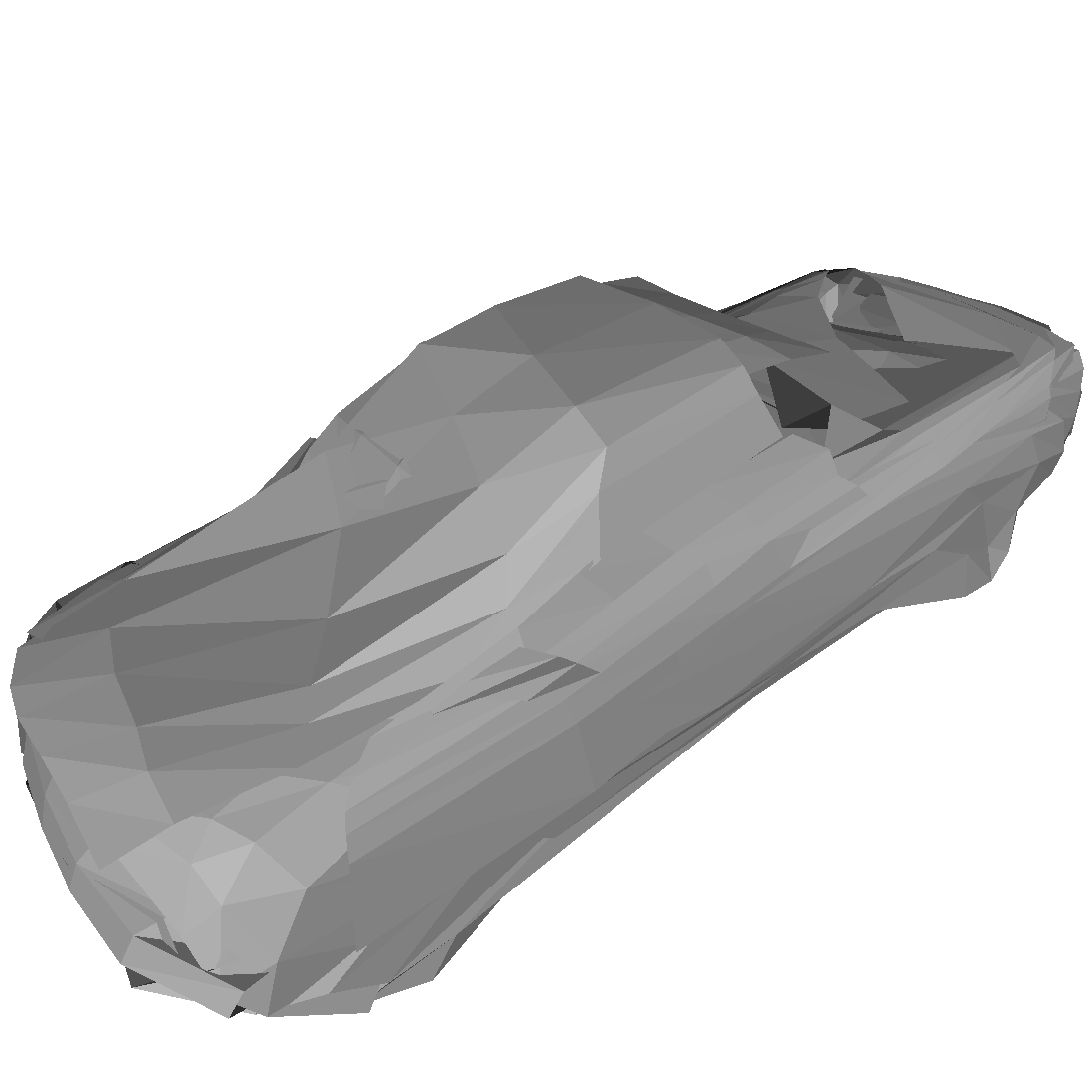}} &
{\includegraphics[width=0.22\linewidth]{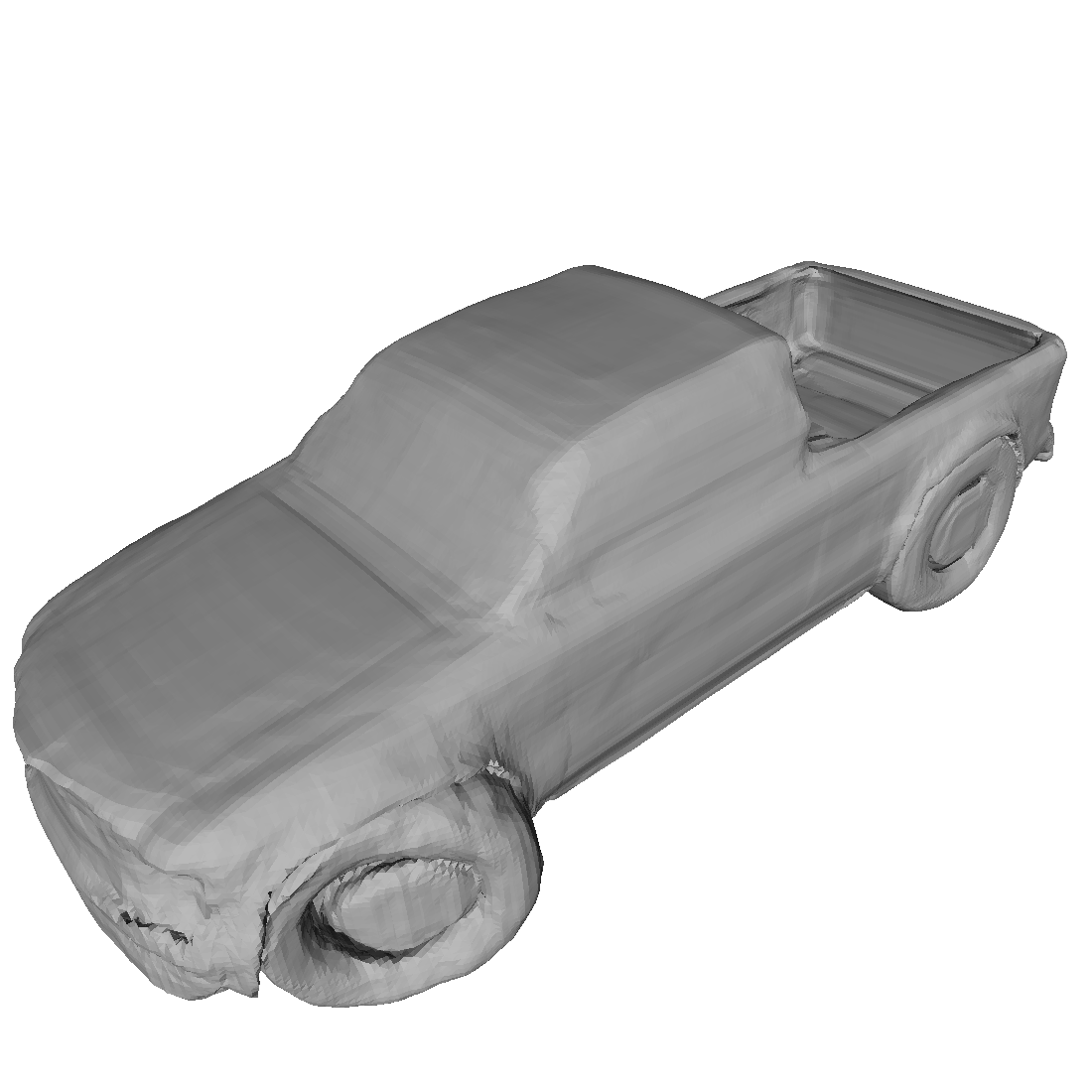}}
\\
{\includegraphics[width=0.25\linewidth]{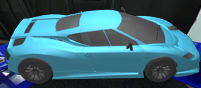}} &
{\includegraphics[width=0.22\linewidth]{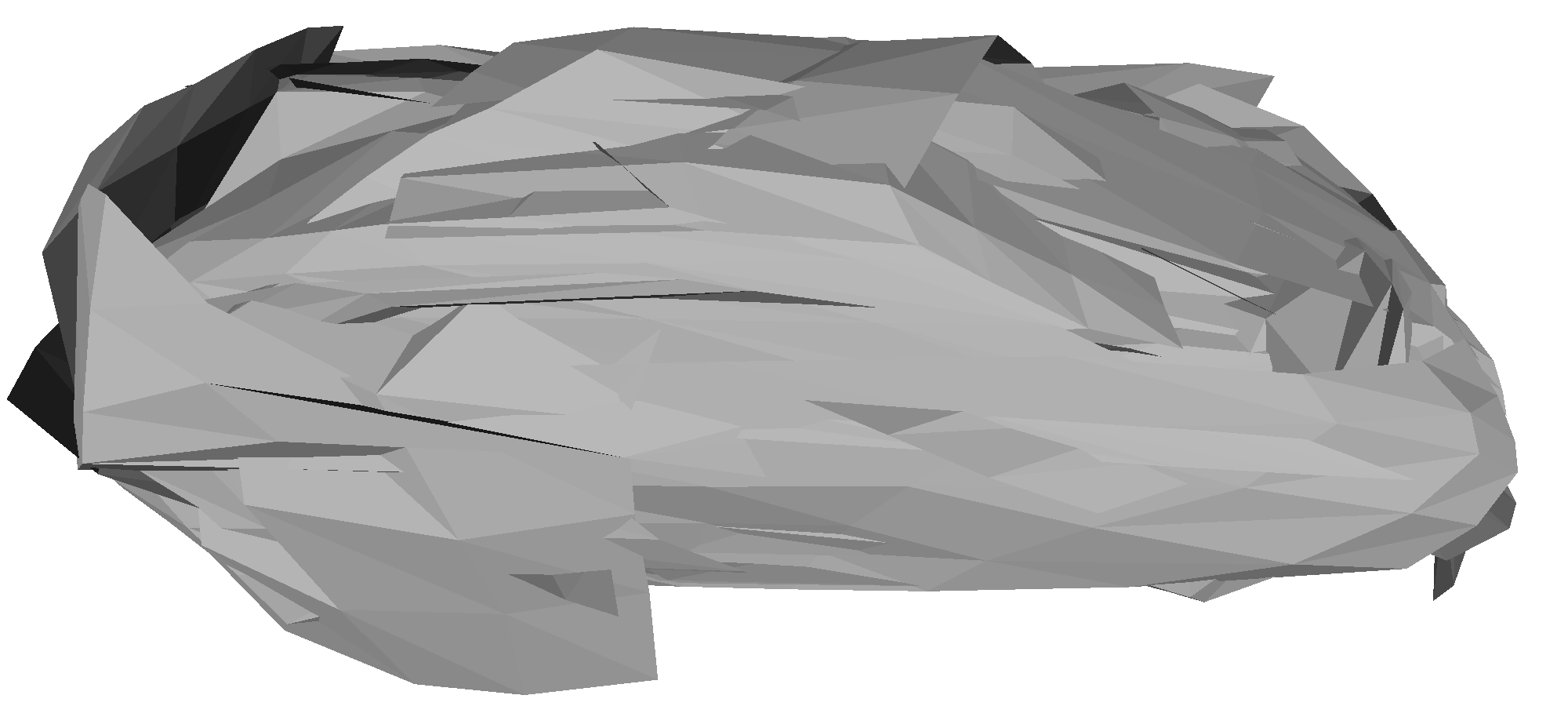}} &
{\includegraphics[width=0.22\linewidth]{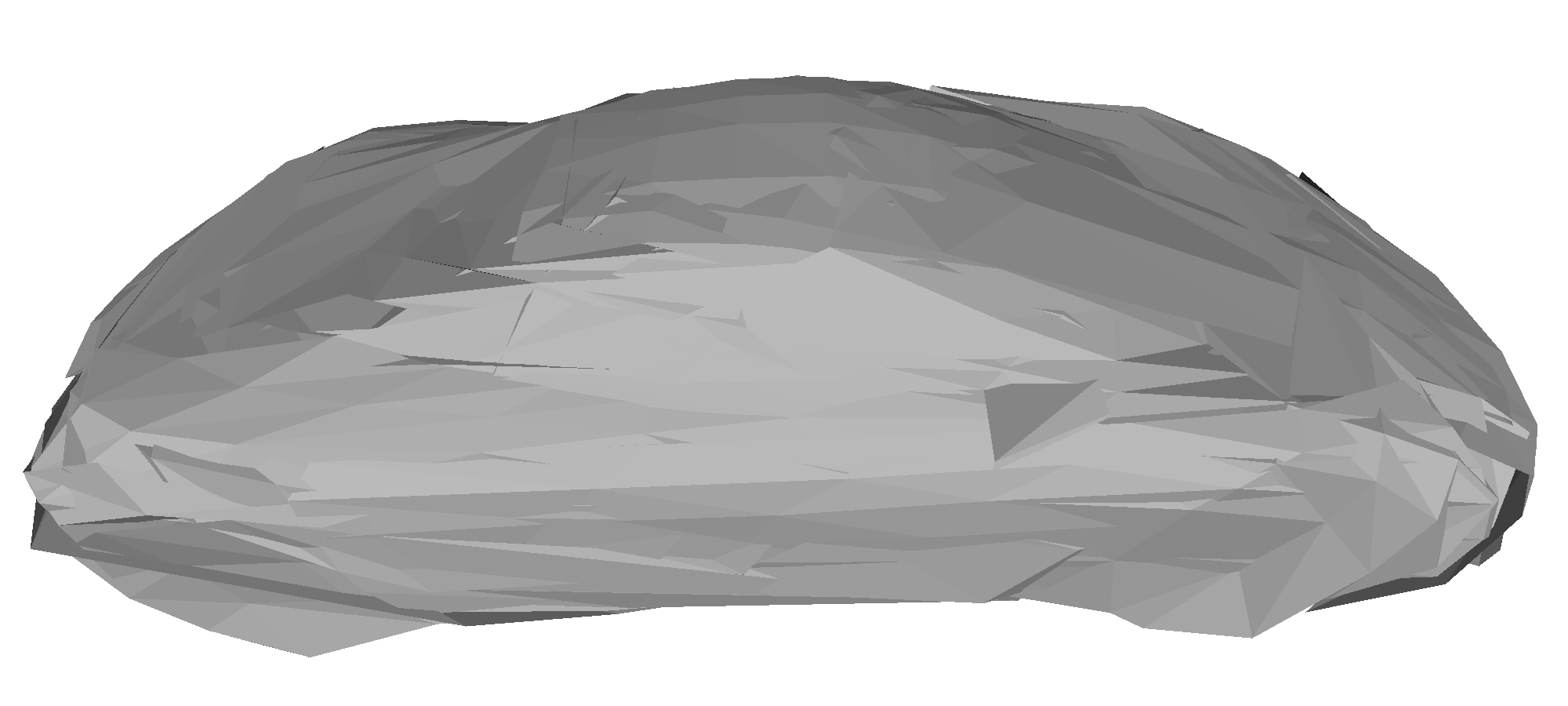}} &
{\includegraphics[width=0.22\linewidth]{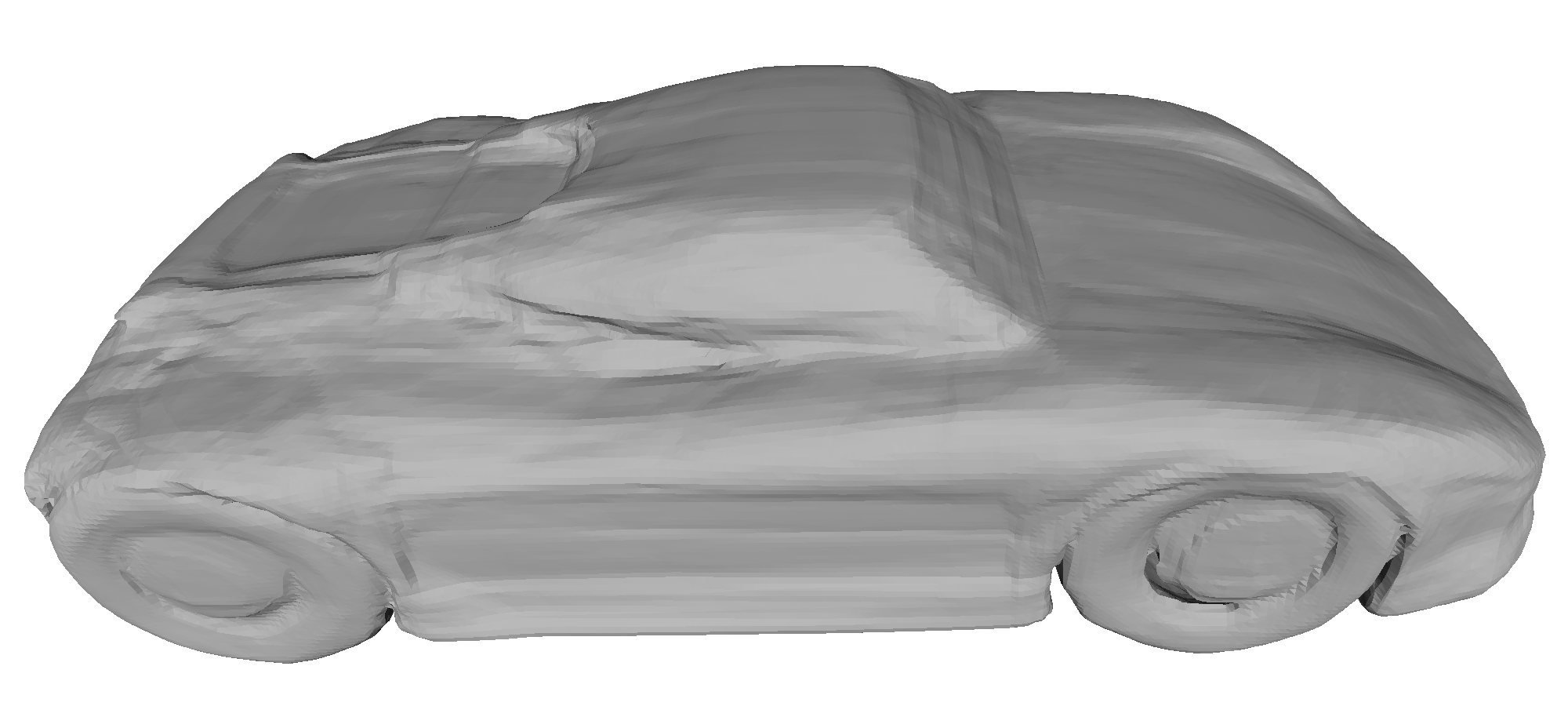}}
\\
\end{tabular}
\centering
\caption{Comparison on 3D shape prediction from multi-view images on the PMO test set. Our method maintains good surface details, while PMO suffers from the mesh representation and may not effectively optimize the shape.
}
\label{fig::mv_comp_pmo}
\vspace{-3pt}
\end{figure}

\noindent
\textbf{Generalization Capability}
To further evaluate the generalization capability, we compare to PMO on some unseen data and initialization. 
We first evaluate both methods on a testing set generated using different camera focal lengths, and the quantitative comparison is in \figref{fig::mv_ablation}~(a).
It clearly shows that our method generalizes well to the new images, while PMO suffers from overfitting or domain gap.
To further test the effectiveness of the geometric reasoning, we also directly add random noise to the initial latent code.
The performance of PMO again drops significantly, while our method is not affected since the initialization is randomized (\figref{fig::mv_ablation}~(b)).
Some qualitative results are shown in \figref{fig::mv_comp_pmo}.
Our method produces accurate shapes with detailed surfaces.
In contrast, PMO suffers from two main issues: 1) the low resolution mesh is not capable of maintaining geometric details; 2) their geometric reasoning struggles with the initialization from image encoder.

\begin{figure}[!htb]
\centering
\scriptsize
\setlength\tabcolsep{2.3pt} 
\begin{tabular}{ccc}
Video sequence & PMO & \textbf{Ours}
\\
{\includegraphics[width=0.35\linewidth]{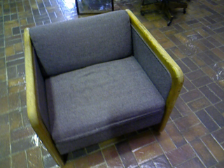}} &
{\includegraphics[width=0.28\linewidth]{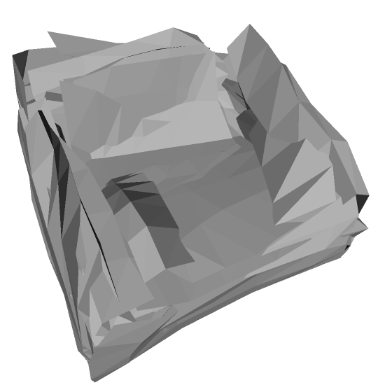}} &
{\includegraphics[width=0.28\linewidth]{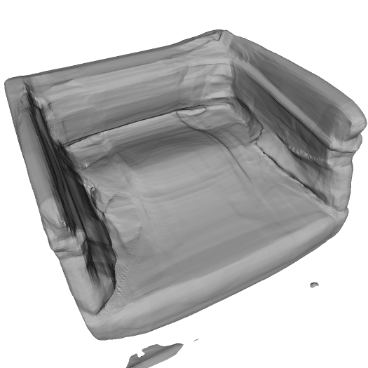}}
\\
{\includegraphics[width=0.35\linewidth]{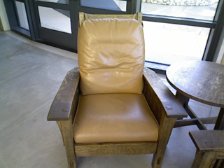}} &
{\includegraphics[width=0.28\linewidth]{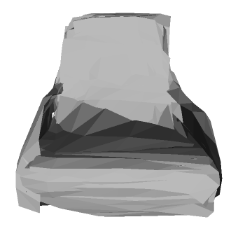}} &
{\includegraphics[width=0.28\linewidth]{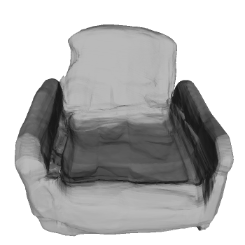}}
\\
\end{tabular}
\vspace{5pt}
\centering
\caption{Comparison on 3D shape prediction from multi-view images on real-world dataset~\cite{choi2016large}. 
It is in general challenging for shape prediction on real image. Comparatively, our method produces more reasonable results with correct structure.
}
\label{fig::mv_comp_real_resized}
\end{figure}

We further show comparison on real data in \figref{fig::mv_comp_real_resized}.
Following PMO, since the provided initial similarity transformation is not accurate in some cases, we also optimize over the similarity transformation in addition to the shape code. 
As can be seen, both methods perform worse on this challenging dataset.
In comparison, our method produces shapes with higher quality and correct structures, while PMO only produce very rough shapes.
Overall, our method shows better generalization capability and robustness against domain change.

\section{Conclusion}
We propose a differentiable sphere tracing algorithm to render 2D observations such as depth maps, normals, silhouettes, from implicit signed distance functions parameterized as a neural network.
This enables geometric reasoning in 3D shape prediction from both single and multiple views in conjunction with the high capacity 3D neural representation.
Extensive experiments show that our geometry based optimization algorithm produces 3D shapes that are more accurate than SOTA, generalizes well to new datasets, and is robust to imperfect or partial observations.
Promising directions to explore using our renderer include self-supervised learning, recovering other properties jointly with geometry, and neural image rendering.

\section*{Acknowledgements}
This work is partly supported by National Natural Science Foundation of China under Grant No. 61872012, National Key R\&D Program of China (2019YFF0302902), and Beijing Academy of Artificial Intelligence (BAAI).

{\small
\bibliographystyle{ieee_fullname}
\bibliography{egbib}
}

\section*{Appendix}
\appendix

In this supplementary material, we provide detailed analysis of the proposed renderer,
implementation details, and more qualitative results. 

\section{More Analysis on the Design of Differentiable Sphere Tracing}
\subsection{Benefits of Aggressive Marching}
We use an aggressive strategy (Sec. 3.2 in main submission) to speed up the sphere tracing.
Instead of marching with the step size as the SDF of the current location, we march $\alpha$ times of it, where $\alpha$ is larger than 1 and set as 1.5 by default.
This brings two benefits - faster convergence and stable training.

\paragraph{Faster Convergence}

\begin{figure}[tb]
\centering
{
\small
\includegraphics[width=0.8\linewidth]{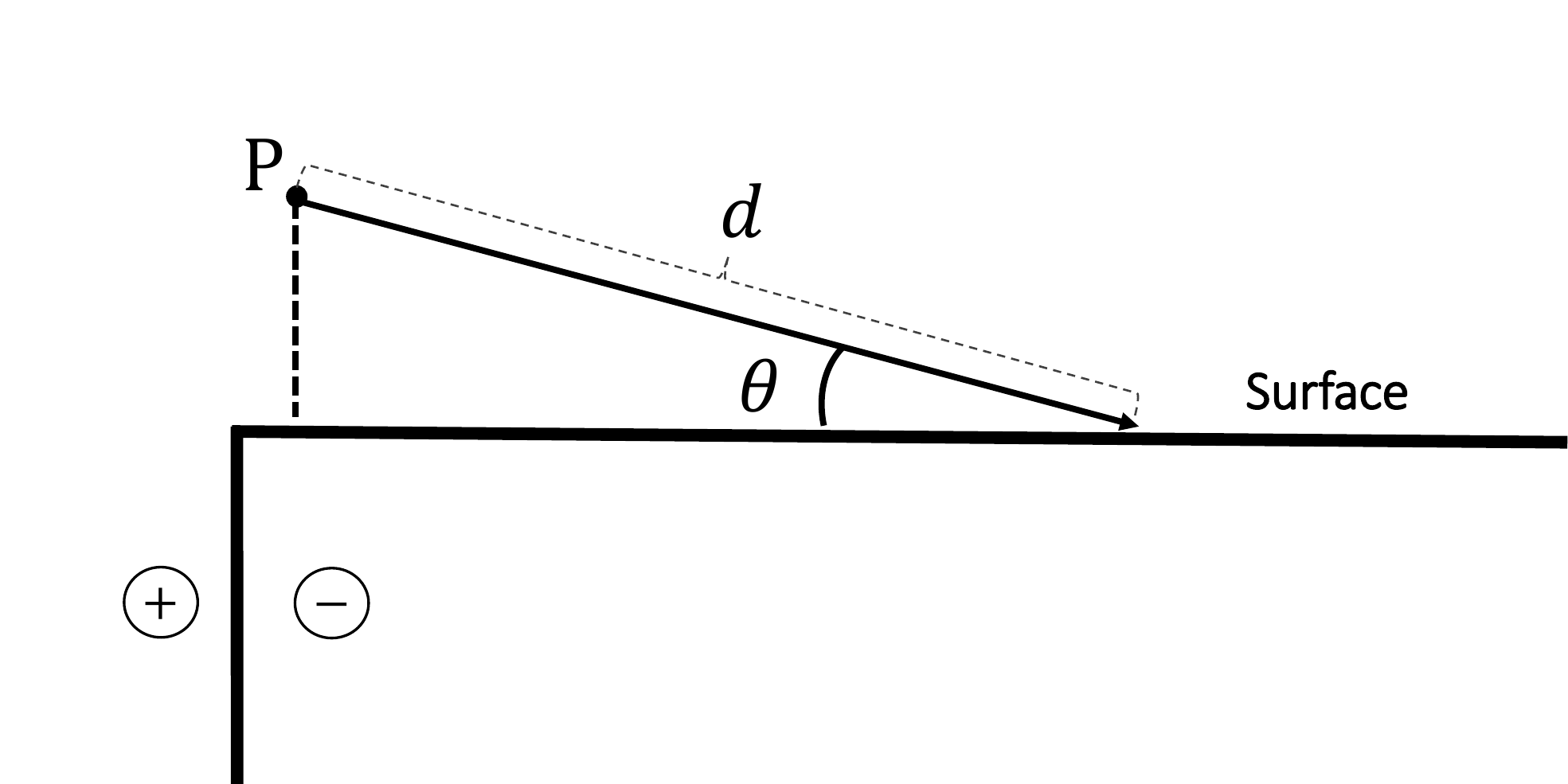}
}
\caption{Illustration of ill-posed conditions for sphere tracing algorithm. When $\theta$ is relatively small, the queried SDF is much lower than the actual ray distance $d$ towards the surface, making the sphere tracing process slower than expected.}
\label{fig::aggressive_marching}
\end{figure}

As shown in \figref{fig::aggressive_marching}, the sphere tracing algorithm can become unexpectedly slow when the angle between the camera ray and the surface is relatively small. From an initial point with a ray distance $d$ towards the surface, the marching step $k$ needs to satisfy the equation below to reach convergence:

\begin{equation}
    |d(1-\alpha sin\theta)^k| < \epsilon,
\end{equation}
where $\alpha$ equals to 1.0 in the conventional sphere tracing algorithm. When $|1-\alpha sin\theta| < 1$, we can easily derive the minimum marching step needed,
\begin{equation}
\label{eq::minimum-step}
    k > k_{min}=\frac{log\epsilon - logd}{log|1-\alpha sin\theta|}.
\end{equation}
By taking an aggressive strategy with $\alpha$ greater than 1 (we set it to 1.5 by default), the convergence can be speeded up under the ill-posed conditions. For example, suppose $d=1.0$, $\epsilon=5 \times 10^{-5}$, the minimum number of convergence steps decreases from 52 to 33 when $\theta$ equals to 10 degrees.

\begin{figure}[tb]
\centering
{
\small
\includegraphics[width=0.7\linewidth]{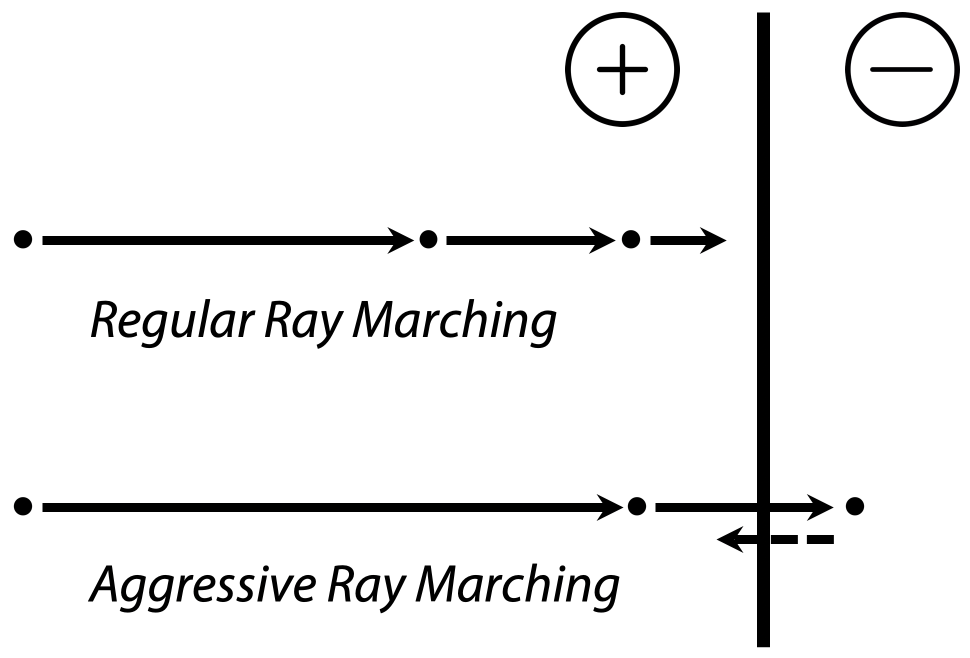}
}
\caption{Compared to the regular ray marching algorithm, the aggressive ray marching strategy can march inside the surface and bounce back-and-forth between the inside and outside areas. This gives more samples on the negative side of implicit signed distance function and benefits the optimization.}
\label{fig::supp_aggressive_marching_negative}
\end{figure}

\paragraph{Stable Training}
Besides speeding up the overall marching, the aggressive marching also allows more samplings from the locations behind the surface, which adds supervision at the interior of the shape and stabilize the training.
As shown in \figref{fig::supp_aggressive_marching_negative}, in traditional ray marching, the front end of the ray approaches the surface from the camera side (\ie $SDF>0$) and less likely to trespass the surface.
In contrast, our marching is more likely to pass through the surface (and for sure under the ground truth SDF when the ray direction is orthogonal to the surface since the marching step is larger than SDF).
This will not add much computational overhead when working conjointly with the convergence criteria, but achieves ray convergence from both sides the surface.
This gives the training more supervision with both positive and negative SDF, compared to positive only using regular marching.
The aggressive marching also naturally samples more points near the surface, which are important for network to learn surface details.
Coincidentally, DeepSDF~\cite{park2019deepsdf} also mentioned the importance of sampling more points near surface.
While they need to rely on extra ground truth normal and depth to perform the sampling, our method is fully automatic and sample adaptively according to the SDF field.

\subsection{Convergence Criteria}
\begin{figure}[tb]
\centering
{
\small
\includegraphics[width=0.8\linewidth]{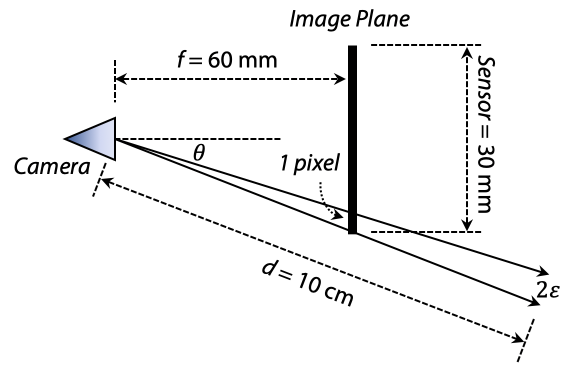}
}
\caption{Illustration on the geometric meaning of the threshold $\epsilon$ of the convergence criteria. }
\label{fig::supp_converge}
\end{figure}

It is important to define a proper convergence criteria as shown in Fig.~7 of the main submission.
In our differentiable sphere tracing, a ray stops marching if the absolute SDF is smaller than certain threshold $\epsilon$.
Essentially, this means that the true intersection on the surface is bounded by a ball with radius of $\epsilon$ centered at our current sampling point.
There are two guidelines to select this threshold. 
On one hand, the threshold should not be too large, since the rendering noise is theoretically bounded by this threshold. Large threshold may result in large error in the rendered depth.
On the other hand, the threshold must not be too small. The rendering time will be significantly longer if it is too small since more queries would be needed. 
Moreover, with a fixed maximum number of tracing steps, some pixels may not converge and thus are considered as background, which causes erosion (as shown in Fig. 7 in our main submission).
Based on these two observations, we propose to define the threshold as the distance where the ray front ends from neighboring pixels are clearly separable.
This is equivalent to finding the radius such that balls centered at the ray front ends of neighboring pixels do not intersect with each other.
\figref{fig::supp_converge} demonstrates how to compute the $\epsilon$.
For a camera with focal length $f$, sensor size $S$, and resolution $R$, we get the following equation for objects roughly $d_{min}$ away from the camera, according to similar triangles:
\begin{equation}
    \frac{S/R \cdot cos(\theta)}{f/cos(\theta)} = \frac{2\epsilon}{d_{min}}
\end{equation}



This gives:
\begin{equation}
    \epsilon=\frac{d_{min}\cdot S\cdot cos^2(\theta)}{2\cdot f \cdot R}
\end{equation}
Taking the common set up shown in~\figref{fig::supp_converge}, where $f=60mm, d_{min}=10cm, S=32mm, R=512$, we get $\epsilon \approx 0.5\times 10^{-4}m$ ($0.05mm$). 

\subsection{Differentiable Rendering of Silhouette}
\begin{figure}[tb]
\scriptsize
\centering
\setlength\tabcolsep{2.3pt} 
\begin{tabular}{ccc}
\\
{\includegraphics[width=0.32\linewidth]{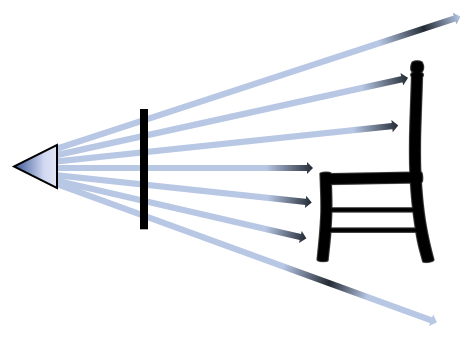}} &
{\includegraphics[width=0.25\linewidth, height=55pt]{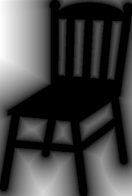}} &
{\includegraphics[width=0.25\linewidth, height=55pt]{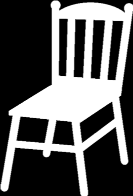}}
\\
(a) Sphere Tracing & (b) $\texttt{min}(\texttt{abs}(SDF))$ & (c) $\texttt{min}(\texttt{abs}(SDF))<\epsilon$
\end{tabular}
\vspace{2pt}
\caption{The differentiable rendering of silhouette. We take minimum absolute SDF value along each ray and determine the silhouette by checking whether the minimum absolute value is less than the threshold $\epsilon$ or not. We also consider the rendered soft silhouette as the minimum absolute queries substracted by $\epsilon$, which is fully differentiable and feasible for optimization.}
\label{fig::silhouette}
\end{figure}

\figref{fig::silhouette} shows how to render the silhouette in a differentiable way.
After running our deep sphere tracing (\figref{fig::silhouette}~(a)), we render the minimal absolute SDF on each pixel, and get the soft silhouette by substracting it by $\epsilon$ (\figref{fig::silhouette}~(b)). In this way the binary silhouette (\figref{fig::silhouette}~(c)) can be easily acquired by checking whether the rendered silhouette is positive (background) or not (foreground).

Because our rendered silhouette is fully differentiable with respect to implicit signed distance functions and camera extrinsic parameters, we can define differentiable loss term over the rendered silhouette $S_r$ and the ground-truth binary silhouette $S_{gt}$. The silhouette loss $\mathcal{L}_s$ can be formulated as below:

\begin{equation}
\label{eq::silhouette-loss}
    \mathcal{L}_s = S_{gt}max(0, S_{r}) + (1 - S_{gt})max(0, -S_{r}).
\end{equation}

\begin{figure}[tb]
{\includegraphics[width=0.8\linewidth,height=100pt]{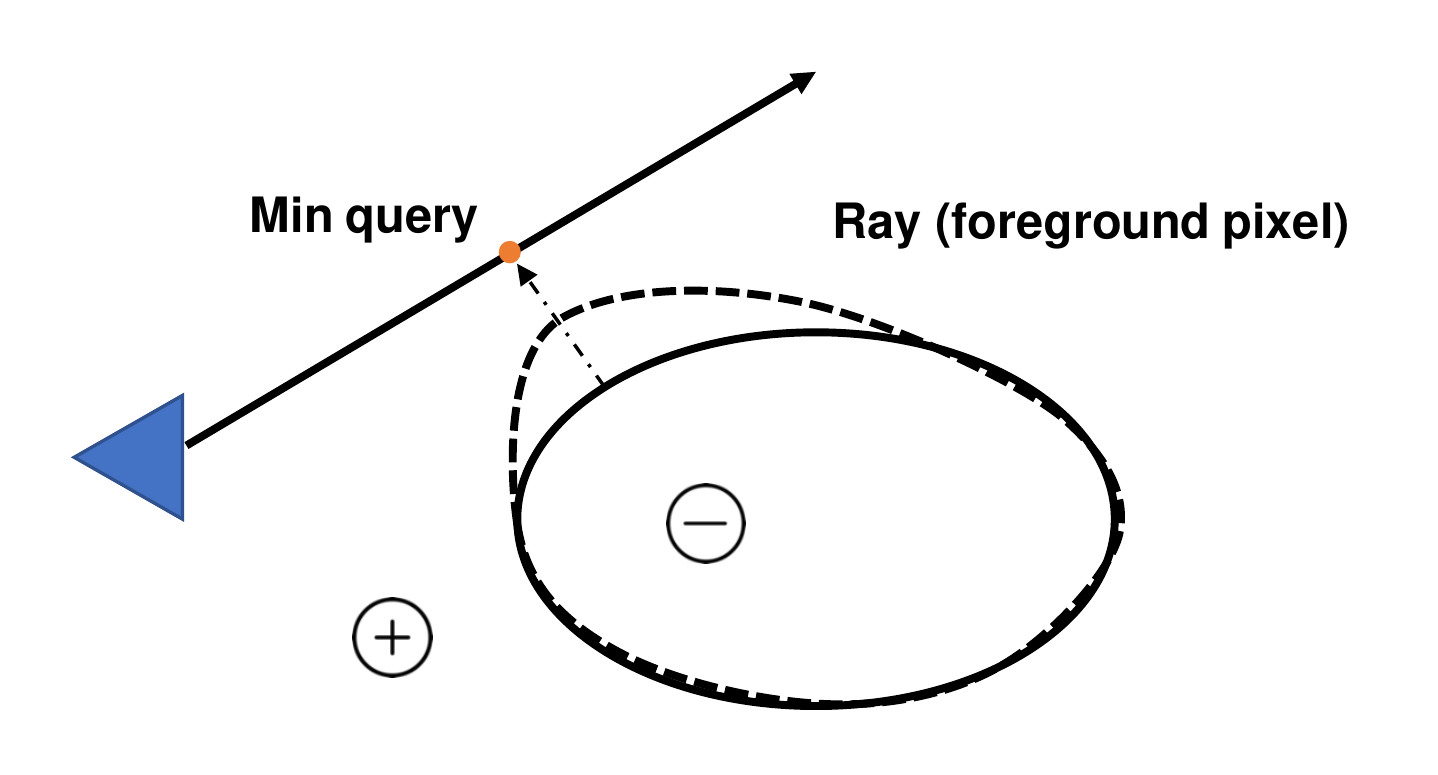}}
\centering
\caption{Differentiable error propagation along the boundary of the foreground. We can make use of the nice property of signed distance fields to optimize the nearest surface.}
\label{fig::diff-silhouette}
\end{figure}

This formulation is able to get the silhouette error differentiably back-propagate to the optimized parameters. Note that by using the minimum absolute query, we utilize the nice individual property for signed distance functions, where the nearest surface with respect to the camera ray is optimized (as shown in Fig. \ref{fig::diff-silhouette}). This strategy makes the shape optimization smooth and effective. Intuitively, combining the rendered silhouette with the distance transform on the 2D image plane can further improve the efficacy of the term, which is left for future work.

As discussed in the main paper, we check whether the ray intersects with the unit sphere to generate an initialization mask, where the ray without intersection with the unit sphere is directly set to background. To make the rendered silhouette on those background pixels differentiable, we set the soft silhouette value on each of those pixels to be the distance from the origin to the corresponding camera ray minus 1.0. This design shares similar spirits with the previous design on differentiable rendering of the silhouette. Because the distance from the origin to each of those camera rays is always greater than 1.0, we can consistently check whether the rendered silhouette is negative to determine its corresponding binary silhouette.

\subsection{Drawbacks}

\begin{figure}[tb]
\centering
\setlength\tabcolsep{2.3pt} 
\scriptsize
\begin{tabular}{cccc}
parallel & + dynamic & + aggressive & + coarse-to-fine
\\
{\includegraphics[trim={40 40 40 40}, clip, width=0.21\linewidth]{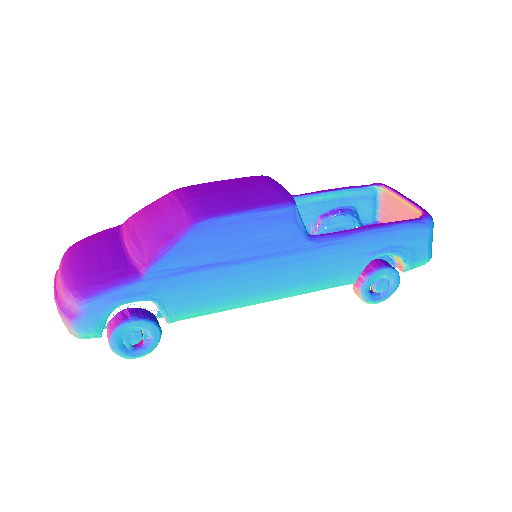}} &
{\includegraphics[trim={40 40 40 40}, clip, width=0.21\linewidth]{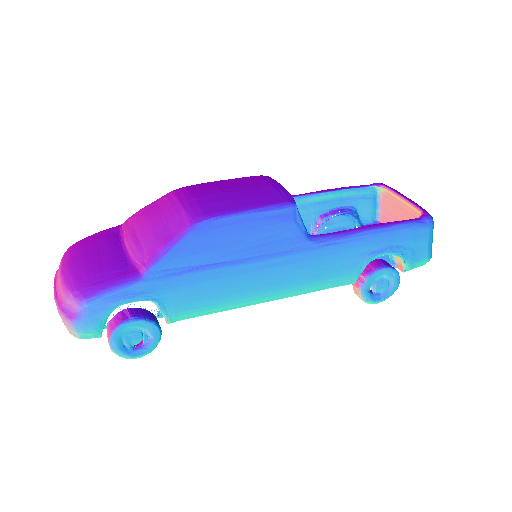}} &
{\includegraphics[trim={40 40 40 40}, clip, width=0.21\linewidth]{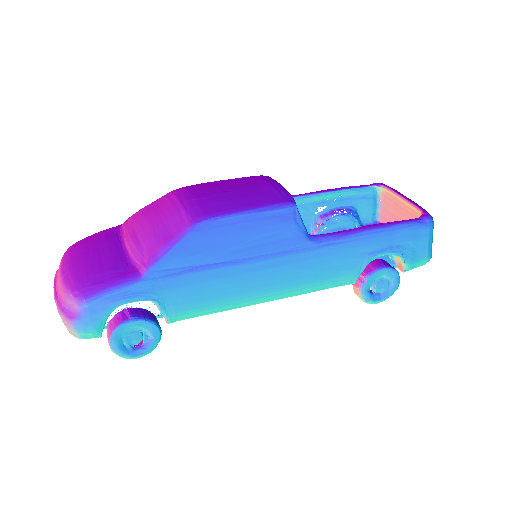}} &
{\includegraphics[trim={40 40 40 40}, clip, width=0.21\linewidth]{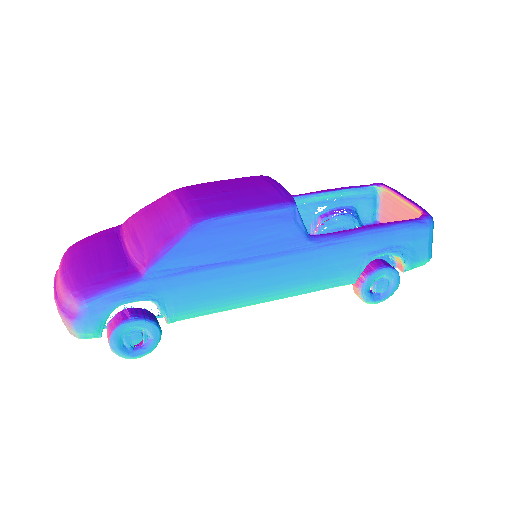}}
\\
\end{tabular}
\centering
\caption{Illustration of the small artifacts induced by the aggressive strategy. Small holes could occur on thin surface areas. Better viewed when zoomed in. }
\label{fig::artifact}
\end{figure}

Most of our acceleration strategy does not affect the rendering quality, except the aggressive marching.
Fig.~\ref{fig::artifact} shows the drawback of the aggressive tracing strategy. 
This mostly happens for the case when the geometry is super thin such that a single step of marching may trespass two surfaces.
As a result, the ray front end is still considered out of the shape and will keep marching till infinite, which causes artifacts shown in the back part of the truck. 

Another potential drawback is the well-known aliasing effect, since for each pixel there is only one ray shot from the pixel center. Many well-known anti-aliasing strategies can be directly applied to our renderer to mitigate this issue.

\section{Implementation Details}
\subsection{Network Architecture}
We follow the same network architecture as DeepSDF \cite{park2019deepsdf}, which consists of 9 fully connected layers. Each hidden layer has a dimension of 512. For the texture re-rendering applications, we concatenate the shape code and texture code together and feed the concatenated code to the texture network that employs the same architecture as the geometry network. Both the shape code and texture code has a dimension of 256. Note that the network architecture of DeepSDF \cite{park2019deepsdf} is much heavier than the backbone used in \cite{sitzmann2019scene} (4 layers with 256 dimensions for each). However, with our proposed advanced sphere tracing strategies, we can produce high-resolution images within limited time consumption overhead. 

\subsection{Implementation of Dynamic Synchronized Inference}
Here we present some details on the implementation of dynamic synchronized inference with the off-the-shelf deep learning framework. We maintain a binary flag over each camera ray during the sphere tracing process. For each step, we concatenate all unfinished camera rays together and perform feedforward in a batch-wise manner and map them back to the original image resolution. Then, we check whether the tracing on each ray converges or gets out of the unit sphere and set the corresponding binary flags to zero. Note that because the operation is performed on a concatenated tensor in the computational graph, trivially computing the minimum absolute SDF value for each ray results in unaffordable memory consumption. To address this issue, we introduce an implementation trick that the location of each query is saved globally, and minimization is performed over a detached tensor graph. After this operation, we get the minimum K queries for each ray and feedforward those queries again with the gradients attached. This strategy enables our renderer to produce much high resolution images ($2048\times2048$) on a single GTX-1080Ti. 

\subsection{Depth / Normal / Silhouette / Color Loss}
Our method renders 2D observations in a differentiable manner and computes the loss on the image plane. The depth loss and normal loss are computed over the foreground region determined by the rendered silhouette. For the multi-view shape reconstruction application, we compute the photometric error over the commonly visible pixels in two views.
The visibility can be determined by computing the difference $\Delta_d$ between the reprojected depth from the source view and the directly rendered depth of the target view. In our experiments, the pixels with $\Delta_d^2 < 0.001$ are considered as visible.
$L_1$ loss is used for depth and photometric error computation, and negative dot product is computed to measure the loss of surface normals \cite{eigen2015predicting}. For the rendered silhouette (substracting minimum absolute query with $\epsilon$), we require that the value should be negative over the foreground and positive over the background and use the loss term defined in Eq. (\ref{eq::silhouette-loss}). 

\subsection{More Details and Hyperparameters}
Our framework is implemented in PyTorch and code will be made publicly available. For all experiments, we use the Adam optimizer with the initial learning rate 1e-2. To compute the Chamfer Distance for evaluation, we sample 30k points on depth completion and 10k points on multi-view shape reconstruction respectively. We follow the common practice to report the distance scaled by 1000. 
In the multi-view scenario, the 3D shapes from the DeepSDF decoder may consist of structures inside some objects but the ground-truth meshes only have the structure on the surface. Therefore, we report the Chamfer Distance only in the direction of gt$\rightarrow$pred for fair comparison.
For the sampling stretegy when rendering depth, empirically, $K=1$ already works reasonably well. We use $K=3$ for shape completion and $K=1$ for the multi-view shape reconstruction. The maximum marching step we use is 100.
Empirically, a value choice ranging from 1.2 to 1.8 can produce an effective $\alpha$. Small $\alpha$ results in slow convergence while large $\alpha$ can lead to small holes in thin areas.

For shape completion, we initialize the latent code to be zero (which denotes the mean shape) and perform optimization for 100 iterations. The loss weights of the depth loss and silhouette loss are set to 10.0 and 1.0 respectively. We also follow \cite{park2019deepsdf} to add an $\ell_2$ regularizer over the shape code during optimization and the loss weight of the regularization term is set to 1.0. 

For multi-view shape reconstruction under the PMO \cite{lin2019photometric} synthetic test set, for each iteration we sample 8 views uniformly distributed $360^{\circ}$ around the object, warp each of them and calculate the photometric loss to the closest next view based on the estimated depth image. We downsample input images to its half size $112\times112$ in order to back-propagate the photometric and regularization losses from all 8 views together. Since there are 72 views provided in the dataset for each object, we run 9 iterations for each epoch and 20 epochs in total. 

As for the real-world multi-view dataset, in contrast to PMO \cite{lin2019photometric} which uses over 100 images of an object, we only picked between 20 and 30 views and further downsample them from $480\times640$ to $96\times128$. 6 views are selected during each iteration.  Since the provided initial similarity transformation is not accurate in some cases, we also optimize over this similarity transformation in addition to the shape code. We find out that it is usually enough to acquire an accurate similarity transformation after only 1 or 2 epochs. The weight of the photometric loss is set to 5.0 for both synthetic and real-world experiments.

\section{Rendering Demos}
\begin{figure*}[tb]
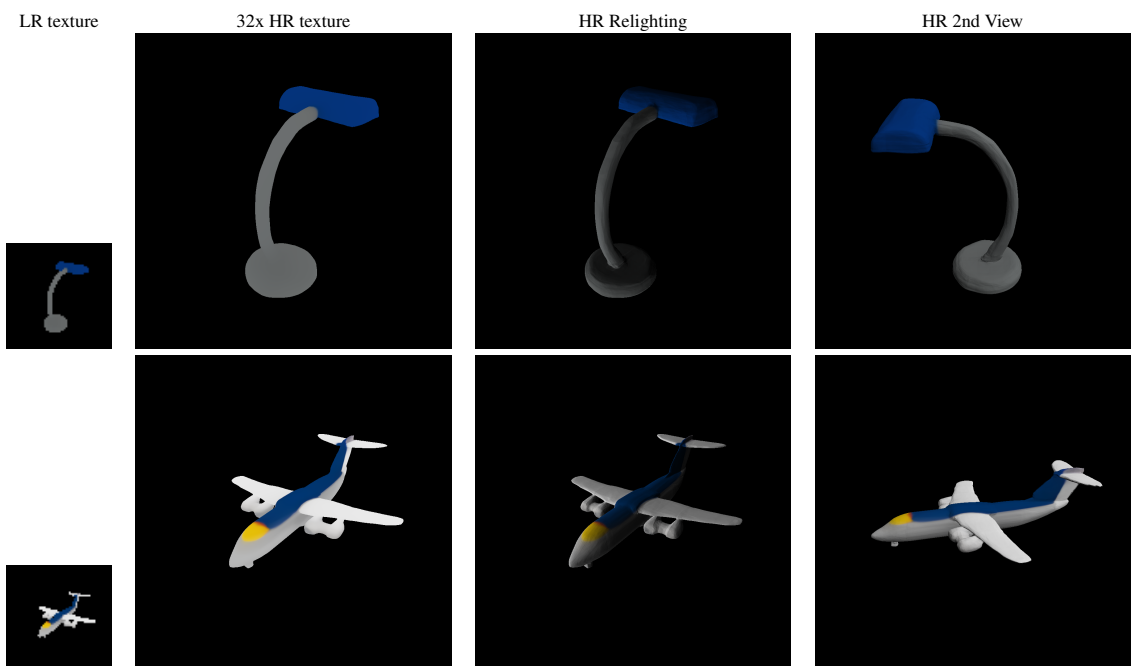

\centering
\scriptsize
\setlength\tabcolsep{4.6pt} 
\begin{tabular}{cccc}
LR texture & 32x HR texture &  HR Relighting & HR 2nd View
\\
{\includegraphics[width=0.08\linewidth]{img_src/demo_rerender/lamp_10_lr.png}} &
{\includegraphics[width=0.24\linewidth]{img_src/demo_rerender/lamp_10_sr_16x.png}} &
{\includegraphics[width=0.24\linewidth]{img_src/demo_rerender/lamp_10_sr_16x_relight.png}} &
{\includegraphics[width=0.24\linewidth]{img_src/demo_rerender/lamp_10_sr_16x_nv.png}}\\
{\includegraphics[width=0.08\linewidth]{img_src/demo_rerender/plane_27_lr.png}} &
{\includegraphics[width=0.24\linewidth]{img_src/demo_rerender/plane_27_sr.png}} &
{\includegraphics[width=0.24\linewidth]{img_src/demo_rerender/plane_27_sr_relight2.png}} &
{\includegraphics[width=0.24\linewidth]{img_src/demo_rerender/plane_27_sr_nv2.png}}
\end{tabular}
\centering
\caption{
Qualitative results on the applications on texture re-rendering, where we can generate high-resolution outputs under various resolution, camera viewpoints and illumination.
}
\label{fig::supp_texture_render}
\end{figure*}

We attach a video demo in the supplementary material to show that our method can render high-resolution depth, surface normals, silhouette and RGB image with various lighting conditions and camera viewpoints. Also, more qualitative results on 
multi-view reconstruction are included in the video. We also show a larger version on the texture re-rendering demo (Fig. 8 in the main paper) in~\figref{fig::supp_texture_render}.

\section{Additional Experimental Results}

\subsection{Qualitative Comparison for Shape Completion}
\begin{figure*}[!tb]
\centering
\scriptsize
\setlength\tabcolsep{2.3pt} 
\begin{tabular}{c|cccccc}
\hline
& dense & 50\% pts & 10\% pts & 100 pts & 50 pts & 20 pts \\
\hline
Input depth (direct view) &
{\includegraphics[width=0.12\linewidth]{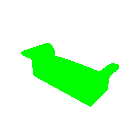}} &
{\includegraphics[width=0.12\linewidth]{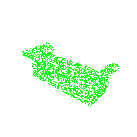}} &
{\includegraphics[width=0.12\linewidth]{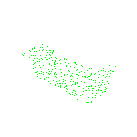}} &
{\includegraphics[width=0.12\linewidth]{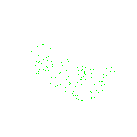}} &
{\includegraphics[width=0.12\linewidth]{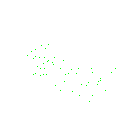}} &
{\includegraphics[width=0.12\linewidth]{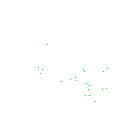}}
\\
\hline
DeepSDF \cite{park2019deepsdf} (direct view) &
{\includegraphics[width=0.12\linewidth]{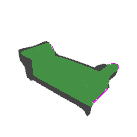}} &
{\includegraphics[width=0.12\linewidth]{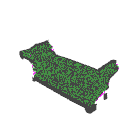}} &
{\includegraphics[width=0.12\linewidth]{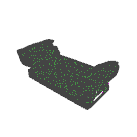}} &
{\includegraphics[width=0.12\linewidth]{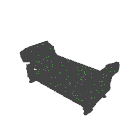}} &
{\includegraphics[width=0.12\linewidth]{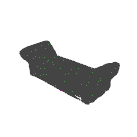}} &
{\includegraphics[width=0.12\linewidth]{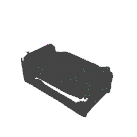}}
\\
DeepSDF \cite{park2019deepsdf} (second view) &
{\includegraphics[width=0.12\linewidth]{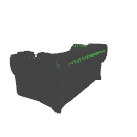}} &
{\includegraphics[width=0.12\linewidth]{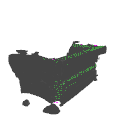}} &
{\includegraphics[width=0.12\linewidth]{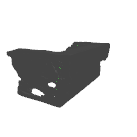}} &
{\includegraphics[width=0.12\linewidth]{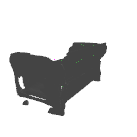}} &
{\includegraphics[width=0.12\linewidth]{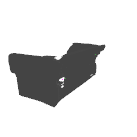}} &
{\includegraphics[width=0.12\linewidth]{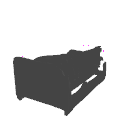}}
\\
DeepSDF \cite{park2019deepsdf} (third view) &
{\includegraphics[width=0.12\linewidth]{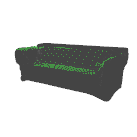}} &
{\includegraphics[width=0.12\linewidth]{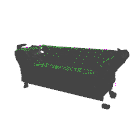}} &
{\includegraphics[width=0.12\linewidth]{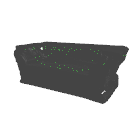}} &
{\includegraphics[width=0.12\linewidth]{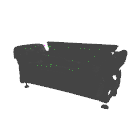}} &
{\includegraphics[width=0.12\linewidth]{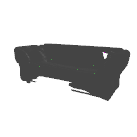}} &
{\includegraphics[width=0.12\linewidth]{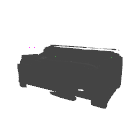}}
\\
\hline
Ours (direct view) &
{\includegraphics[width=0.12\linewidth]{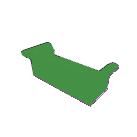}} &
{\includegraphics[width=0.12\linewidth]{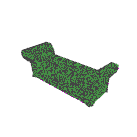}} &
{\includegraphics[width=0.12\linewidth]{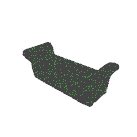}} &
{\includegraphics[width=0.12\linewidth]{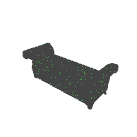}} &
{\includegraphics[width=0.12\linewidth]{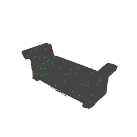}} &
{\includegraphics[width=0.12\linewidth]{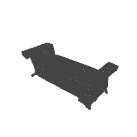}}
\\
Ours (second view) &
{\includegraphics[width=0.12\linewidth]{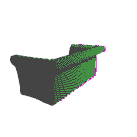}} &
{\includegraphics[width=0.12\linewidth]{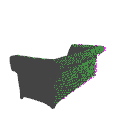}} &
{\includegraphics[width=0.12\linewidth]{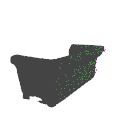}} &
{\includegraphics[width=0.12\linewidth]{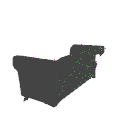}} &
{\includegraphics[width=0.12\linewidth]{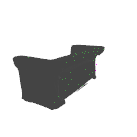}} &
{\includegraphics[width=0.12\linewidth]{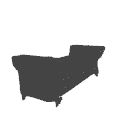}}
\\
Ours (third view) &
{\includegraphics[width=0.12\linewidth]{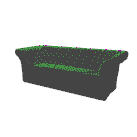}} &
{\includegraphics[width=0.12\linewidth]{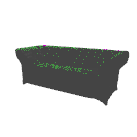}} &
{\includegraphics[width=0.12\linewidth]{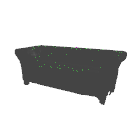}} &
{\includegraphics[width=0.12\linewidth]{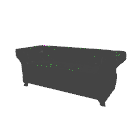}} &
{\includegraphics[width=0.12\linewidth]{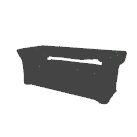}} &
{\includegraphics[width=0.12\linewidth]{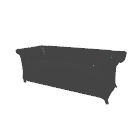}}
\\
\hline
Ours w. mask (direct view) &
{\includegraphics[width=0.12\linewidth]{img_src/supp_depth_completion/sofa_44/ours_10_direct_output.png}} &
{\includegraphics[width=0.12\linewidth]{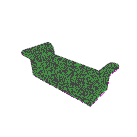}} &
{\includegraphics[width=0.12\linewidth]{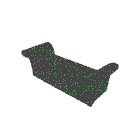}} &
{\includegraphics[width=0.12\linewidth]{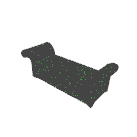}} &
{\includegraphics[width=0.12\linewidth]{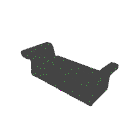}} &
{\includegraphics[width=0.12\linewidth]{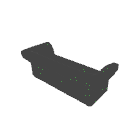}}
\\
Ours w. mask (second view) &
{\includegraphics[width=0.12\linewidth]{img_src/supp_depth_completion/sofa_44/ours_10_az210_output.png}} &
{\includegraphics[width=0.12\linewidth]{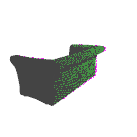}} &
{\includegraphics[width=0.12\linewidth]{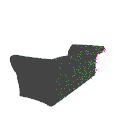}} &
{\includegraphics[width=0.12\linewidth]{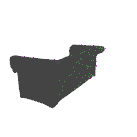}} &
{\includegraphics[width=0.12\linewidth]{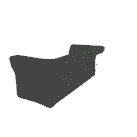}} &
{\includegraphics[width=0.12\linewidth]{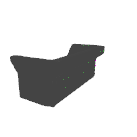}}
\\
Ours w. mask (third view) &
{\includegraphics[width=0.12\linewidth]{img_src/supp_depth_completion/sofa_44/ours_10_az120_output.png}} &
{\includegraphics[width=0.12\linewidth]{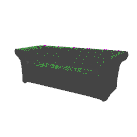}} &
{\includegraphics[width=0.12\linewidth]{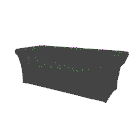}} &
{\includegraphics[width=0.12\linewidth]{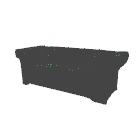}} &
{\includegraphics[width=0.12\linewidth]{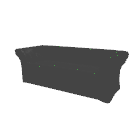}} &
{\includegraphics[width=0.12\linewidth]{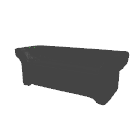}}
\\
\hline
\end{tabular}
\vspace{5pt}
\centering
\caption{Qualitative comparisons on shape completion under different sparsity of input depth (sofa).}

\label{fig::supp_depth_completion_sofa}
\vspace{-5pt}
\end{figure*}

\begin{figure*}[!tb]
\centering
\scriptsize
\setlength\tabcolsep{2.3pt} 
\begin{tabular}{c|cccccc}
\hline
& dense & 50\% pts & 10\% pts & 100 pts & 50 pts & 20 pts \\
\hline
Input depth (direct view) &
{\includegraphics[width=0.12\linewidth]{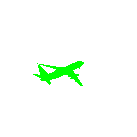}} &
{\includegraphics[width=0.12\linewidth]{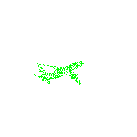}} &
{\includegraphics[width=0.12\linewidth]{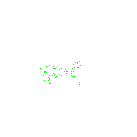}} &
{\includegraphics[width=0.12\linewidth]{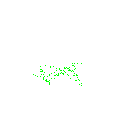}} &
{\includegraphics[width=0.12\linewidth]{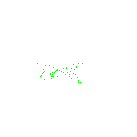}} &
{\includegraphics[width=0.12\linewidth]{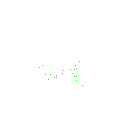}}
\\
\hline
DeepSDF \cite{park2019deepsdf} (direct view) &
{\includegraphics[width=0.12\linewidth]{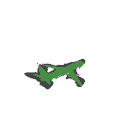}} &
{\includegraphics[width=0.12\linewidth]{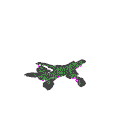}} &
{\includegraphics[width=0.12\linewidth]{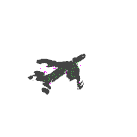}} &
{\includegraphics[width=0.12\linewidth]{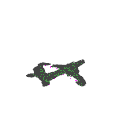}} &
{\includegraphics[width=0.12\linewidth]{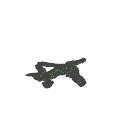}} &
{\includegraphics[width=0.12\linewidth]{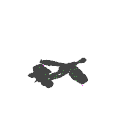}}
\\
DeepSDF \cite{park2019deepsdf} (second view) &
{\includegraphics[width=0.12\linewidth]{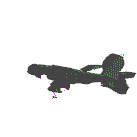}} &
{\includegraphics[width=0.12\linewidth]{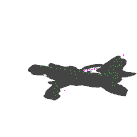}} &
{\includegraphics[width=0.12\linewidth]{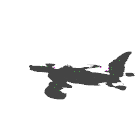}} &
{\includegraphics[width=0.12\linewidth]{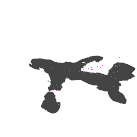}} &
{\includegraphics[width=0.12\linewidth]{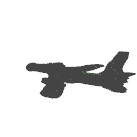}} &
{\includegraphics[width=0.12\linewidth]{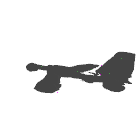}}
\\
DeepSDF \cite{park2019deepsdf} (third view) &
{\includegraphics[width=0.12\linewidth]{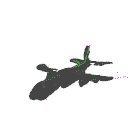}} &
{\includegraphics[width=0.12\linewidth]{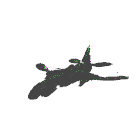}} &
{\includegraphics[width=0.12\linewidth]{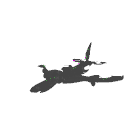}} &
{\includegraphics[width=0.12\linewidth]{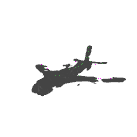}} &
{\includegraphics[width=0.12\linewidth]{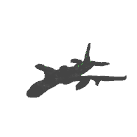}} &
{\includegraphics[width=0.12\linewidth]{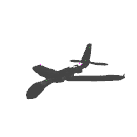}}
\\
\hline
Ours (direct view) &
{\includegraphics[width=0.12\linewidth]{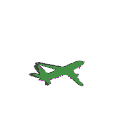}} &
{\includegraphics[width=0.12\linewidth]{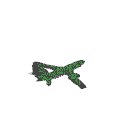}} &
{\includegraphics[width=0.12\linewidth]{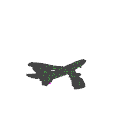}} &
{\includegraphics[width=0.12\linewidth]{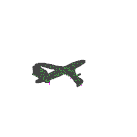}} &
{\includegraphics[width=0.12\linewidth]{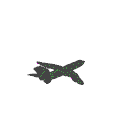}} &
{\includegraphics[width=0.12\linewidth]{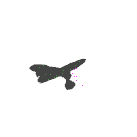}}
\\
Ours (second view) &
{\includegraphics[width=0.12\linewidth]{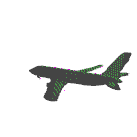}} &
{\includegraphics[width=0.12\linewidth]{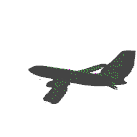}} &
{\includegraphics[width=0.12\linewidth]{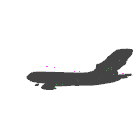}} &
{\includegraphics[width=0.12\linewidth]{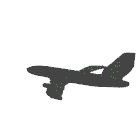}} &
{\includegraphics[width=0.12\linewidth]{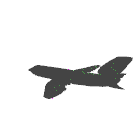}} &
{\includegraphics[width=0.12\linewidth]{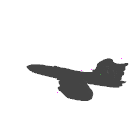}}
\\
Ours (third view) &
{\includegraphics[width=0.12\linewidth]{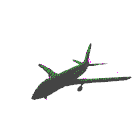}} &
{\includegraphics[width=0.12\linewidth]{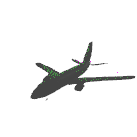}} &
{\includegraphics[width=0.12\linewidth]{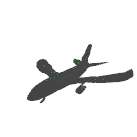}} &
{\includegraphics[width=0.12\linewidth]{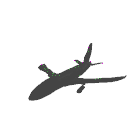}} &
{\includegraphics[width=0.12\linewidth]{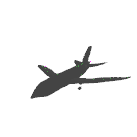}} &
{\includegraphics[width=0.12\linewidth]{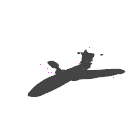}}
\\
\hline
Ours w. mask (direct view) &
{\includegraphics[width=0.12\linewidth]{img_src/supp_depth_completion/plane_53/ours_10_direct_output.png}} &
{\includegraphics[width=0.12\linewidth]{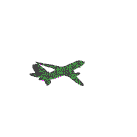}} &
{\includegraphics[width=0.12\linewidth]{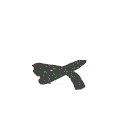}} &
{\includegraphics[width=0.12\linewidth]{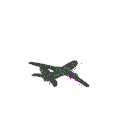}} &
{\includegraphics[width=0.12\linewidth]{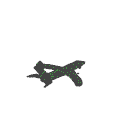}} &
{\includegraphics[width=0.12\linewidth]{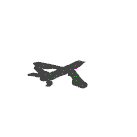}}
\\
Ours w. mask (second view) &
{\includegraphics[width=0.12\linewidth]{img_src/supp_depth_completion/plane_53/ours_10_az210_output.png}} &
{\includegraphics[width=0.12\linewidth]{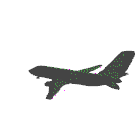}} &
{\includegraphics[width=0.12\linewidth]{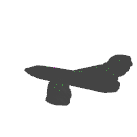}} &
{\includegraphics[width=0.12\linewidth]{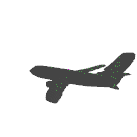}} &
{\includegraphics[width=0.12\linewidth]{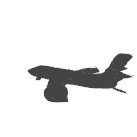}} &
{\includegraphics[width=0.12\linewidth]{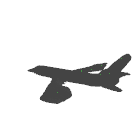}}
\\
Ours w. mask (third view) &
{\includegraphics[width=0.12\linewidth]{img_src/supp_depth_completion/plane_53/ours_10_az120_output.png}} &
{\includegraphics[width=0.12\linewidth]{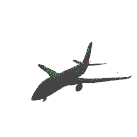}} &
{\includegraphics[width=0.12\linewidth]{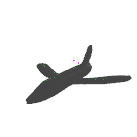}} &
{\includegraphics[width=0.12\linewidth]{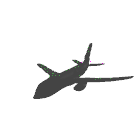}} &
{\includegraphics[width=0.12\linewidth]{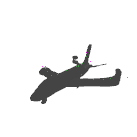}} &
{\includegraphics[width=0.12\linewidth]{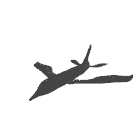}}
\\
\hline
\end{tabular}
\vspace{5pt}
\centering
\caption{Qualitative comparisons on shape completion under different sparsity of input depth (plane).}

\label{fig::supp_depth_completion_plane}
\vspace{-5pt}
\end{figure*}

\begin{figure*}[!tb]
\centering
\scriptsize
\setlength\tabcolsep{2.3pt} 
\begin{tabular}{c|cccccc}
\hline
& dense & 50\% pts & 10\% pts & 100 pts & 50 pts & 20 pts \\
\hline
Input depth (direct view) &
{\includegraphics[width=0.12\linewidth]{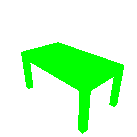}} &
{\includegraphics[width=0.12\linewidth]{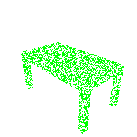}} &
{\includegraphics[width=0.12\linewidth]{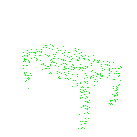}} &
{\includegraphics[width=0.12\linewidth]{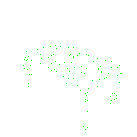}} &
{\includegraphics[width=0.12\linewidth]{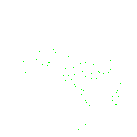}} &
{\includegraphics[width=0.12\linewidth]{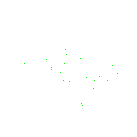}}
\\
\hline
DeepSDF \cite{park2019deepsdf} (direct view) &
{\includegraphics[width=0.12\linewidth]{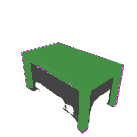}} &
{\includegraphics[width=0.12\linewidth]{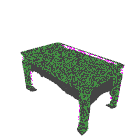}} &
{\includegraphics[width=0.12\linewidth]{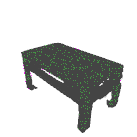}} &
{\includegraphics[width=0.12\linewidth]{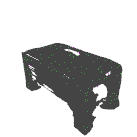}} &
{\includegraphics[width=0.12\linewidth]{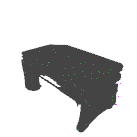}} &
{\includegraphics[width=0.12\linewidth]{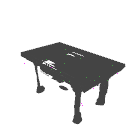}}
\\
DeepSDF \cite{park2019deepsdf} (second view) &
{\includegraphics[width=0.12\linewidth]{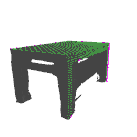}} &
{\includegraphics[width=0.12\linewidth]{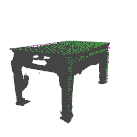}} &
{\includegraphics[width=0.12\linewidth]{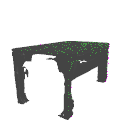}} &
{\includegraphics[width=0.12\linewidth]{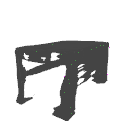}} &
{\includegraphics[width=0.12\linewidth]{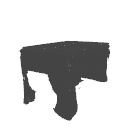}} &
{\includegraphics[width=0.12\linewidth]{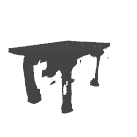}}
\\
DeepSDF \cite{park2019deepsdf} (third view) &
{\includegraphics[width=0.12\linewidth]{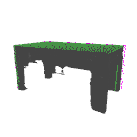}} &
{\includegraphics[width=0.12\linewidth]{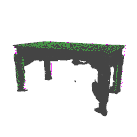}} &
{\includegraphics[width=0.12\linewidth]{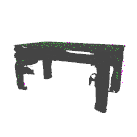}} &
{\includegraphics[width=0.12\linewidth]{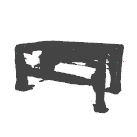}} &
{\includegraphics[width=0.12\linewidth]{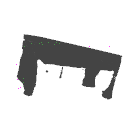}} &
{\includegraphics[width=0.12\linewidth]{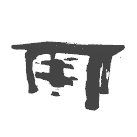}}
\\
\hline
Ours (direct view) &
{\includegraphics[width=0.12\linewidth]{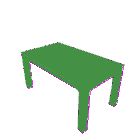}} &
{\includegraphics[width=0.12\linewidth]{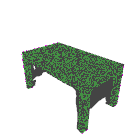}} &
{\includegraphics[width=0.12\linewidth]{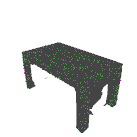}} &
{\includegraphics[width=0.12\linewidth]{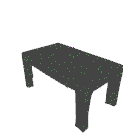}} &
{\includegraphics[width=0.12\linewidth]{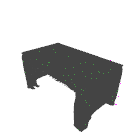}} &
{\includegraphics[width=0.12\linewidth]{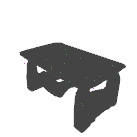}}
\\
Ours (second view) &
{\includegraphics[width=0.12\linewidth]{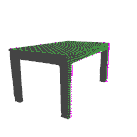}} &
{\includegraphics[width=0.12\linewidth]{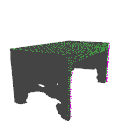}} &
{\includegraphics[width=0.12\linewidth]{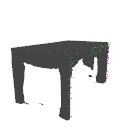}} &
{\includegraphics[width=0.12\linewidth]{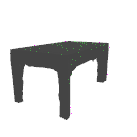}} &
{\includegraphics[width=0.12\linewidth]{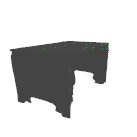}} &
{\includegraphics[width=0.12\linewidth]{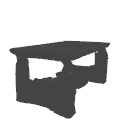}}
\\
Ours (third view) &
{\includegraphics[width=0.12\linewidth]{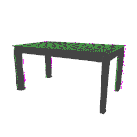}} &
{\includegraphics[width=0.12\linewidth]{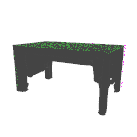}} &
{\includegraphics[width=0.12\linewidth]{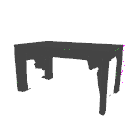}} &
{\includegraphics[width=0.12\linewidth]{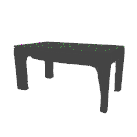}} &
{\includegraphics[width=0.12\linewidth]{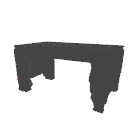}} &
{\includegraphics[width=0.12\linewidth]{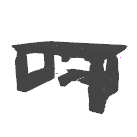}}
\\
\hline
Ours w. mask (direct view) &
{\includegraphics[width=0.12\linewidth]{img_src/supp_depth_completion/table_194/ours_10_direct_output.png}} &
{\includegraphics[width=0.12\linewidth]{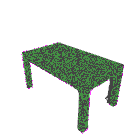}} &
{\includegraphics[width=0.12\linewidth]{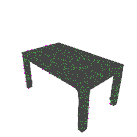}} &
{\includegraphics[width=0.12\linewidth]{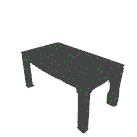}} &
{\includegraphics[width=0.12\linewidth]{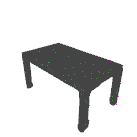}} &
{\includegraphics[width=0.12\linewidth]{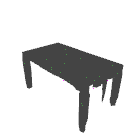}}
\\
Ours w. mask (second view) &
{\includegraphics[width=0.12\linewidth]{img_src/supp_depth_completion/table_194/ours_10_az210_output.png}} &
{\includegraphics[width=0.12\linewidth]{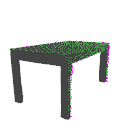}} &
{\includegraphics[width=0.12\linewidth]{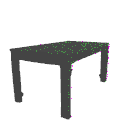}} &
{\includegraphics[width=0.12\linewidth]{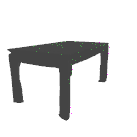}} &
{\includegraphics[width=0.12\linewidth]{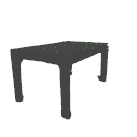}} &
{\includegraphics[width=0.12\linewidth]{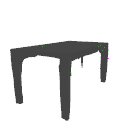}}
\\
Ours w. mask (third view) &
{\includegraphics[width=0.12\linewidth]{img_src/supp_depth_completion/table_194/ours_10_az120_output.png}} &
{\includegraphics[width=0.12\linewidth]{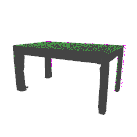}} &
{\includegraphics[width=0.12\linewidth]{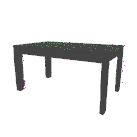}} &
{\includegraphics[width=0.12\linewidth]{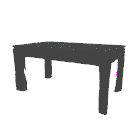}} &
{\includegraphics[width=0.12\linewidth]{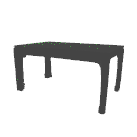}} &
{\includegraphics[width=0.12\linewidth]{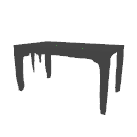}}
\\
\hline
\end{tabular}
\vspace{5pt}
\centering
\caption{Qualitative comparisons on shape completion under different sparsity of input depth (table).}

\label{fig::supp_depth_completion_table}
\vspace{-5pt}
\end{figure*}

We show qualitative comparison on shape completion under different sparsity of input depth in~\figref{fig::supp_depth_completion_sofa}, \figref{fig::supp_depth_completion_plane} and \figref{fig::supp_depth_completion_table}. The visual quality of our optimized mesh clearly outperforms the baseline method DeepSDF \cite{park2019deepsdf}. By employing perspective camera model and using online computed error (compared to the fixed sampling strategy in DeepSDF \cite{park2019deepsdf}) to perform inverse optimization on the rendered 2D observations, our method generates less holes in the predicted mesh when the input depth is sparse, particularly from the original view where the input depth is captured. When the silhouette information is available, our method can work reasonably well when the input depth is extremely sparse. Note that we give DeepSDF \cite{park2019deepsdf} the surface normal predicted by the dense ground-truth depth to make its sampling feasible. On the contrary, we do not use the surface normal information in the optimization of our method, which potentially can further improve our performance. Our method can generate reasonable occluded part with the assistance of the pretrained shape model prior. However, our method also could fail when the view of the camera cannot give sufficient information and the depth completion problem becomes extremely ill-posed.


\subsection{Qualitative Comparison for Multi-view Shape Prediction}
\begin{figure*}[!tb]
\centering
\newcommand{\rgb}{0.22\linewidth}
\scriptsize
\setlength\tabcolsep{6pt} 
\begin{tabular}{cccc}
Video sequence & PMO (rand init)& PMO & \textbf{Ours (rand init)}
\\
{\includegraphics[width=\rgb]{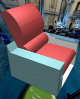}} &
{\includegraphics[width=0.24\linewidth]{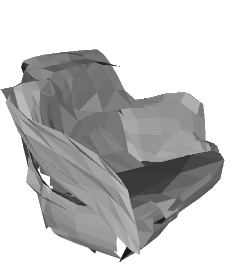}} &
{\includegraphics[width=0.22\linewidth]{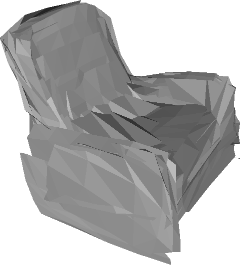}} &
{\includegraphics[width=0.22\linewidth]{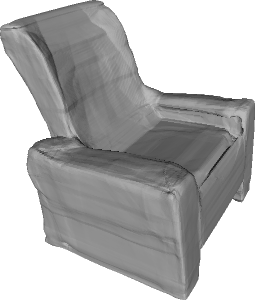}}
\\
{\includegraphics[width=\rgb]{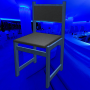}} &
{\includegraphics[width=0.18\linewidth]{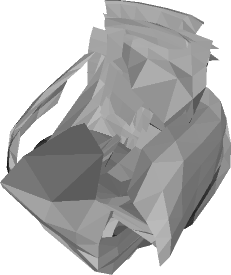}} &
{\includegraphics[width=0.12\linewidth]{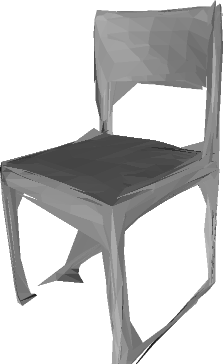}} &
{\includegraphics[width=0.14\linewidth]{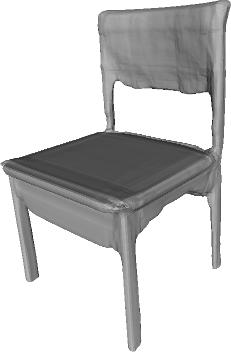}}
\\
{\includegraphics[width=\rgb]{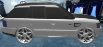}} &
{\includegraphics[width=0.2\linewidth]{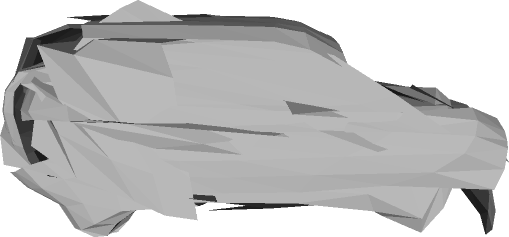}} &
{\includegraphics[width=0.2\linewidth]{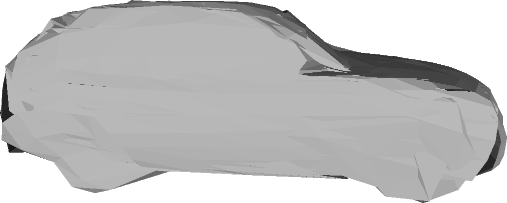}} &
{\includegraphics[width=0.2\linewidth]{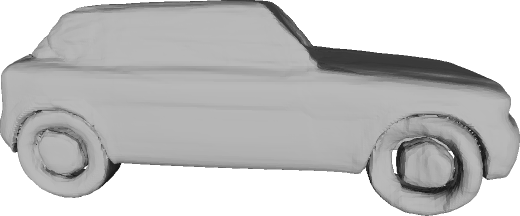}}
\\
{\includegraphics[width=\rgb]{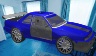}} &
{\includegraphics[width=0.18\linewidth]{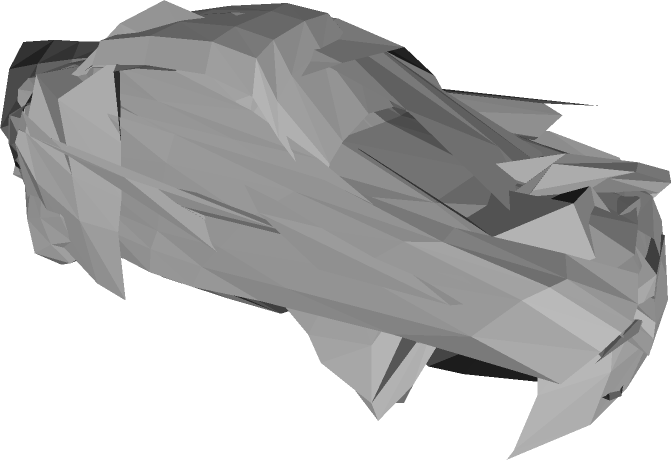}} &
{\includegraphics[width=0.2\linewidth]{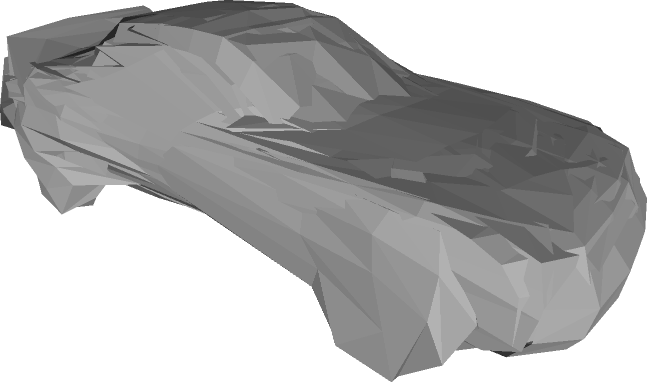}} &
{\includegraphics[width=0.2\linewidth]{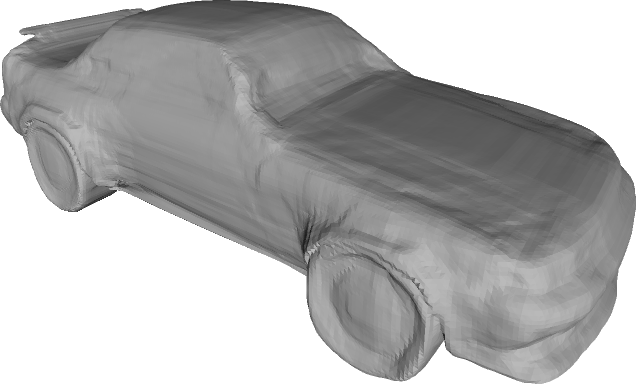}}
\\
{\includegraphics[width=\rgb]{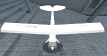}} &
{\includegraphics[width=0.11\linewidth]{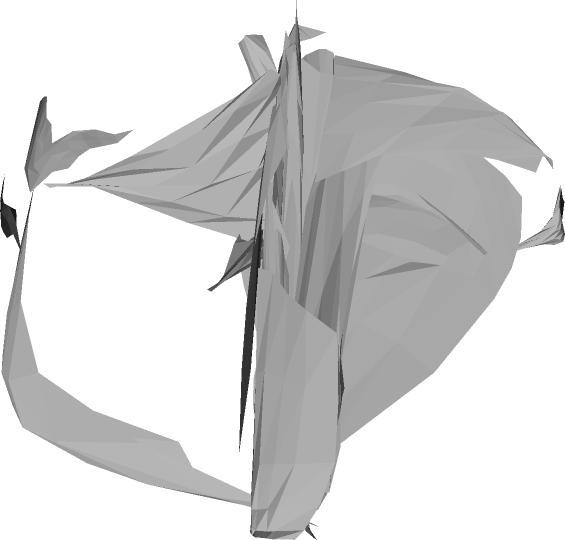}} &
{\includegraphics[width=0.18\linewidth]{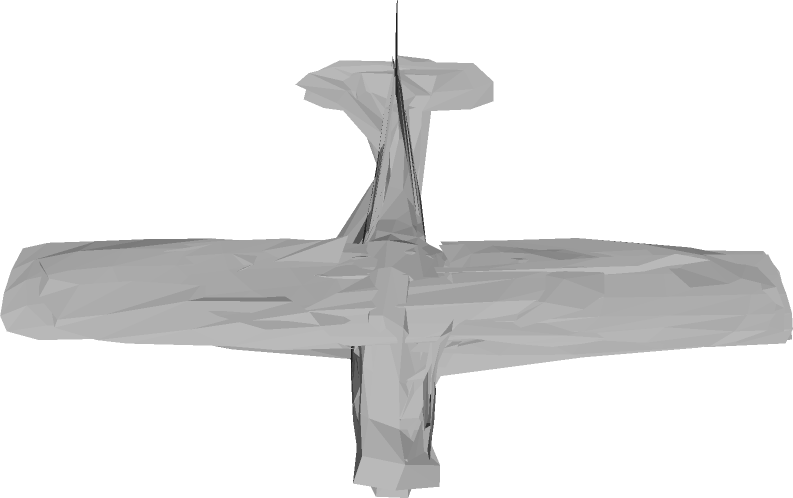}} &
{\includegraphics[width=0.18\linewidth]{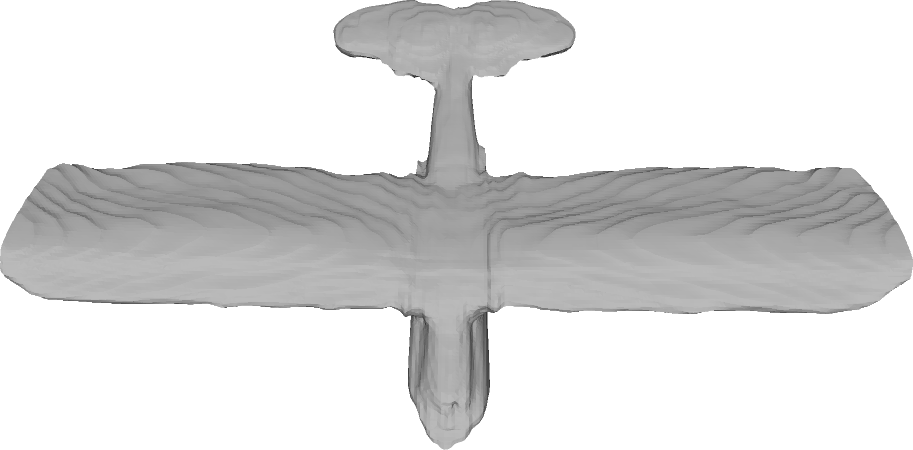}}
\\
{\includegraphics[width=\rgb]{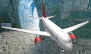}} &
{\includegraphics[width=0.16\linewidth]{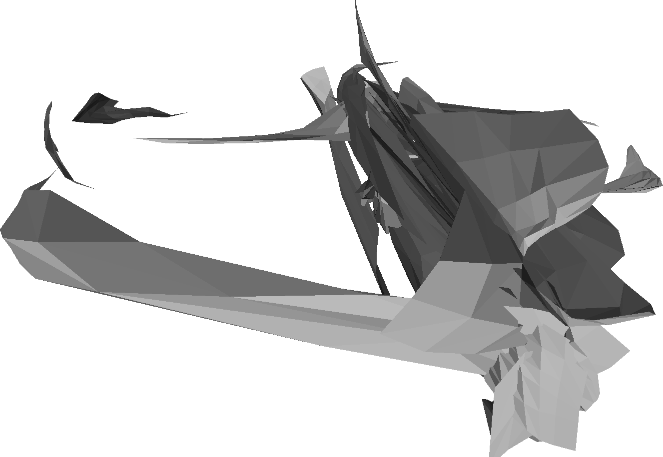}} &
{\includegraphics[width=0.18\linewidth]{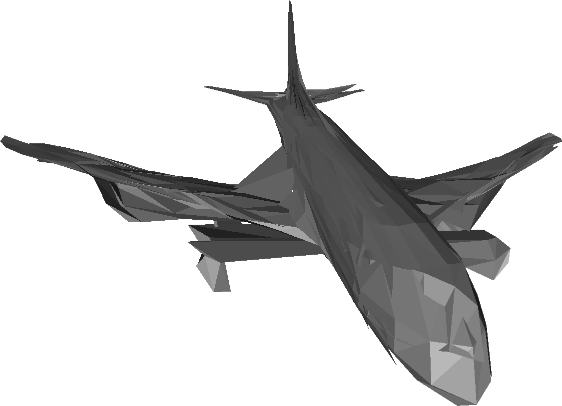}} &
{\includegraphics[width=0.18\linewidth]{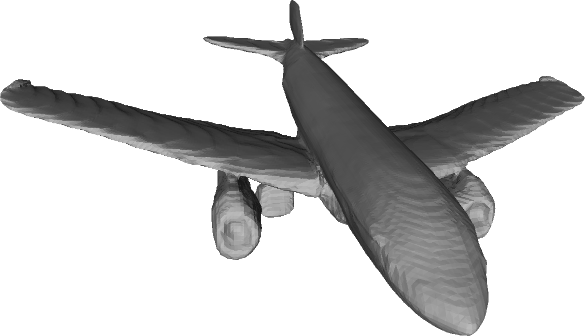}}
\\
{\includegraphics[width=\rgb]{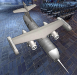}} &
{\includegraphics[width=0.2\linewidth]{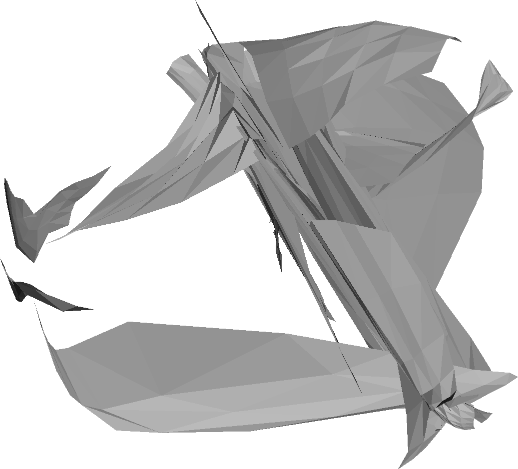}} &
{\includegraphics[width=0.22\linewidth]{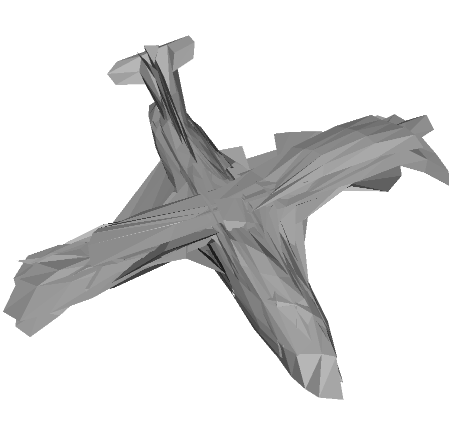}} &
{\includegraphics[width=0.21\linewidth]{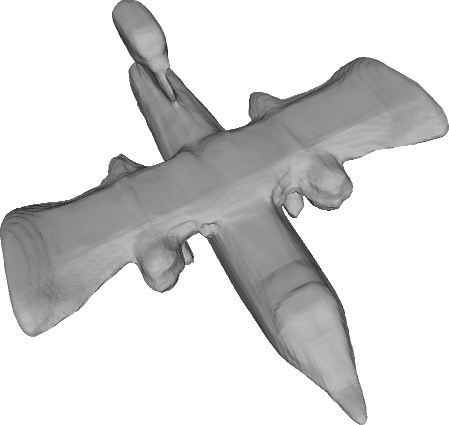}}
\\
\end{tabular}
\vspace{5pt}
\centering
\caption{More quatitative comparisons on 3D shape prediction from multi-view images on the PMO test set.}

\label{fig::supp_mv_comp_pmo_resized}
\end{figure*}

We firstly show more qualitative comparisons on PMO test dataset in~\figref{fig::supp_mv_comp_pmo_resized}. It can be noticed that our method produces much more visually satisfactory 3D shapes with only random initializations. In contrast, even if PMO~\cite{lin2019photometric} uses a much better initialization from the encoder pretrained on their PMO training set, their 3D shapes have low resolutions because of the limited number of vertices. Moreover, if the random initialized codes are applied, PMO fails to generate reasonable 3D shapes in most of the cases. When checking closely the second column in~\figref{fig::supp_mv_comp_pmo_resized}, the results even converge to rather similar shapes, especially for chairs and planes.

\begin{figure*}[!tb]
\centering
\newcommand{\rgb}{0.32\linewidth}
\newcommand{\shape}{0.2\linewidth}
\scriptsize
\setlength\tabcolsep{24pt} 
\begin{tabular}{ccc}
Video sequence & PMO & \textbf{Ours (rand init)}
\\
{\includegraphics[width=\rgb]{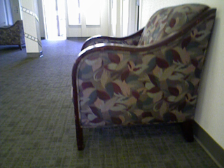}} &
{\includegraphics[width=\shape]{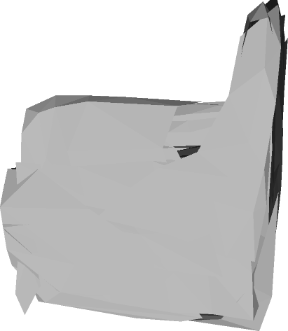}} &
{\includegraphics[width=\shape]{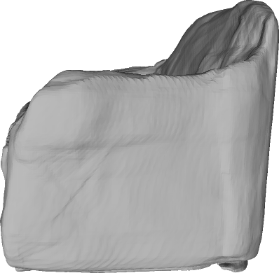}}
\\
{\includegraphics[width=\rgb]{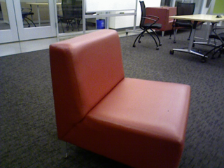}} &
{\includegraphics[width=\shape]{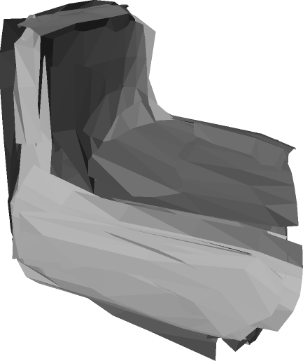}} &
{\includegraphics[width=\shape]{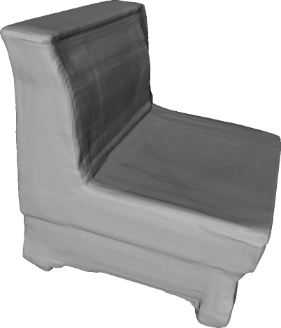}}
\\
{\includegraphics[width=\rgb]{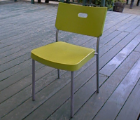}} &
{\includegraphics[width=0.15\linewidth]{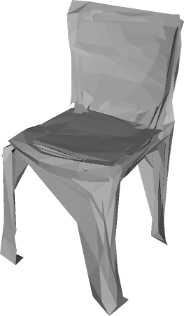}} &
{\includegraphics[width=\shape]{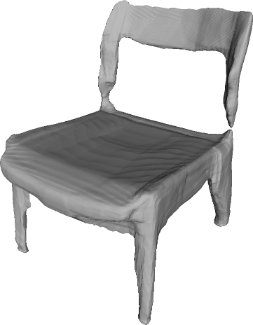}}
\\
{\includegraphics[width=\rgb]{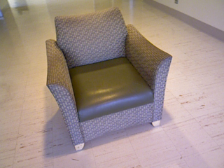}} &
{\includegraphics[width=\shape]{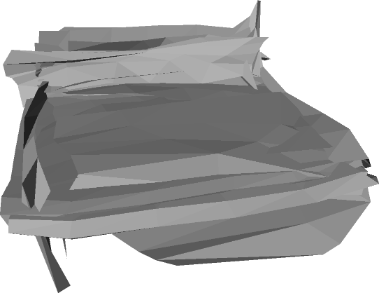}} &
{\includegraphics[width=\shape]{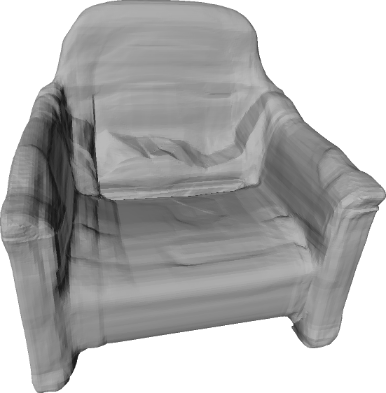}}
\\
{\includegraphics[width=\rgb]{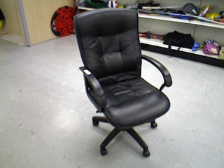}} &
{\includegraphics[width=0.18\linewidth]{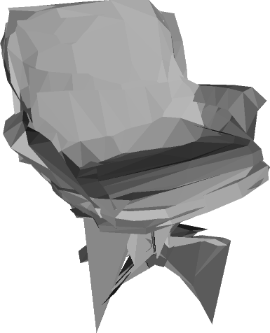}} &
{\includegraphics[width=0.12\linewidth]{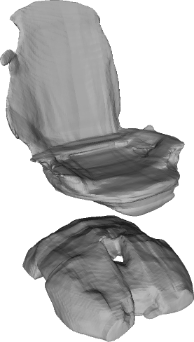}}
\end{tabular}
\vspace{5pt}
\centering
\caption{More Comparisons on 3D shape prediction from multi-view images on real-world chair dataset~\cite{choi2016large}. 
It is in general challenging for shape prediction on real images. Comparatively, our method produces more reasonable results with correct structure. 
}
\label{fig::supp_mv_comp_real_resized}
\end{figure*}

We also illustrate more results on the real-world multi-view chair dataset in~\figref{fig::supp_mv_comp_real_resized}. 
Note that similar to PMO, we now also optimize over the similarity transformation including rotation, translation and scale, together with the shape code. It can be noticed easily that our method with random initialization again generates much superior outputs over PMO.
We also show a failure case in the last row.  Our method fails when there is insufficient texture on foreground or background as photometric cues. Moreover, our method may fail when the similarity transformation can not be correctly estimated.

\end{document}